\documentclass[preprint,12pt]{elsarticle}

\usepackage{amssymb}
\usepackage{amsthm}
\usepackage[T5]{fontenc}
\usepackage[utf8]{inputenc}
\usepackage{booktabs}
\usepackage{float}
\usepackage{adjustbox}
\usepackage{caption}
\usepackage{subcaption}
\usepackage{graphicx}
\usepackage{tabularx}
\usepackage{multirow}
\usepackage{amsmath}
\usepackage{amsfonts}
\usepackage{algorithm}
\usepackage{algpseudocode}
\usepackage{url}
\usepackage{pifont}
\newcommand{\tabitem}{\textbullet~~}
\newcommand{\cmark}{\ding{51}}
\newcommand{\xmark}{\ding{55}}

\author{Nghia Hieu Nguyen$^{a, b}$}
\author{Duong T. D. Vo$^{a, b}$}
\author{Kiet Van Nguyen$^{a, b}$}
\author{Ngan Luu-Thuy Nguyen$^{a, b}$}

\address{$^a$Faculty of Information Science and Engineering, University of Information Technology, Ho Chi Minh city, Vietnam}
\address{$^b$Vietnam National University, Ho Chi Minh city, Vietnam}

\begin{document}

\begin{frontmatter}



\title{OpenViVQA: Task, Dataset, and Multimodal Fusion Models for Visual Question Answering in Vietnamese}


\begin{abstract}
In recent years, visual question answering (VQA) has attracted attention from the research community because of its highly potential applications (such as virtual assistance on intelligent cars, assistant devices for blind people, or information retrieval from document images using natural language as queries) and challenge. The VQA task requires methods that have the ability to fuse the information from questions and images to produce appropriate answers. Neural visual question answering models have achieved tremendous growth on large-scale datasets which are mostly for resource-rich languages such as English. However, available datasets narrow the VQA task as the answers selection task or answer classification task. We argue that this form of VQA is far from human ability and eliminates the challenge of the answering aspect in the VQA task by just selecting answers rather than generating them. In this paper, we introduce the OpenViVQA (\textbf{Open}-domain \textbf{Vi}etnamese \textbf{V}isual \textbf{Q}uestion \textbf{A}nswering) dataset, the first large-scale dataset for VQA with open-ended answers in Vietnamese, consists of \textbf{11,000+} images associated with \textbf{37,000+} question-answer pairs (QAs). Moreover, we proposed FST, QuMLAG, and MLPAG which fuse information from images and answers, then use these fused features to construct answers as humans iteratively. Our proposed methods achieve results that are competitive with SOTA models such as SAAA, MCAN, LORA, and M4C. The dataset\footnote{\url{https://github.com/hieunghia-pat/OpenViVQA-dataset}} is available to encourage the research community to develop more generalized algorithms including transformers for low-resource languages such as Vietnamese.
\end{abstract}





\begin{keyword}

Visual question answering \sep vision-language understanding \sep low-resource languages \sep information fusion \sep multimodal representation

\end{keyword}

\end{frontmatter}


\section{Introduction}
\label{sect:introduction}
Visual Question Answering (VQA) is an information fusion task that takes an image and a question as input, the computers are required to produce an answer for that question based on the information from the image \cite{VQA}. As one of the most challenging tasks, visual question answering recently has attracted lots of attention from both the computer vision (CV) and natural language processing (NLP) research communities. This task requires approaches to have the ability to understand the linguistic concepts of the given question and then use the visual concepts in the respective image to answer the given question. VQA task is motivated by the need to retrieve information from multiple sources including using natural questions to query information from videos or document images.

Despite having been existing since 2015 \cite{VQA} and being researched widely for high-resource languages such as English, there are few studies exploring VQA tasks for Vietnamese as one of the low-resource languages. In 2021, Tran et al. \cite{tran-etal-2021-vivqa-vietnamese} built the ViVQA dataset, the first dataset for research on the VQA task for Vietnamese. The ViVQA dataset covers a small part of the VQAv2 dataset as it was constructed using the semi-automatic method \cite{tran-etal-2021-vivqa-vietnamese}. We assume with the lack of validation as well as the drawbacks of the machine translation results, the effectiveness of the semi-automatic method proposed in \cite{tran-etal-2021-vivqa-vietnamese} is not ensured to reach human performance when translating a VQA dataset for English to Vietnamese. The ViVQA dataset therefore can not be a reliable benchmark for exploring and constructing experiments as well as evaluating VQA systems, and we need a novel and manually annotated VQA dataset to use as a benchmark for research on VQA tasks in Vietnamese.

On the other hand, as most VQA datasets treat VQA task as classification task \cite{goyal2017making}, we argue this approach is not reasonable and far from human ability because humans can normally answer questions using various forms of natural languages such as words, phrases or full sentences. 

To overcome the above limitations arising from the datasets and the task itself, we define a different form of the VQA task, the open-ended VQA, which has open-ended questions and open-ended answers (Section \ref{sect:open-ended}). Then based on this novel definition, we introduce the OpenViVQA (Open-domain Visual Question Answering for Vietnamese) dataset including \textbf{11,199} images together with \textbf{37,914} question-answer pairs (QA) annotated manually. In particular, images in the OpenViVQA dataset are captured in Vietnam and thus are valuable resources for the research community to study and explore the difference between images captured in Vietnam and those captured in other regions outside of Vietnam. This region-specific set of images can motivate the development of proper pre-trained models that suit our dataset. 

In addition, through our experiments, we show that former methods on English VQA datasets fail when tackling the open-ended VQA task, especially on our novel dataset. For this reason, we propose three multimodal and answer generation methods and how they obtain better results. These proposed methods can be used as preliminary baselines for further research on our dataset particularly or open-ended VQA tasks generally. 

The structure of this paper is detailed as follows: Section \ref{sect:related_works} presents works related to our study. In Section \ref{sect:dataset}, we describe the data creation as well as data validation and display challenges of the OpenViVQA dataset. Then, we introduce our three proposed methods in Section \ref{sect:method}. Section \ref{sect:experiment} provides information on how we design the experiments as well as evaluate the results of the baselines and our proposed methods. We provide explanations of why challenges from OpenViVQA influence the results of the methods in Section \ref{sect:result_analysis}. Finally, we summarize our study as well as propose future works in Section \ref{sect:conclusion}.

 \section{Related Works}
\label{sect:related_works}

\begin{figure}[ht]
    \centering
    \includegraphics[width=\textwidth]{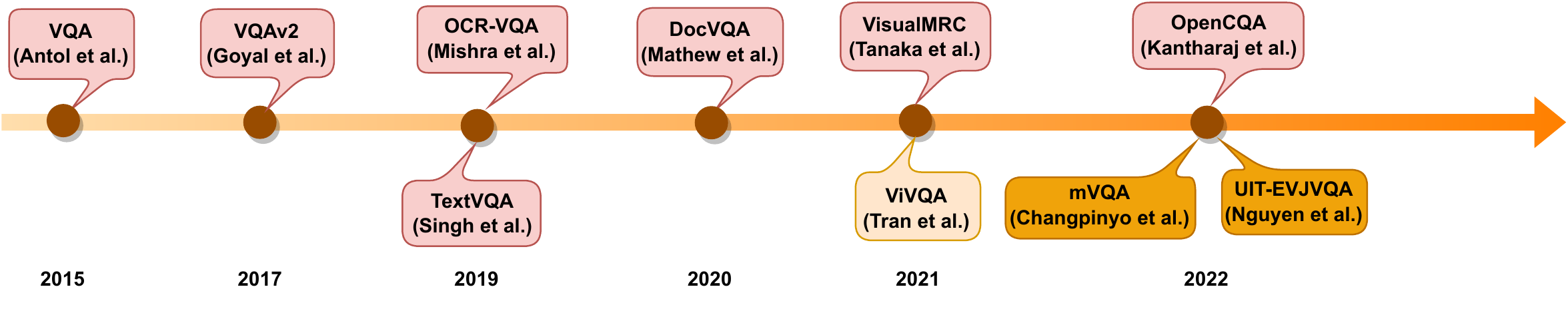}
    \caption{Timeline of VQA datasets}
    \label{fig:vqa_timeline}
\end{figure}

In this section, we briefly review noticeable works of constructing VQA datasets including VQA datasets in English and Vietnamese (Figure \ref{fig:vqa_timeline}). Then we review studies that impact our ideas of designing multimodal methods.

\subsection{Visual Question Answering datasets}

\subsubsection{Former Visual Question Answering Datasets}

Antol et al. \cite{VQA} introduced the VQAv1 dataset including 254,721 images with 764,163 questions and 4,598,610 answers, defining the VQA task in English. The VQAv1 dataset uses 204,721 images from MS COCO \cite{lin2014microsoft} dataset and 50,000 additional abstract scene images. With the release of the VQAv1 dataset, lots of attention are drawn and many methods are proposed \cite{kazemi2017show,lu2016hierarchical,yang2016stacked} to tackle the VQA task. 

Although Antol et al. \cite{VQA} intended to introduce the open-ended VQA task via the VQAv1 dataset, Teney et al. \cite{teney2018tips} after conducting many estimations as well as statistics they showed the answers in the VQAv1 dataset can be treated as a category. They henceforth proposed to tackle the VQA task as a classification task where the machine is trained to classify answers, or in the other word, select answers from a defined set of answers rather than generate answers. 

However, as with other classification tasks, the imbalance in categories is significant and usually leads to the overconfident performance of machine learning methods. Goyal et al. \cite{goyal2017making} discovered this phenomenon while they observed and statistically analyzed the results of VQA methods. Particularly, the phenomenon resembles the class imbalance in classification tasks where models tend to give most appearance answers for a particular type of question. For instance, when models are given a question starting with "how many", they immediately select "two" as the respective answer since this is the most occurrence answer among the answers for questions on quantity. Goyal et al. \cite{goyal2017making} named this phenomenon language-prior to emphasize the dependency of models on questions when providing answers while totally ignoring images. 

To overcome this phenomenon, Goyal et al. \cite{goyal2017making} reannotated the VQAv1 dataset by providing a question with different answers over different images, hence balancing the occurrence of answers for any particular type of questions, forcing the models to use information from images to give answers. The novel balanced dataset was then released under the name VQAv2. It proved that many SOTA VQA methods at that time suffered such language-prior phenomenon hence their good results on the VQAv1 test dataset are not reliable. Later, based on the classification approaches, together with the instruction of attention mechanism \cite{vaswani2017attention,bahdanau2014neural,luong2015effective}, most SOTA methods were designed based on Co-Attention strategy where they attempted to use attention on the information of questions and images in a fusing fashion and then proceeded with the selection of answers.

\subsubsection{Visual Question Anwering Datasets with Reading Comprehension}

Although there has been studied for many years, the research community mostly concentrates on tackling VQA tasks in which questions query details about objects and context in images. In 2019, Singh et al. \cite{singh2019towards} pointed out that VQA methods do not have the ability to read and therefore introduced a novel dataset, the TextVQA dataset, in which every question exploits the scene texts that appear in the images. This novel definition of the VQA task has attracted lots of works \cite{singh2019towards}. Many similar datasets were released at the same time such as OCR-VQA \cite{mishraICDAR19}, DocVQA \cite{Mathew_2021_WACV} or VisualMRC \cite{visualmrc}. 

However, in Figure \ref{fig:question_length_results}, Figure \ref{fig:answer_length_results} and \ref{appendix-A}, we provide statistics as well as examples that such English VQA datasets are not as challenging as the OpenViVQA dataset. Answers in the OpenViVQA dataset associate the scene texts from the images as a part to provide more detailed and specific information as a human-like style of description. On the contrary, answers of VQA datasets in English are simply plain scene texts available in the images.

\subsubsection{Open-ended Visual Question Answering Datasets}

Along with the trend of VQA research in open-ended form, there are similar datasets where its answers are annotated similarly to answers in the OpenViVQA dataset in form of open-ended form such as the VisualMRC dataset \cite{visualmrc} or OpenCQA \cite{Kantharaj2022OpenCQAOQ}. Another dataset released by Google is mVQA \cite{Changpinyo2022TowardsMV} which is a multilingual version of the VQAv2 dataset where answers are in open-ended form after being translated into non-English languages. 

Constructing an open-ended VQA dataset has many advantages. Open-ended VQA datasets eliminate the language-prior phenomenon, another color of imbalance classes in the classification task. Moreover, open-ended answers of these datasets guide the community to research and explore methods that can give answers by using human languages naturally and fluently rather than selecting an answer from a defined set, thus bridging the gap between humans and machines.

\subsubsection{Visual Question Answering Datasets in Vietnamese}

Although there are many benchmarks for researching the VQA task in English, there is no resource available for low-resource languages, particularly Vietnamese. In 2019, Tran et al. \cite{tran-etal-2021-vivqa-vietnamese} published the ViVQA dataset, the first VQA dataset in Vietnamese. They took advantage of machine translation to translate questions and answers from a portion of the VQAv2 dataset to Vietnamese, then they manually validate the correctness and fluency of translated questions and answers. However, our examples in \ref{appendix-A} point out that this kind of semi-automatic method does not ensure the quality of translated sentences hence this dataset is not reliable to be used as a benchmark for researching and evaluating Vietnamese VQA models. To this end, we construct and introduce the first large-scale and manually annotated VQA dataset to the NLP research community, providing a novel benchmark for further research as well as validating pre-trained vision-language models for VQA in Vietnamese.

\subsection{Visual Question Answering methods}

VQA methods include three main components: the external information embedding module, the multi-source information fusion module, and the answer classifier or answer generator module. The development of VQA methods currently concentrates on improving the external information embedding module and the multi-source information fusion module.

Initial approaches used pre-trained image models such as ResNet \cite{He2015DeepRL} to extract features from images \cite{kazemi2017show,lu2016hierarchical,yang2016stacked}. Anderson et al. \cite{Anderson2017BottomUpAT} proposed the Bottom-up Top-down mechanism for the VQA task by using FasterRCNN \cite{ren2015faster} to extract region features from images. This way of feature extraction is effective as it reduces noises introduce by the regions of images that are not relevant to given questions. Jiang et al. \cite{jiang2020defense} conducted experiments to prove that when using grid features as the input features of images for the attention-based deep learning method they can perform approximately the same performance as the Bottom-up Top-down attention mechanism. Recently there are pre-trained models that leverage the information extraction for VQA tasks such as Oscar \cite{li2020oscar} or VinVL \cite{zhang2021vinvl}, such models enhance significantly results of VQA methods on various datasets \cite{li2020oscar,zhang2021vinvl}.

Former methods used pre-trained word embedding such as FastText \cite{bojanowski2017enriching} or GloVe \cite{pennington-etal-2014-glove} together with LSTM \cite{hochreiter1997long} network to extract linguistic features. Recent studies \cite{yu2019deep,hu2020iterative} used large language models (LLM) such as BERT \cite{Devlin2019BERTPO} to leverage the linguistic feature extraction and achieved positive results.

Besides the way of extracting information from multiple sources, fusing information is also a crucial part of output features that select or construct appropriate answers. Most VQA methods concentrated on the attention mechanism \cite{luong2015effective,bahdanau2014neural,vaswani2017attention} to perform information fusion, or in other words, they perform the attention mechanism to determine the correlation among multiple sources of information. According to the survey study \cite{ZHANG2019268}, the attention strategy for information fusion can be divided into two categories based on the number of attention layers: single-hop attention and multi-hop attention. Previous works implemented single-hop attention such as \cite{yang2016stacked,lu2016hierarchical,kazemi2017show} but these methods did not obtain promising results while multi-hop attention methods such as \cite{yu2019deep,ZHANG20211,ZHENG202114,ZHANG2020116} achieved better ones. These results showed that the attention module with a single layer can not model properly the reasoning ability required in VQA tasks. Recent studies enhance this view of point by performing co-attention mechanisms with multiple transformer layers \cite{vaswani2017attention} such as ViLBERT \cite{lu2019vilbert}, VisualBERT \cite{li2019visualbert}, LXMERT \cite{tan2019lxmert}, VL-BERT \cite{su2019vl}, Unicoder-VL \cite{li2020unicoder}, Uniter \cite{chen2020uniter}, X-LXMERT \cite{cho2020x}, Pixel-BERT \cite{huang2020pixel}, VLMo \cite{lu2022unified}, or SimVLM \cite{wang2021simvlm}.

\section{OpenViVQA Dataset}
\label{sect:dataset}

The OpenViVQA dataset is designed in a way that exhibits the open-ended property of questions and answers to facilitate research on generating answers to the VQA task. Moreover, our dataset must exploit visual scenes from Vietnam to associate with the Vietnamese language aspects and clearly express our contribution to the research community in Vietnam. In this way, we can also question the information from the scene text in such images in Vietnamese. In this section, we first give a detailed definition of the open-ended VQA task. After that, we describe the process of creating the OpenViVQA dataset, from image collection to questions and answers creation. Finally, we show the overall statistics and analysis of our dataset.

\subsection{Open-ended Visual Question Answering Task} \label{sect:open-ended}
Previous studies \cite{VQA, goyal2017making,Kantharaj2022OpenCQAOQ,visualmrc} used the term open-ended questions without defining them explicitly. In this paper, based on works in linguistics, we define the open-ended questions and open-ended answers for the VQA task, then explain carefully how we can ensure that questions and answers in the OpenViVQA dataset are open-ended.

According to Worley et al. \cite{worley2015open}, the openness and closeness of questions are defined depending on two aspects: conception and grammar. Conceptually, open-ended questions "contain or invite tensions, conflicts, or controversies in the concepts contained within the question itself" \cite{worley2015open} and close-ended questions otherwise. Grammatically open-ended questions are questions that demand answers having more than one word or answers that are short-phrase. 

In the conceptual aspect, open-ended questions have open topics, and their answers can be formed from diverse domains of knowledge and point of view (invite tensions, conflicts, or controversies in the concepts \cite{worley2015open}). While in the VQA task, answers are limited to the context or content available in the images only. Hence defining the open-ended questions following the conceptual aspect is not suitable.

Considering the definition of open-ended questions in the grammatical aspect, Worley et al. \cite{worley2015open} grammatically defined the open-ended questions of a VQA dataset are questions that require more than one-word answers or short-phrase answers \cite{worley2015open}. VQA datasets such as VQAv2 \cite{goyal2017making} was stated they are open-ended questions VQA datasets. However, according to Table \ref{tab:level_comparison}, one-word answers occupy 89,41\% total answers, which means most of the questions in VQAv2 are close-ended, which is opposed to the statement of previous studies \cite{goyal2017making}. From that on, determining the open-ended feature of questions by the grammar of their answers is not reasonable in the spirit of previous works. This color is available as well in the latter VQA datasets such as OCR-VQA dataset \cite{mishraICDAR19} and TextVQA dataset \cite{singh2019towards} where close-ended questions are the most ones.

To comprehensively inspire the open-ended definition of previous studies, we define the open-ended feature of questions in a VQA dataset based on their grammar rather than the grammar of their answers. Particularly, a VQA dataset is an open-ended question VQA dataset if all questions in that dataset do not share common patterns that can be determined easily by a heuristic algorithm. From that on, questions in the VQAv1 dataset, VQAv2 dataset, TextVQA dataset, or OpenViVQA dataset are open-ended questions as they were constructed by crowdsourcing and have the diverse length as well as complicated semantic dependencies (Section \ref{sect:comparison}). In contrast, questions in OCR-VQA are not open-ended as they were defined by particular patterns of questions (\ref{appendix-A}).

We define a VQA dataset as a close-ended answers VQA dataset if it satisfies one of the following conditions:
\begin{itemize}
    \item All answers are words or phrases (the way of determining when an answer is a word or a phrase is detailed in Section \ref{sect:comparison}).
    \item VQA methods that sample answers from a defined set (or answer selection) achieved better results than VQA methods that construct answers by sequentially sampling tokens from a defined set (or answer generation).
\end{itemize}

VQA datasets that are not close-ended answers VQA dataset are open-ended answers VQA dataset.

The first condition identifies with the definition of close-ended questions in the study of Worley et al. \cite{worley2015open}. The second definition generalizes the classification of answers in question answering (QA) tasks (e.g., multiple choice QA where the QA methods have to make the most accurate selection A, B, C, or D). Note that we do not mention the extracted answer of some QA tasks as their essence is the information extraction tasks rather than QA tasks.

Why do we have the second condition? Is using the behavior of VQA methods to determine the openness of answers reasonable? Actually, we propose this second condition inspired by the study of Teney et al. \cite{teney2018tips}. In this study, Teney et al. and various later studies \cite{goyal2017making,yu2019deep,lu2016hierarchical,kazemi2017show} proposed on the VQAv1 \cite{VQA} and VQAv2 dataset \cite{goyal2017making} VQA methods that sample answers from a defined set. That is, if we define on the VQAv2 dataset, for example, the probability measure:

\begin{equation}
    \mu(a) = \mu(\{a\}) = \frac{\#a}{\#\omega}
\end{equation}
where $\omega$ is the set of all answers in the VQAv2 dataset, $a \in \omega$ is an answer, $\#a$ indicates the total times of appearance of $a$ in $\omega$, and $\#A = \sum_{a \in A}(\#a)$ $(A \subset \omega$), then, the existence of answers selection VQA methods implies for most $a \in \omega$, $\mu(a)$ is so significant that using VQA methods to approximate the distribution of answers is more effective than generating answers from individual tokens. This approach is absolutely identified with the multiple choice QA tasks where answers are sampled from a defined set. This is why using the behavior of VQA methods to indicate the openness of answers is reasonable.

Following our definition of open-ended answers, answers in the VQAv1 \cite{VQA} and VQAv2 \cite{goyal2017making} are close-ended answers as indicated by Teney et al. \cite{teney2018tips}. On the other hand, our experiments showed that answer selection VQA methods failed when approaching the OpenViVQA dataset, thus breaking the second condition of our definition of close-ended answers. Moreover, from Table \ref{tab:linguistic_comparison}, answers in the OpenViVQA dataset have diverse linguistic levels, which violates the first condition of close-ended answers. Hence answers in the OpenViVQA dataset are ensured to be open-ended.


Finally, we define the open-ended VQA dataset comprising open-ended questions and their answers. Open-ended VQA tasks are VQA tasks that are defined by open-ended VQA datasets.

\subsection{Dataset Creation}

\begin{figure}[ht]
    \centering
    \includegraphics[width=1\linewidth]{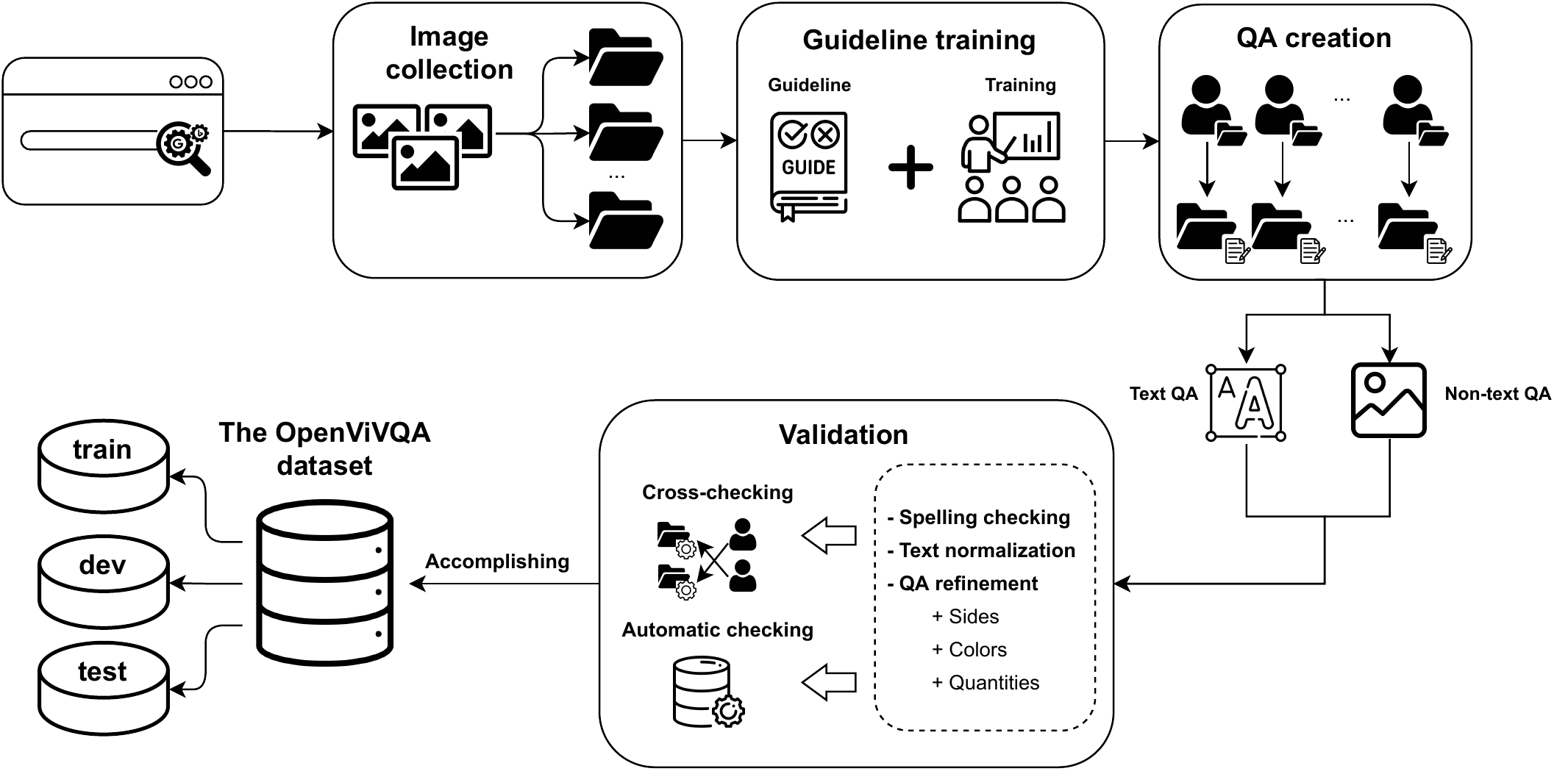}
    \caption{Overall process for the creation of the OpenViVQA dataset.}
    \label{fig:openvivqa-creation}
\end{figure}

The creation process of the OpenViVQA dataset can be described in the following diagram in Figure \ref{fig:openvivqa-creation}. First, we collect the images and then distribute them into multiple subsets. We then design the guideline and base it on training the employed crowd-workers. After the training process, the questions and answers (QAs) creation stage officially starts. Pairs of questions and answers are formed for each image and are classified into Text QA or Non-text QA. The next step is dataset validation and adjustment, where we check for spelling mistakes and ensure consistency toward the expected quality. After that, the OpenViVQA dataset is completed and divided into training, development, and test sets for the experiments.

\subsubsection{Image Collection}
In our work, we diversify the image set from other related works in English since they only exploit the visual scenes in Western locations \cite{lin2014microsoft,VQA}, which do not truly represent the distinct characteristics of scenery in Vietnam. Specifically, images that originate from Vietnamese scenes often depict the populous streets with motorcycles and vendors and the unmistakable Vietnamese lifestyles of the people. These images will facilitate the use of Vietnamese words to question and describe culturally specific concepts. Moreover, since we also need to use the Vietnamese scene text for our research, images captured in Vietnam are the best fit to be the foundation of the OpenViVQA dataset.

We first prepare a set of keywords that represent a wide range of concepts, such as human lifestyle and activities, means of transport, interiors, markets, streets, and cultural sites. We also appended Vietnamese locations like Hanoi or Saigon to some of the keywords for more diversity in geological and cultural contexts. For instance, some of the keywords can be "quán ăn ở Hà Nội" (eateries in Hanoi), "đường phố ở Sài Gòn" (streets in Saigon), "chợ Việt Nam" (Vietnamese markets), "bảng hiệu cửa hàng" (store signboards) or "sinh viên đi dã ngoại" (students taking a trip). We then pass these keywords into Google Images and gain multiple collections of images corresponding to the keywords using both the Octoparse \footnote{https://www.octoparse.com/} scrapping tool and Python code. 

After collecting all the images, we proceed with the filtering stages. Since we have to keep most of the details visible for questioning, we only retain images whose resolution is 500x400 pixels or above. Image files that are of file types other than JPEG or PNG are also excluded. In addition, we manually observe and eliminate broken or too blurry images to ensure the visibility of the details. Finally, all images are distributed into multiple subsets, with 100 images in each.

\subsubsection{Question-Answer Pair Creation} \label{sect:qa-creation}

For this stage, we first employ a number of Vietnamese crowd workers with sufficient language proficiency. Taking advantage of crowdsourcing, we expect to have enough quantity and necessary variation to ensure the diversity of linguistic representation through the formation of ground truth questions and answers, as well as the broadness of the vocabulary.

As of major significance, guidelines are designed to monitor the quality of the dataset. The crowd workers are asked to conform to the following guideline in Table \ref{tab:guideline}. To ensure the open-ended characteristics of the questions and answers, we require the questions to be more extractive rather than to be in the form of verification like binary or selective questions. The answers are also expected to be longer than single words. Moreover, we decide to control the questioning on quantities, colors, and directions since these factors may take an important role in indicating and distinguishing among objects but can be mistaken easily due to linguistic variation and inconsistency during crowdsourcing. The examples to portray these criteria are shown in \ref{appendix-B}.

\begin{table}[ht]
\caption{Guidelines for the creation of questions and answers from the OpenViVQA dataset}\label{tab:guideline}
\begin{adjustbox}{width=\textwidth}
\begin{tabular}{@{}ll@{}}
\toprule
\multicolumn{1}{c}{\textbf{Criteria}} & \multicolumn{1}{c}{\textbf{Rules}} \\ \toprule
Number of QAs & Create at least 3 QAs for each image \\ \midrule
Answer complexity & Encouraged answers are in the form of phrases or sentences, not single words \\ \midrule
Question type & Must not be yes/no or selective questions \\ \midrule
Quantities & \begin{tabular}[c]{@{}l@{}} \tabitem Write quantities in alphabet characters rather than numeric characters.\\ \tabitem Quantities are not greater than 10.\end{tabular} \\ \midrule
Colors & \begin{tabular}[c]{@{}l@{}} \tabitem Only from this provided list: black, white, red, orange, yellow, green, blue, sky blue,\\ purple, pink, brown, and gray.\\ \tabitem Ignore color property in the QA if the above colors cannot exactly represent the true\\ color of the object.\end{tabular} \\ \midrule
\begin{tabular}[c]{@{}l@{}}Directions \\ (left and right)\end{tabular} & \begin{tabular}[c]{@{}l@{}}If following that direction words is an object, then such direction is defined based on that\\ object, else using the perspective of the viewer to define the direction.\end{tabular} \\ \bottomrule
\end{tabular}
\end{adjustbox}
\end{table}

The crowd-workers are required to use the prepared tool to create the questions and answers for each image in the assigned subsets. They are encouraged to make QAs for as many images as possible. In case the number of QAs for an image is below the minimum quantity specified in the guidelines since there are only a few details in that image, crowd-workers can form QAs with similar meanings to the existing ones. However, if the image is too vague and without any certain details, it can be skipped. The questions and answers creation stage lasts until all assigned subsets are completed.

\subsubsection{QA Type Classification} \label{sub:qa_type_classification}

To better conduct the analyses and experiments on the scene-text property of the OpenViVQA dataset, we decide to classify each of the QA in the dataset to be whether \emph{"Text QA"} or \emph{"Non-text QA"}. Specifically, Non-text QAs are QAs that focus on exploiting the information of the objects, including their attributes and relationship with others. For instance, any QA for objects or concepts indication, colors, quantities, or positions of objects is labeled as Non-text QA. On the other hand, we classify any QA as Text QA if it exploits information in the appeared scene text itself or utilizes the scene text to question other specific objects in a similar way as Non-text QA.

\begin{figure*}[ht]
    \centering
    \begin{subfigure}{0.35\textwidth}
        \includegraphics[width=\textwidth, height=1.03\textwidth]{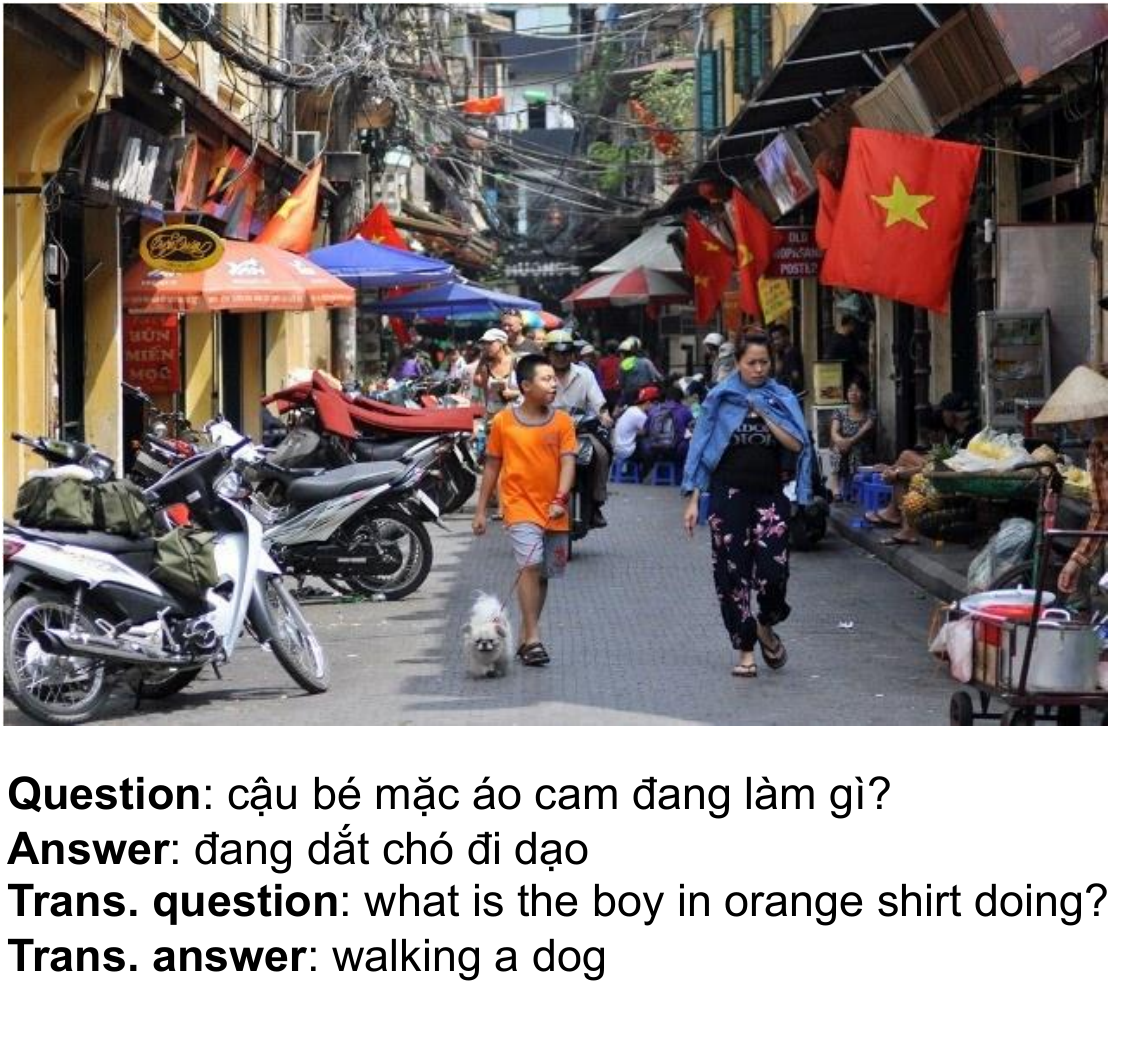}
        \caption{Non-text QA}
        \label{fig:non-text_qa}
    \end{subfigure}
    \begin{subfigure}{0.5\textwidth}
        \includegraphics[width=\textwidth]{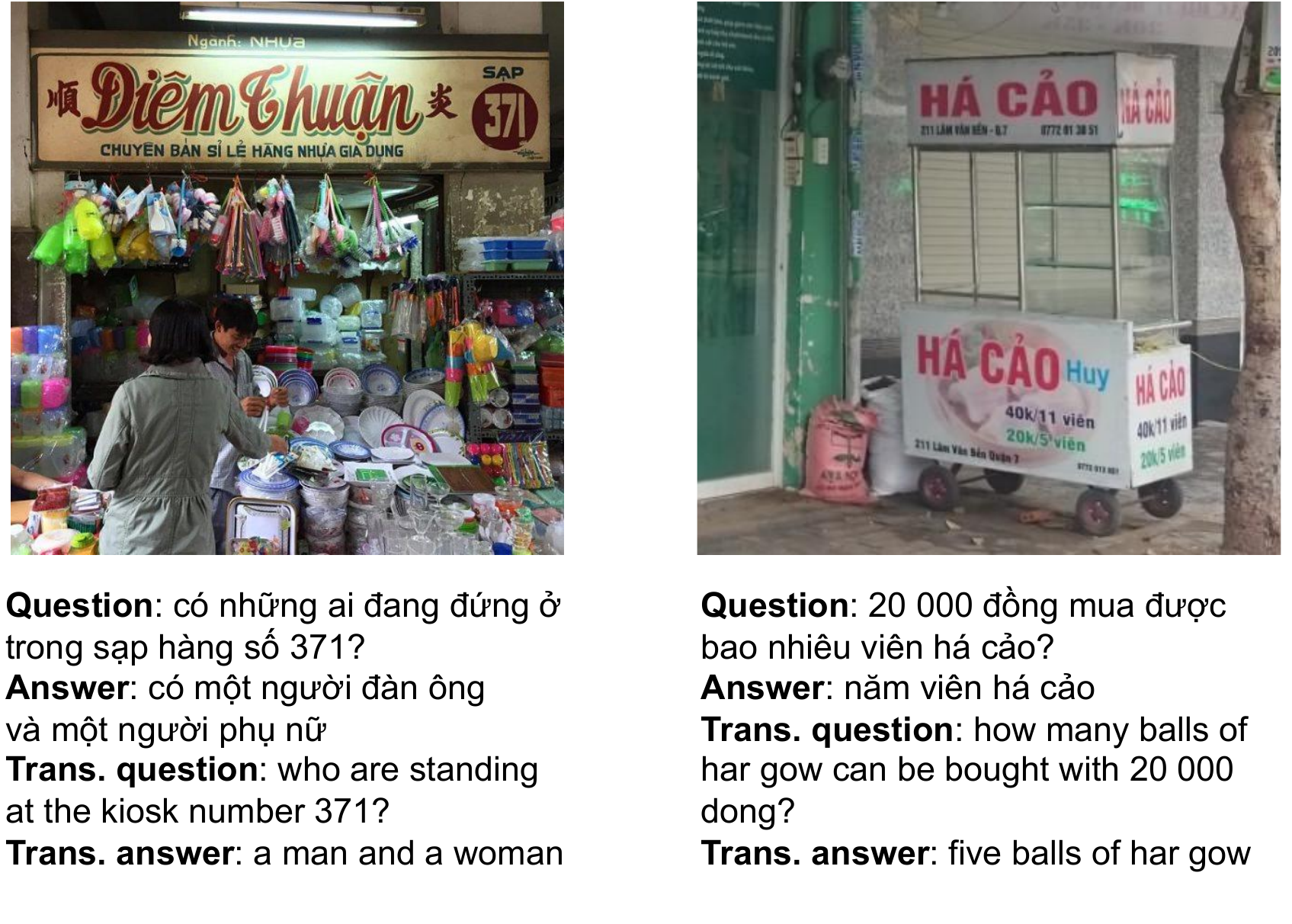}
        \caption{Text QAs}
        \label{fig:text_qa}
    \end{subfigure}
    \caption{Typical examples of Non-text QA and Text QA}
    \label{fig:text_nontext_examples}
\end{figure*}

We show some typical examples of Non-text QA and Text QA in Figure \ref{fig:text_nontext_examples}. For the leftmost image (Figure \ref{fig:non-text_qa}), a crowd worker only questioned the activity of a boy, hence this QA is a regular case of Non-text QA. On the other hand, the QAs in Figure \ref{fig:text_qa} have traces of using scene text from the images, which is performed in two cases. The first case has the question taking the scene text (the kiosk number) to help indicate explicitly the location of the questioned subjects. Meanwhile, the question in the second one directly focuses on the content of the appeared scene text from which the answer is expected to be extracted (the quantity of har gow balls with the corresponding price is shown on the signboard).

The QA-type classification stage involves active participation from a group of crowd-workers. Afterward, we proceed to pick out two subsets so as to meticulously monitor the classification agreement among 14 crowd-workers. This is done to quantify the mutual agreement in perceiving each QA as a Text QA or a Non-text QA. On all the QAs of these subsets, we use the traditional Percent Agreement score calculated as the number of QAs that all annotators treat as the same type divided by the total number of QAs. We also employ the Fleiss' Kappa \cite{fleiss1971measuring} to quantify the agreement regarding the consistency of classification and account for the agreement that may occur by chance. The kappa value is determined by:

\begin{equation}
    \kappa = \frac{\bar{P}-\bar{P_e}}{1-\bar{P_e}}
\end{equation}
where $1-\bar{P}$ is the degree of agreement attainable above chance while $\bar{P}-\bar{P_e}$ indicates the degree of agreement that is actually achieved above chance. The formula derives from the $P_i$ term, the extent to which annotators agree on i-th QA, and $p_j$ term, the proportion of all classifications as j-th QA type.
\begin{equation}
    P_i = \frac{1}{n(n-1)}\sum^{k}_{j=1}n_{ij}(n_{ij}-1)
\end{equation}
\begin{equation}
    p_j = \frac{1}{Nn}(\sum^{N}_{i=1}n_{ij})
\end{equation}
where $N$ is the number of QA (491 in our case), $n$ is the number of annotators (14 annotators), and $k$ is the number of classification categories (2 QA types). $n_{ij}$ is the number of annotators who assigned type j-th to the i-th QA. From here, we calculate $\bar{P} = \frac{1}{N}\sum^{N}_{i=1}P_i$ and $\bar{P_e}=\sum^{k}_{j=1}p_{j}^2$, hence the $\kappa$ value.

As a result, we obtain the Fleiss' Kappa of 0.8975 and the Percent Agreement score of 87.37\%. These results show a high level of agreement amongst the crowd-workers achieved above chance and a clear distinction between Text and Non-text QA.

\subsubsection{Dataset Validation}

To ensure the quality and consistency of the dataset, we take it through the validation process where thorough checks and refinements are executed, as shown as one of the steps in the pipeline in Figure \ref{fig:openvivqa-creation}. 

We first assigned several crowd workers with random subsets throughout the dataset and asked them to fix any spelling or syntax mistakes they could find. To facilitate the training process, we preprocess the QAs in the dataset by setting the text to lowercase, placing whitespaces between words and punctuation marks, and normalizing prices from the scene text. As for the prices which have essentiality for questioning in reality but are displayed in a wild writing variation, we automatically convert them to the form that looks like this: "\#,\#\#\# đồng" where \# is a numeral character, the comma is a widely used thousands separator in Vietnam, and "đồng" is the Vietnamese currency.

\subsection{Dataset Analysis}

\subsubsection{Initial Statistic}

\begin{table}[ht]
    \centering
    \begin{tabular}{lrrr}
\hline
      & \textbf{Images} & \textbf{Text} & \textbf{Non-text} \\ \hline
Train & 9,129           & 13,104        & 17,729            \\
Dev   & 1,070           & 1,733         & 1,772             \\
Test  & 1,000           & 1,766         & 1,770             \\
Total & 11,199          & 16,643        & 21,271            \\ \hline
\end{tabular}
    \caption{Statistic of images and QAs.}
    \label{tab:qa_image_statistic}
\end{table}

The OpenViVQA dataset consists of 11,199 images associated with 37,914 question-answer pairs in Text QA or Non-text QA. The detailed information of our dataset is listed in Table \ref{tab:qa_image_statistic}.

As stated in section \ref{sub:qa_type_classification}, QAs in the OpenViVQA dataset are Text QAs or Non-text QAs. These Text QA statistically has significant numbers, which challenge VQA models must have reading scene text ability to fully understand the content of images as well as find the correct information to give answers.

\begin{figure}[ht]
    \centering
    \begin{subfigure}{0.5\textwidth}
        \includegraphics[width=\textwidth]{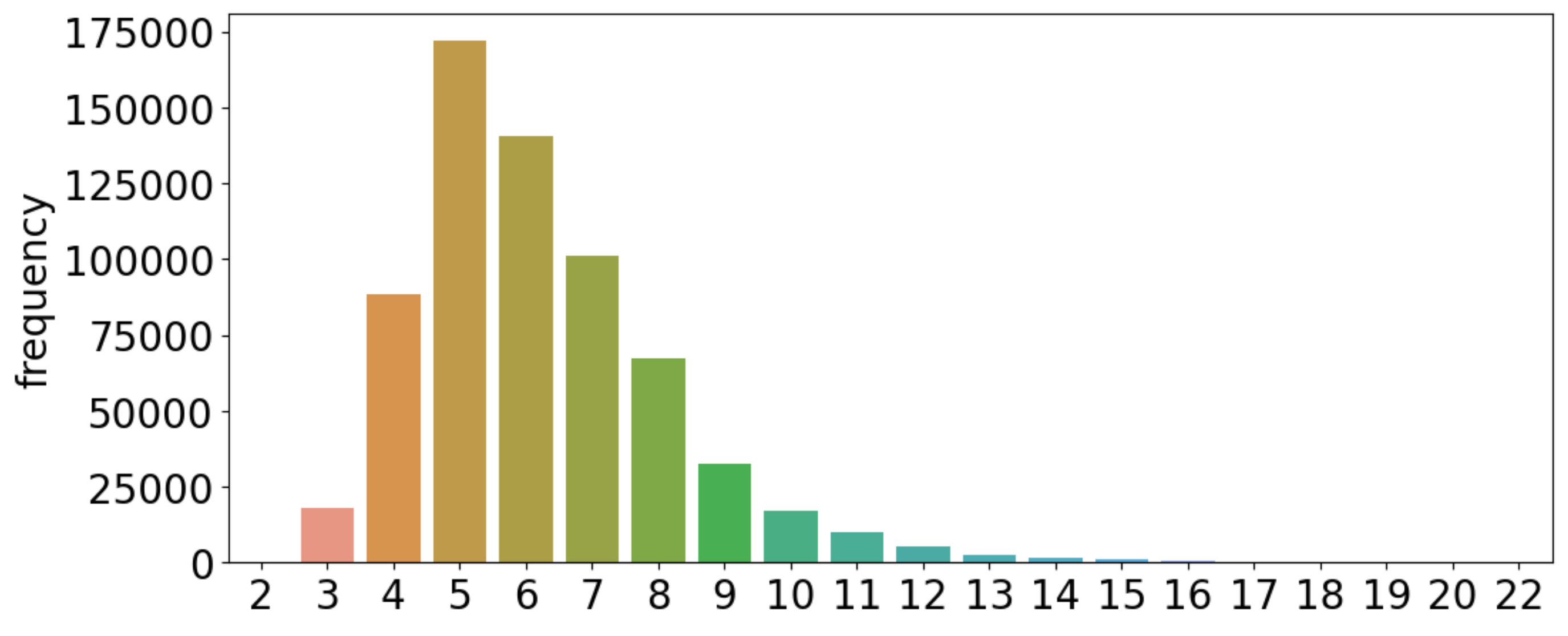}
        \caption{VQAv2}
    \end{subfigure}
    \begin{subfigure}{0.49\textwidth}
        \includegraphics[width=\textwidth]{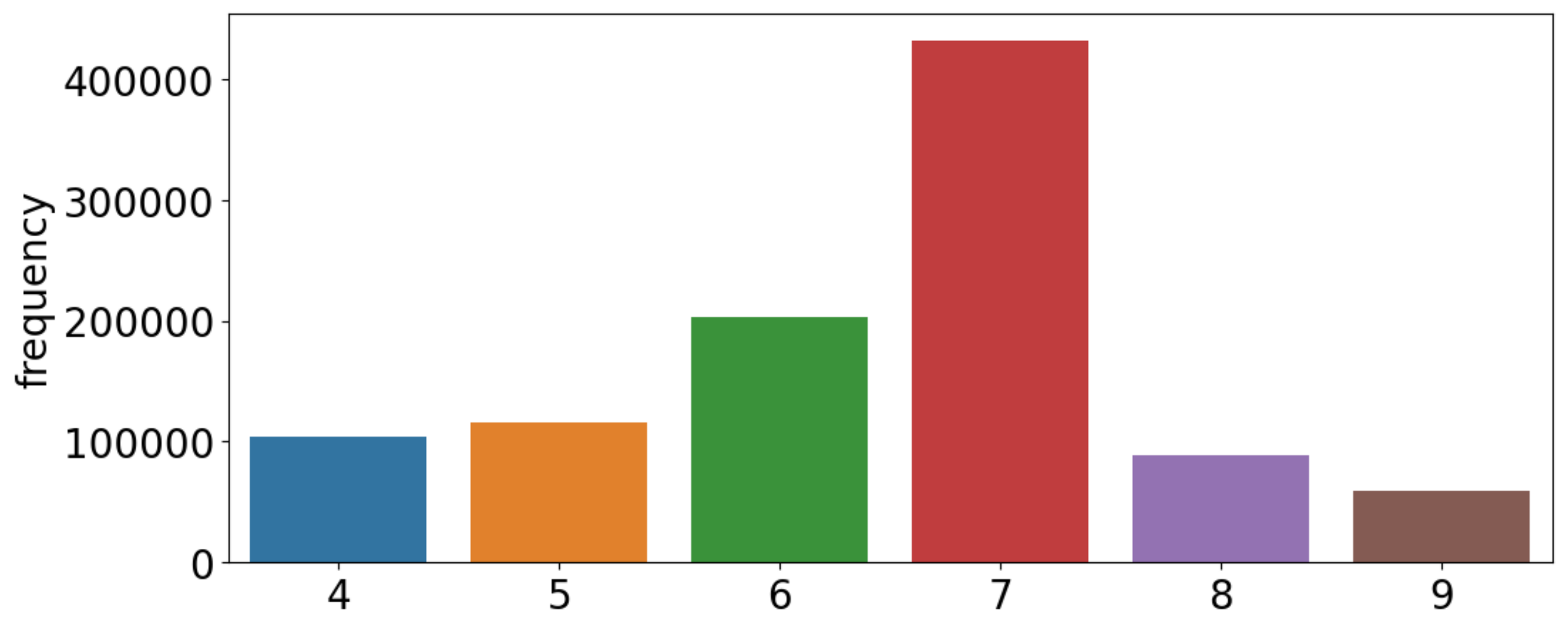}
        \caption{OCR-VQA}
    \end{subfigure}
    \begin{subfigure}{0.49\textwidth}

        \includegraphics[width=\textwidth]{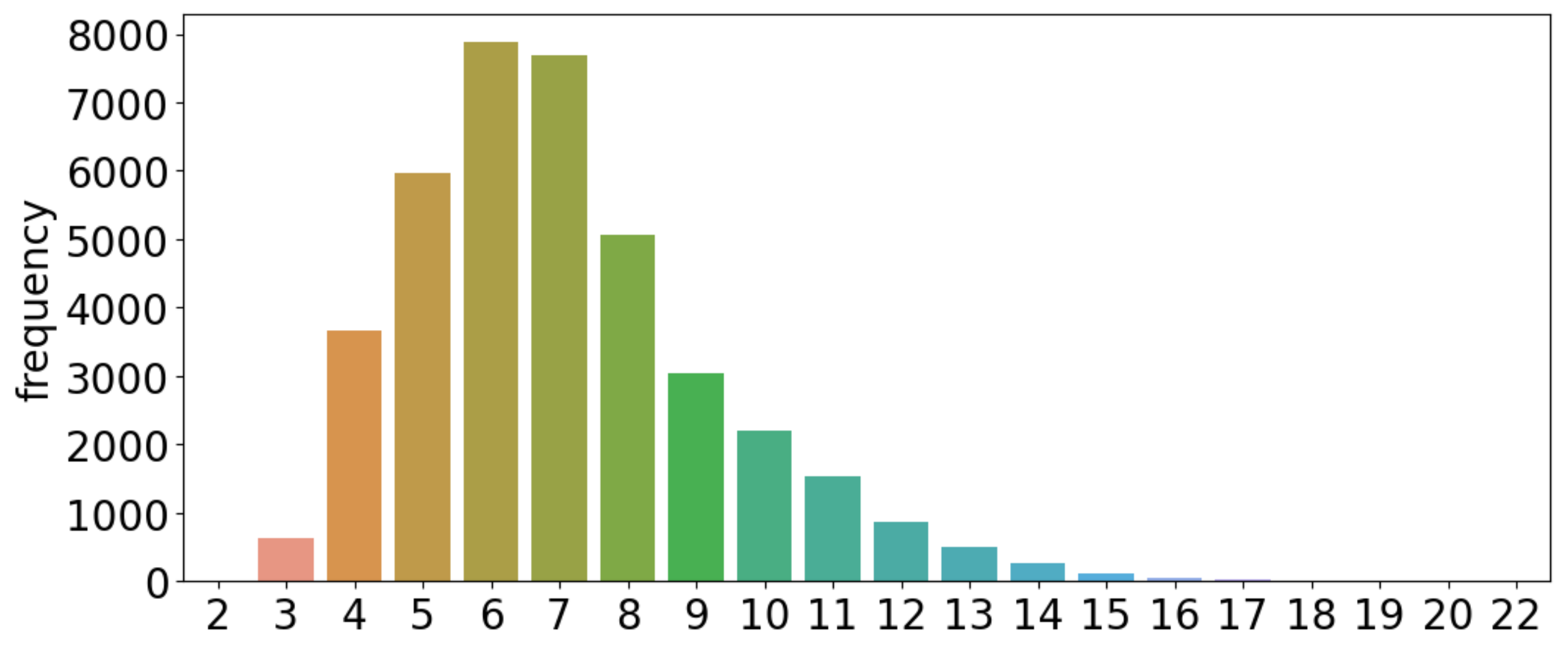}
        \caption{TextVQA}
    \end{subfigure}
    \begin{subfigure}{0.49\textwidth}
        \includegraphics[width=\textwidth]{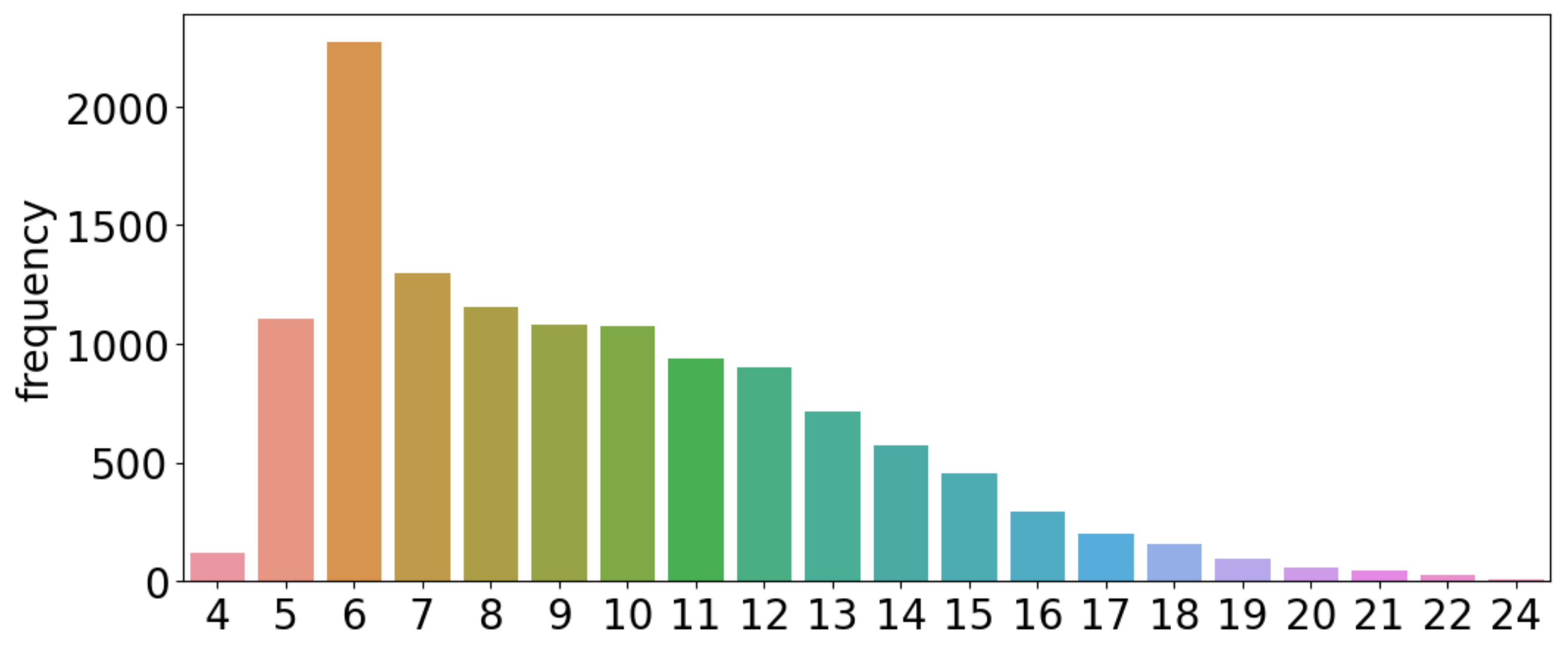}
        \caption{ViVQA}
    \end{subfigure}
    \begin{subfigure}{0.49\textwidth}
        \includegraphics[width=\textwidth]{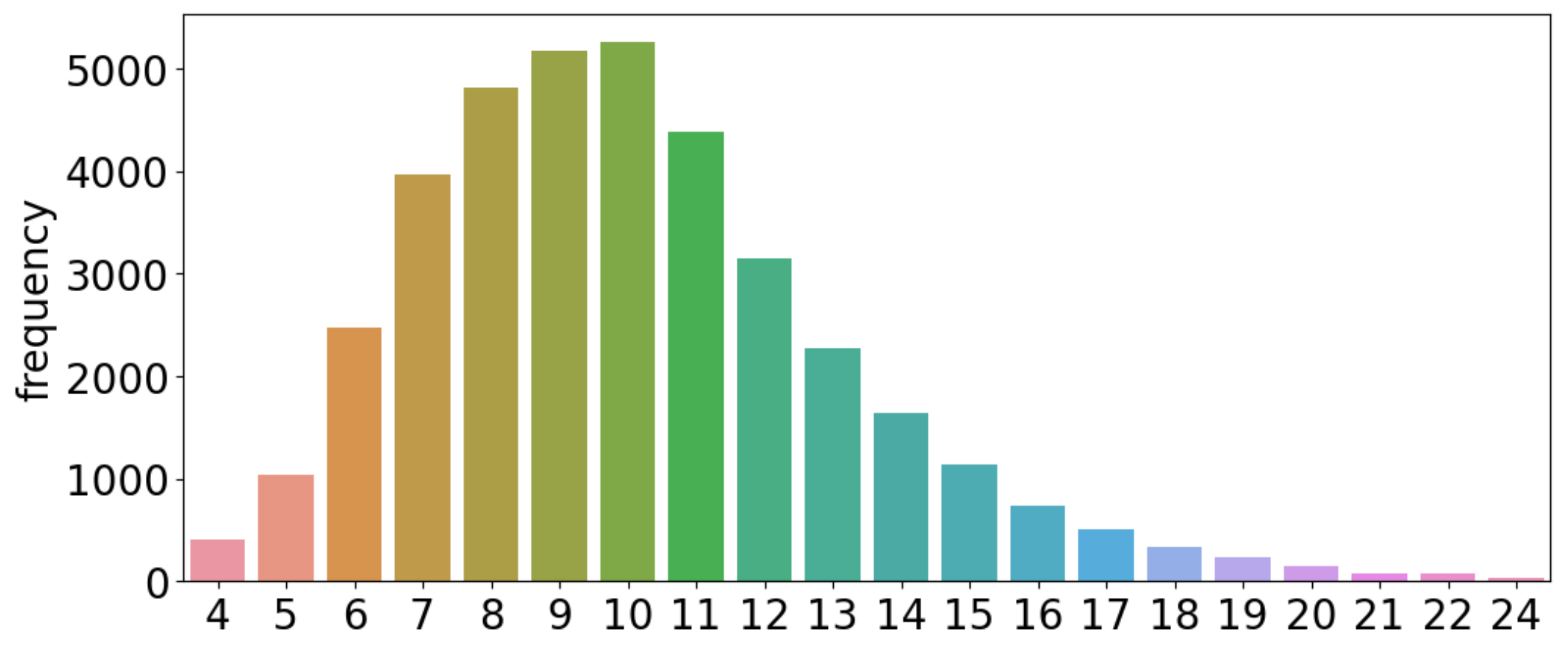}
        \caption{OpenViVQA}
    \end{subfigure}
    \caption{Comparison of question length among VQA datasets.}
    \label{fig:datasets-question-length-statistics}
\end{figure}

\begin{figure}[ht]
    \centering
    \begin{subfigure}{0.5\textwidth}
        \includegraphics[width=\textwidth]{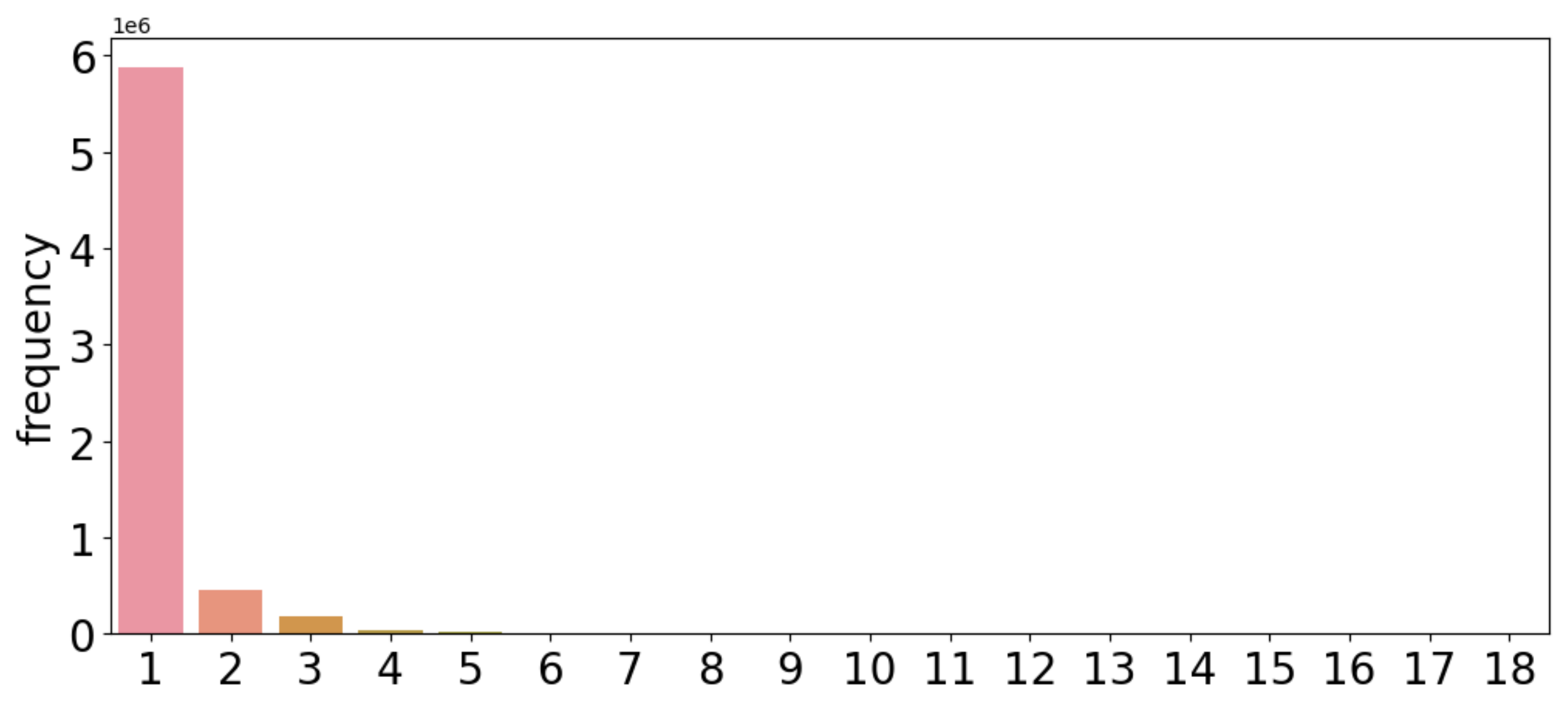}
        \caption{VQAv2}
    \end{subfigure}
    \begin{subfigure}{0.49\textwidth}
        \includegraphics[width=\textwidth]{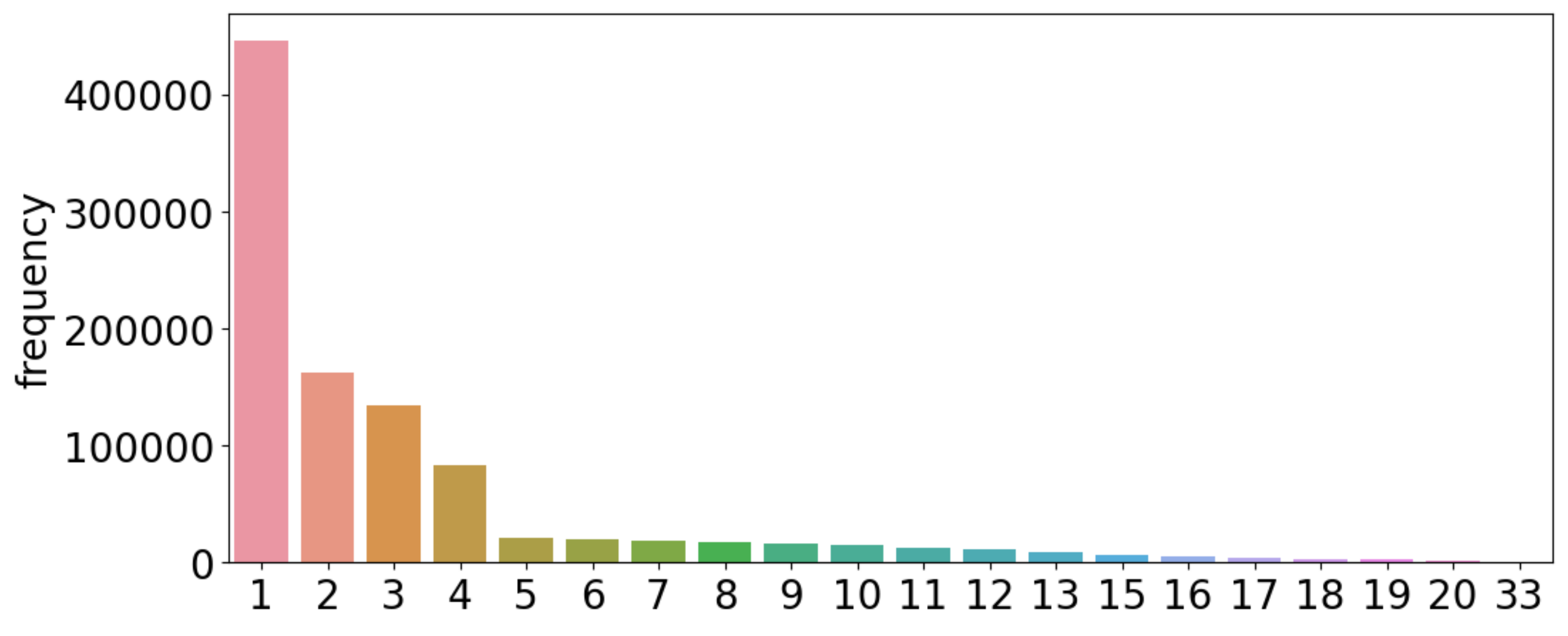}
        \caption{OCR-VQA}
    \end{subfigure}
    \begin{subfigure}{0.49\textwidth}
        \includegraphics[width=\textwidth]{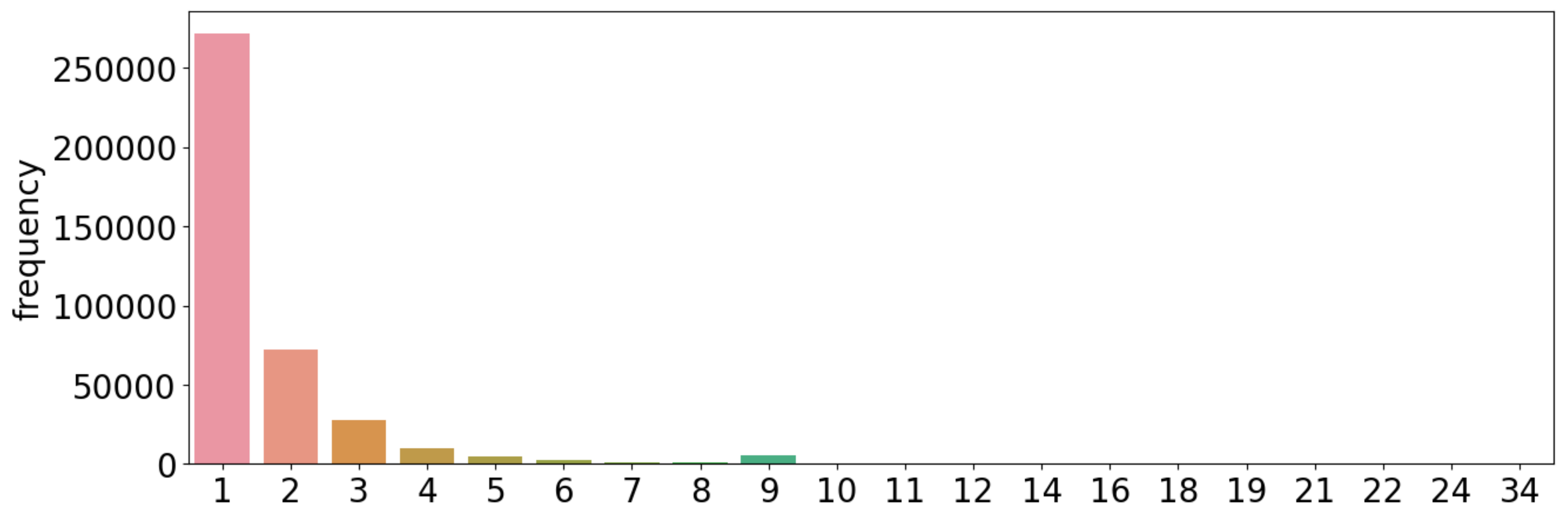}
        \caption{TextVQA}
    \end{subfigure}
    \begin{subfigure}{0.49\textwidth}
        \includegraphics[width=\textwidth]{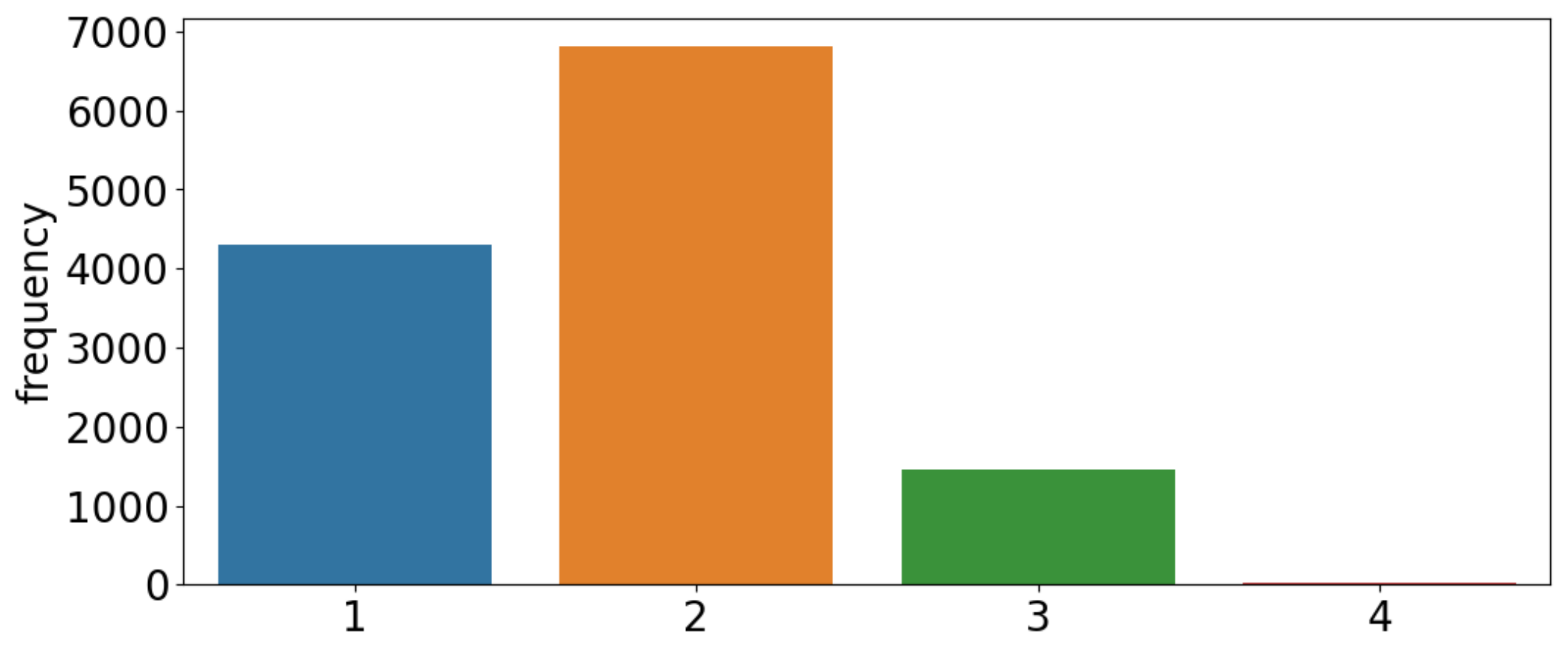}
        \caption{ViVQA}
    \end{subfigure}
    \begin{subfigure}{0.49\textwidth}
        \includegraphics[width=\textwidth]{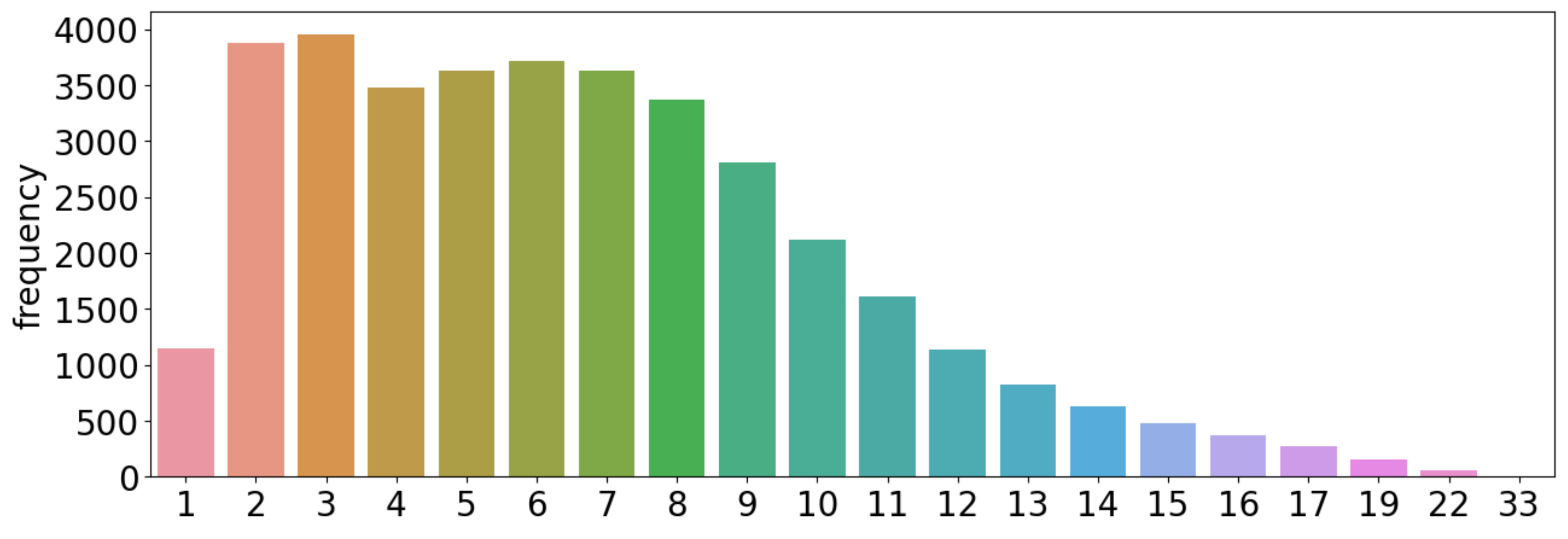}
        \caption{OpenViVQA}
    \end{subfigure}
    \caption{Comparison of answer length among VQA datasets.}
    \label{fig:datasets-answer-length-statistics}
\end{figure}

Moreover, as we define in Section \ref{sect:comparison}, the OpenViVQA introduces the open-ended VQA task, so its questions and answers are open-ended, and we assume such complicated answers challenge VQA models that obtained SOTA results on the VQAv1 \cite{VQA} and VQAv2\cite{goyal2017making} datasets. For more detail, we conducted statistics for the length of questions (Figure \ref{fig:datasets-question-length-statistics}) and answers (Figure \ref{fig:datasets-answer-length-statistics}) based on the number of tokens in each answer.

As shown in Figure \ref{fig:datasets-question-length-statistics}, questions in the OpenViVQA dataset has diverse length and their length distribution is smoother than those of other VQA datasets. Moreover, the distribution of answer length shows the diversity and complication of answers of the OpenViVQA dataset (Figure \ref{fig:datasets-answer-length-statistics}). Most of the answers have lengths that fall between 2 and 10, which indicates the classification approach of current VQA methods is not adaptable on the OpenViVQA dataset, and we should propose ones having answer generation ability to gain better results. Details of this conclusion will be shown in our experiments.

\subsubsection{Comparison with Other Visual Question Answering Datasets} \label{sect:comparison}

We conducted statistics to compare the OpenViVQA dataset with other similar VQA datasets in English and Vietnamese in both statistical aspects (such as distribution of question length and answer length or total images) and linguistic aspects. We define the linguistic aspects of a VQA dataset are the statistical numbers of questions, answers, semantic dependencies in questions or answers, and the height of the semantic tree constructed based on the semantic dependencies. 

\begin{figure}[ht]
    \centering
    \begin{subfigure}{0.24\textwidth}
        \includegraphics[width=\textwidth]{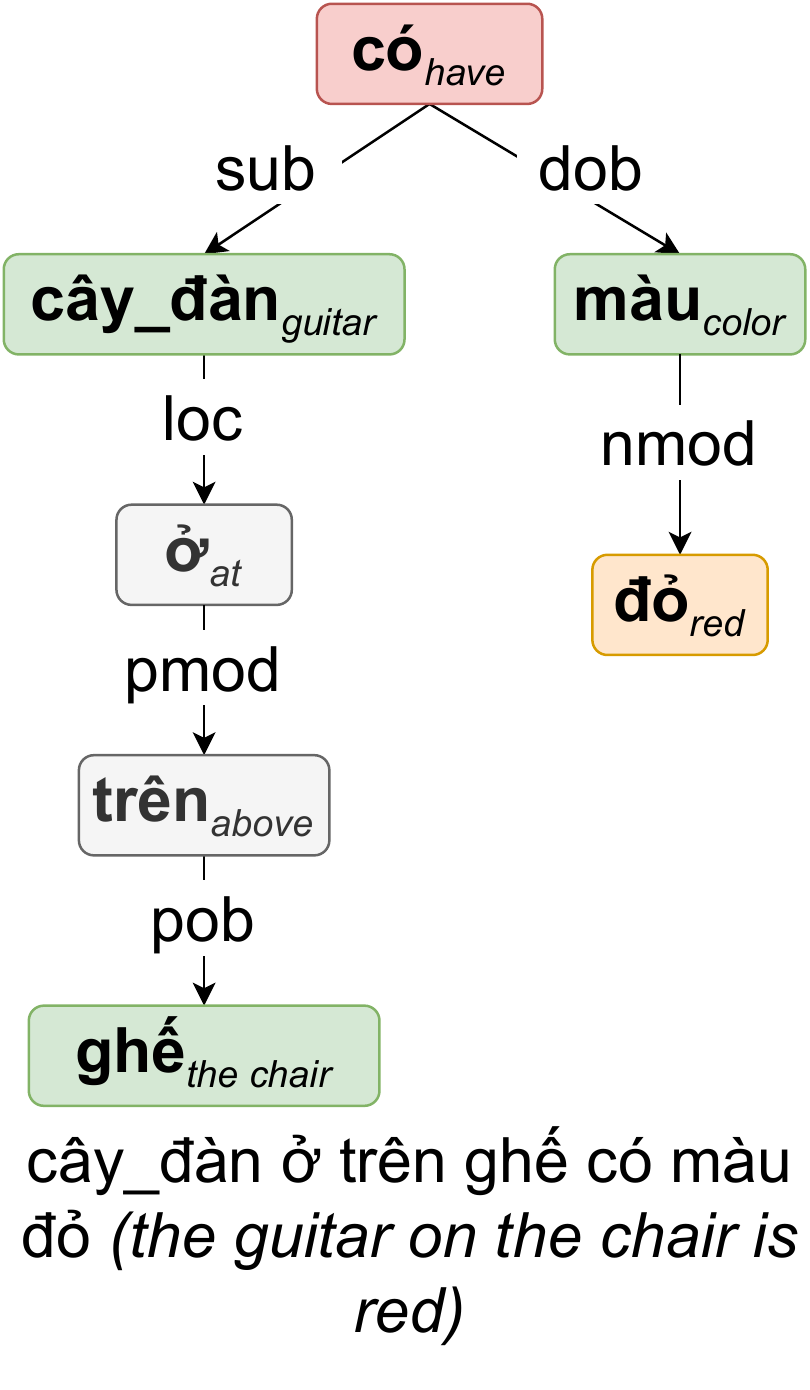}
    \end{subfigure}
    \begin{subfigure}{0.74\textwidth}
        \includegraphics[width=\textwidth]{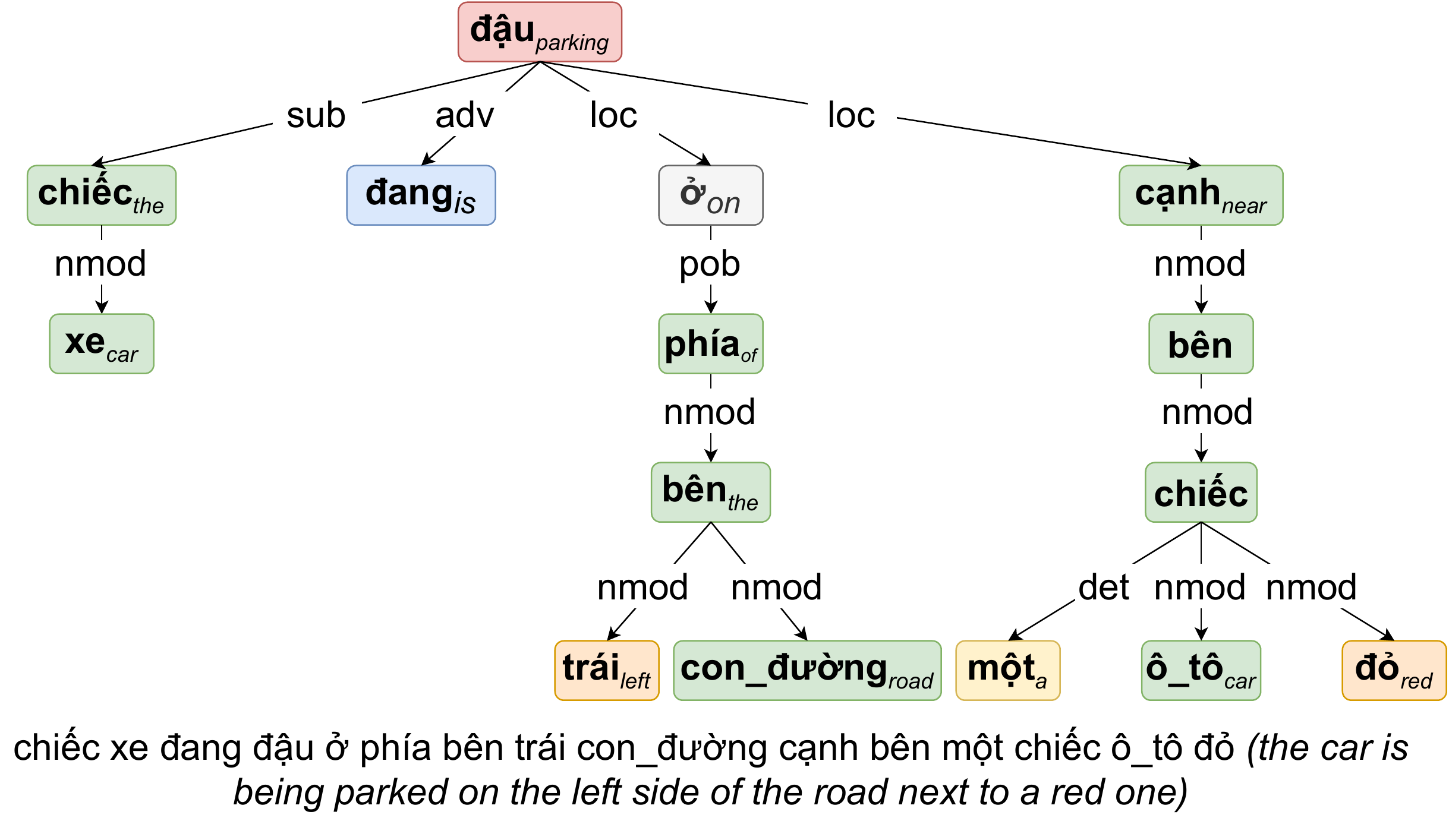}
    \end{subfigure}
    \caption{Trees of semantic dependencies between a simple sentence (left) and a complicated sentence (right). The simple sentence has 6 dependencies and its semantic tree has a height of 4 while the complicated one has 14 dependencies and its semantic tree has a height of 4.}
    \label{fig:dependencies}
\end{figure}

In order to determine the linguistic complexity of a given sentence, we introduced the Linguistic Complexity Specification (LCS) algorithm. First, LCS obtains the statistical number of dependencies between tokens in given sentences based on the results of the appropriate dependency parser for each language. Then using the dependency parsing results LCS constructs the semantic trees and specifies their height. The larger the total number of dependencies as well as the height of semantic trees, the more complicated the given sentence (Figure \ref{fig:dependencies}). We used LCS to point out how the complexity of answers in our dataset compared to other VQA datasets in English, accordingly emphasizing the linguistic difference between English and Vietnamese in the VQA task, hence the necessity of the novel dataset for researching VQA in Vietnamese particularly.

\begin{figure}[ht]
    \centering
    \begin{subfigure}{0.17\textwidth}
        \includegraphics[width=\textwidth]{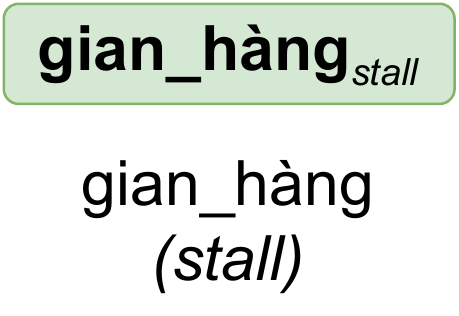}
        \caption{}
    \end{subfigure}
    \begin{subfigure}{0.33\textwidth}
        \includegraphics[width=\textwidth]{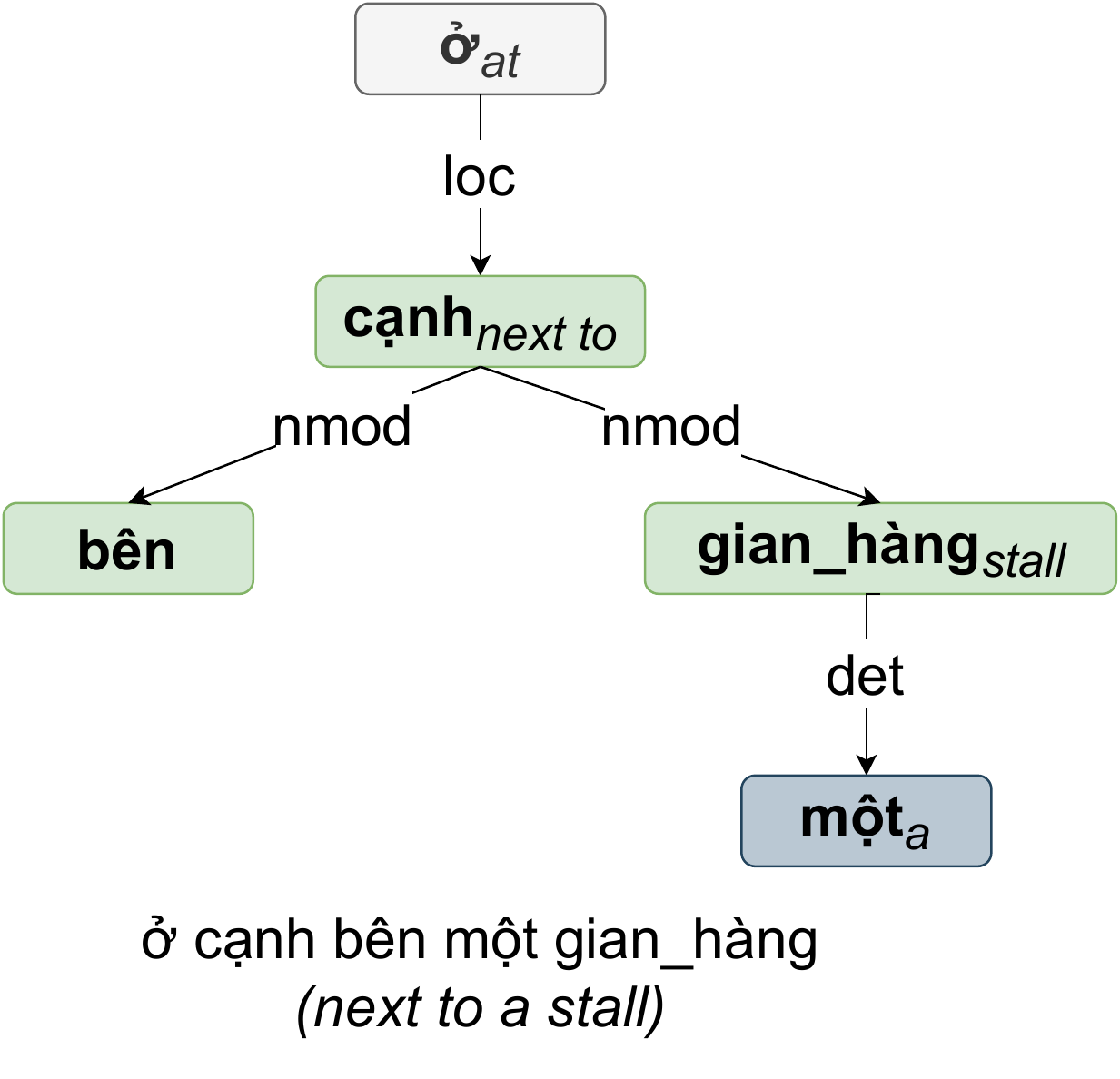}
        \caption{}
    \end{subfigure}
    \begin{subfigure}{0.47\textwidth}
        \includegraphics[width=\textwidth]{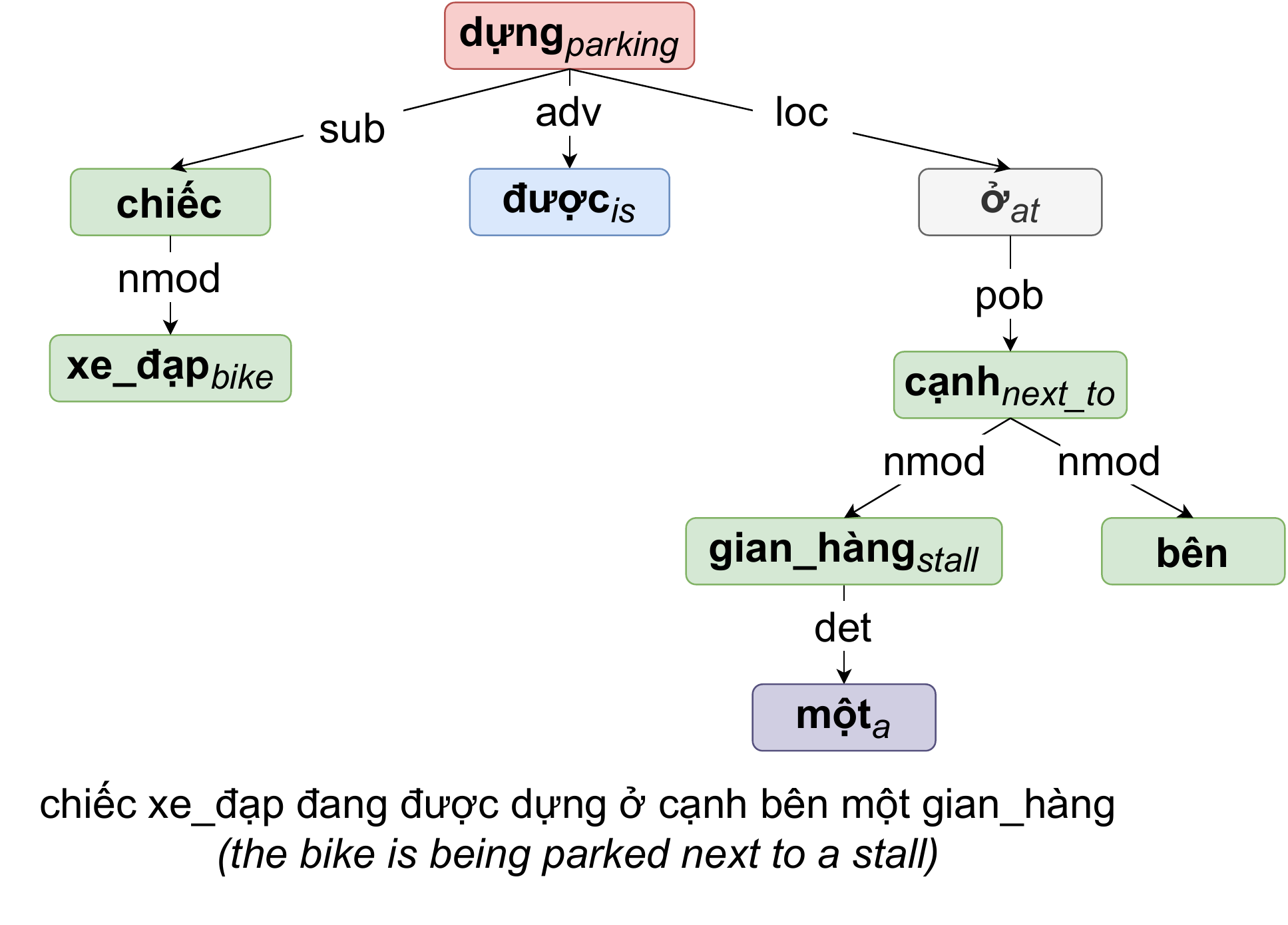}
        \caption{}
    \end{subfigure}
    \caption{Examples for the three linguistic levels of texts: (a) word (b) phrase and (c) sentence.}
    \label{fig:linguistic_level}
\end{figure}

In addition, depending on the dependency parser as well, we constructed an algorithm to specify whether a given text is a word, a phrase, or a sentence. In this paper, this algorithm is called the Linguistic Level Specification algorithm (LLS). The algorithm was constructed based on the assumption: texts that contain one token (word-segmented tokens for Vietnamese or space-split tokens for English) are words, texts that contain a root token as \textit{verb} and a token as \textit{sub} as the subject of that verb is considered sentences, and otherwise, they are phrases (Figure \ref{fig:linguistic_level}). We used LLS to show which linguistic level of sentence humans prefer to make while answering a given question in order that we emphasize the nature and open-ended features of answers in our dataset.

Before applying LCS or LLS algorithm to any VQA dataset, we preprocessed the questions and answers. For Vietnamese VQA datasets such as ViVQA \cite{tran-etal-2021-vivqa-vietnamese} or OpenViVQA, we first used VNCoreNLP \cite{vu-etal-2018-vncorenlp} to perform word segmentation for Vietnamese sentences. This is necessary because words in Vietnamese can have more than two syllables (such as "học sinh" (student) or "đại học" (university), and independently breaking a word into syllables can cause confusion and even misunderstanding in linguistic concept. For example, if we treat the 2-gram "học sinh" as the continuation of two separate words "học" and "sinh" this n-gram can be got as studying (học) biology (sinh). Consequently, the segmentation of words must be performed as the first step before any further analysis. For English VQA datasets, we obtained tokens simply by splitting sentences using space characters. After obtaining the token-preprocessed sentences (word-segmented sentences for Vietnamese or split sentences for English), we formed the semantic dependencies and their semantic trees using dependency parsing provided in PhoNLP \cite{phonlp} for Vietnamese sentences and SpaCy \cite{spacy2} for English sentences.

\begin{table}[ht]
        \resizebox{\textwidth}{!}{
            \begin{tabular}{clccccccccc}
\hline
\multicolumn{2}{c}{\multirow{2}{*}{\textbf{Dataset}}} & \multicolumn{3}{c}{\textbf{Word}}             & \multicolumn{3}{c}{\textbf{Dependency}}       & \multicolumn{3}{c}{\textbf{Height}}           \\ \cline{3-11} 
\multicolumn{2}{c}{}                                  & \textbf{min.} & \textbf{mean} & \textbf{max.} & \textbf{min.} & \textbf{mean} & \textbf{max.} & \textbf{min.} & \textbf{mean} & \textbf{max.} \\ \hline
\multirow{5}{*}{Question}         & VQAv2 \cite{goyal2017making}            & 2             & 6.2           & 23            & 2             & 6.3           & 26            & 1             & 3.3           & 14            \\
                                  & TextVQA \cite{singh2019towards}          & 2             & 7.1           & 33            & 2             & 7.5           & 39            & 1             & 3.9           & 21            \\
                                  & OCR-VQA \cite{mishraICDAR19}          & 4             & 6.5           & 9             & 4             & 6.5           & 10            & 2             & 3.6           & 6             \\
                                  & ViVQA \cite{tran-etal-2021-vivqa-vietnamese}            & 3             & 9.5           & 24            & 2             & 7.3           & 23            & 2             & 5.5           & 14            \\
                                  & OpenViVQA (ours)        & 3             & 10.1          & 32            & 2             & 7.8           & 27            & 2             & 5.2           & 16            \\ \hline
\multirow{5}{*}{Answer}           & VQAv2 \cite{goyal2017making}            & 1             & 1.2           & 18            & 0             & 2.8           & 44            & 1             & 1.0           & 11            \\
                                  & TextVQA \cite{singh2019towards}          & 1             & 1.6           & 85            & 0             & 1.5           & 103           & 1             & 1.3           & 40            \\
                                  & OCR-VQA \cite{mishraICDAR19}          & 1             & 3.3           & 74            & 0             & 2.8           & 100           & 1             & 1.8           & 38            \\
                                  & ViVQA \cite{tran-etal-2021-vivqa-vietnamese}            & 1             & 1.8           & 4             & 0             & 0.5           & 3             & 1             & 1.5           & 3             \\
                                  & OpenViVQA (ours)        & 1             & 6.9           & 56            & 0             & 4.8           & 52            & 1             & 4.0           & 22            \\ \hline
\end{tabular}}
    \caption{Linguistic comparison on questions and answers among VQA datasets. Note that these results were obtained on train-dev sets.}
    \label{tab:linguistic_comparison}
\end{table}

\begin{table}[ht]
    \centering
    
    \begin{tabular}{lrrr}
        \hline
        \multicolumn{1}{c}{\textbf{Dataset}} & \textbf{\#word} & \textbf{\#phrase} & \textbf{\#sentence} \\ \hline
        VQAv2 \cite{goyal2017making}                               &    5,884,207           &       651,128          &        45,775          \\
        OCR-VQA \cite{mishraICDAR19}                             & 3,287         & 302,497         & 15,010           \\
        TextVQA \cite{singh2019towards}                             & 28,317        & 35,964          & 4,947            \\ 
        ViVQA \cite{tran-etal-2021-vivqa-vietnamese}                               &    3,276           &     6,321            &   0               \\
        OpenViVQA (ours)                           & 1,067         & 21,022          & 12,289           \\ \hline
    \end{tabular}
    \caption{Linguistic level comparison among VQA datasets. Note that these results were obtained on train-dev sets.}
    \label{tab:level_comparison}
\end{table}

As shown in Table \ref{tab:linguistic_comparison}, the OpenViVQA has the most dependencies as well as the highest dependency trees in answers compared to other VQA datasets. Although the maximum number of dependencies of answers in the TextVQA and OCR-VQA datasets is more significant than those of the OpenViVQA dataset, the mean of dependencies of answers in OpenViVQA is significant and even the largest mean of dependencies. This indicates the complexity but nature of human answers, especially in Vietnamese, in the OpenViVQA dataset. As the results of our experiments, we will show such complicated answers challenge most SOTA VQA methods on English VQA datasets and that our approaches (e.g. tackle the OpenViVQA dataset by generating answers) are effective. Moreover, in the ViVQA dataset, the amount of semantic dependencies is small, which means the answers in this dataset are simple. According to these statistical numbers of ViVQA and OpenViVQA, we prove the ineffectiveness of the S dataset construction method proposed in \cite{tran-etal-2021-vivqa-vietnamese} and we recommend constructing the benchmark dataset manually rather than using the S method for the assurance of qualification.

\begin{table}[ht]
    \centering
    \resizebox{\textwidth}{!}{
    \begin{tabular}{llccccrrr}
\hline
\textbf{Dataset}        & \textbf{Images Source}          & \textbf{Method} & \textbf{Text QA} & \textbf{Non-text QA} & \textbf{Open-ended} & \multicolumn{1}{c}{\textbf{Images}} & \multicolumn{1}{c}{\textbf{Questions}} & \multicolumn{1}{c}{\textbf{Answers}} \\ \hline
$\text{VQAv2}_{en}$ \cite{goyal2017making}     & MS COCO \cite{lin2014microsoft}                        & H              & \xmark           & \cmark               & \xmark              & 204,721                             & 1,105,904                              & 11,059,040                           \\
$\text{OCR-VQA}_{en}$ \cite{mishraICDAR19}  & From the study \cite{iwana2016judging}          & S               & \cmark           & \xmark               & \cmark              & 207,572                             & 1,002,146                              & 1,002,146                            \\
$\text{TextVQA}_{en}$ \cite{singh2019towards}  & OpenImages \cite{OpenImages}                     & H              & \cmark           & \xmark               & \cmark              & 28,408                              & 45,336                                 & 453,360                              \\
$\text{DocVQA}_{en}$ \cite{Mathew_2021_WACV}   & UCSF Industry Documents Library & H              & \cmark           & \xmark               & \cmark              & 12,767                              & 50,000                                 & 50,000                               \\
$\text{VisualMRC}_{en}$ \cite{visualmrc} & Web pages                       & H              & \cmark           & \xmark               & \cmark              & 10,197                              & 30,562                                 & 30,562                               \\
$\text{OpenCQA}_{en}$ \cite{Kantharaj2022OpenCQAOQ}  & Web pages                       & H              & \cmark           & \xmark               & \cmark              & \multicolumn{1}{c}{7,724}           & 7,724                                  & 7,724                                \\ \hline
$\text{ViVQA}_{vi}$ \cite{tran-etal-2021-vivqa-vietnamese}    & VQAv1 \cite{VQA}                          & S           & \xmark           & \cmark               & \xmark              & 15,000                              & 12,598                                 & 12,598                               \\
$\text{OpenViVQA}_{vi}$ & Google search engine            & H              & \cmark           & \cmark               & \cmark              & 11,199                              & 37,914                                 & 37,914                               \\ \hline
\end{tabular}}
    \caption{Images-questions-answers comparison among VQA datasets. The subscript following the name of each dataset indicates its language (H stands for Human-annotated method and S stands for Semi-automatic method).}
    \label{tab:iqa_comparison}
\end{table}

In addition, as we mentioned previously, QAs in the OpenViVQA dataset are classified into two categories: Text QA and Non-text QA, based on their relevance to scene text in the images. A VQA dataset can then have both Text QA and Non-text QA (Table \ref{tab:iqa_comparison}). Although the OpenViVQA has fewer image-question-answer triplets than other English VQA datasets (Table \ref{tab:iqa_comparison}), it has both types of QAs as well as is the largest Vietnamese VQA dataset. Moreover, as indicated in Table \ref{tab:level_comparison}, approximately 36\% of answers in the OpenViVQA dataset are sentences, while in other similar VQA datasets such as TextVQA, 7\% of answers are sentences, and for OCR-VQA 4\% of answers are sentences. Compared to VQA datasets such as VQAv2, most of the answers are words (89.41\%), while phrases are 9.89\% and sentences are extremely small (0.7\%). For the ViVQA dataset, phrases are twice as words and no sentences in answers.

\section{Our Proposed Methods}
\label{sect:method}


As analyzed in Section 3.2, the OpenViVQA dataset contains open-ended answers with a diverse range of lengths (Figure \ref{fig:datasets-answer-length-statistics}), we believe these open-ended answers are challenged and can not be tackled using the classification approach as SAAA \cite{kazemi2017show}, MCAN \cite{yu2019deep} and LoRRA \cite{singh2019towards} were designed. To prove this statement we propose three answer-generation methods 
which not only keeps the spirit of the respective classifier-based method but also has the ability to give answers as humans. In the following sections, we mention the M4C method and our three proposed methods as generator-based methods for ease of calling. A detailed description of our proposed architectures is given below.

\subsection{Fusing by Staking Together}

\begin{figure}[ht]
    \centering
    \includegraphics[width=0.75\textwidth]{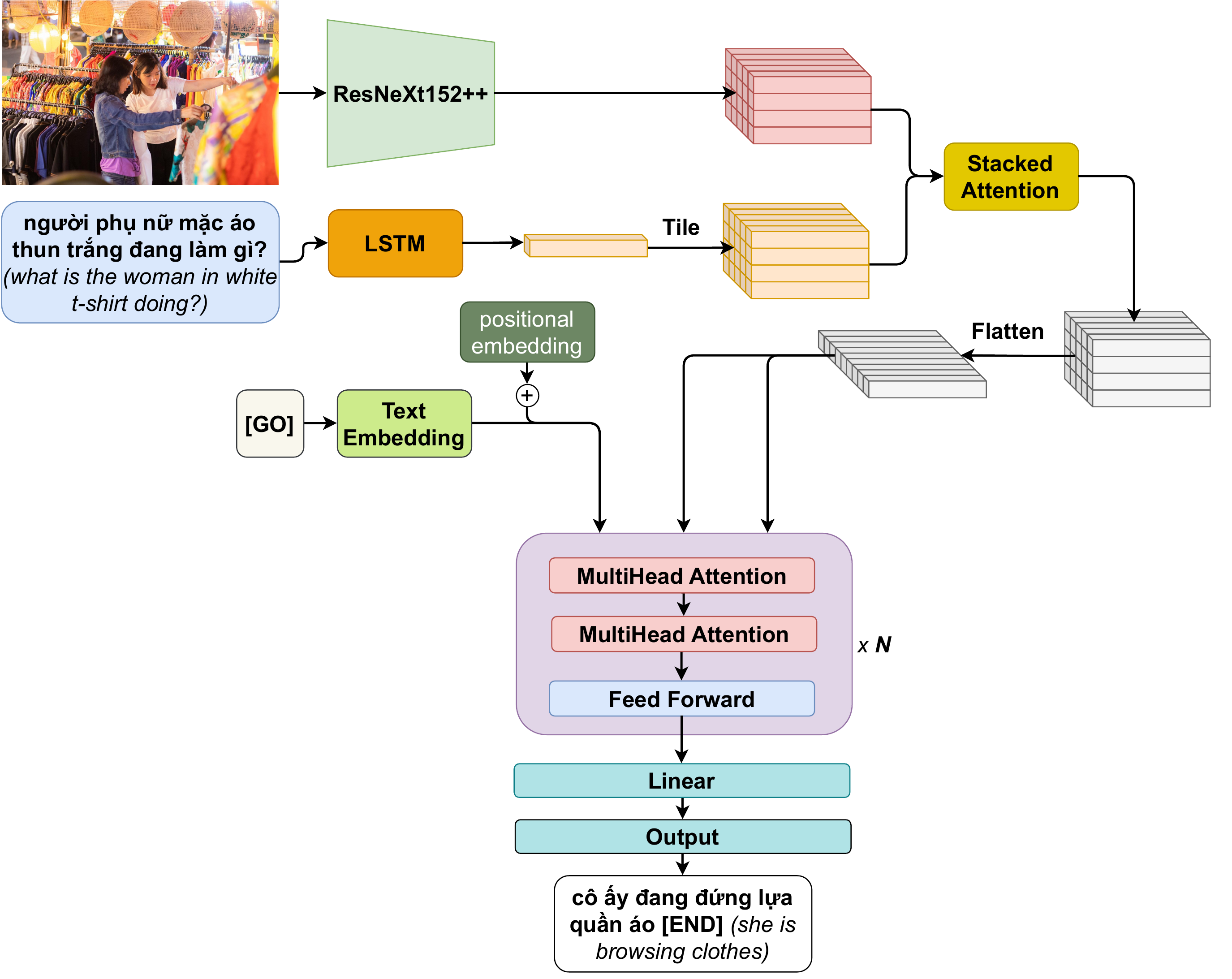}
    \caption{Fusing by Staking Together (FST) Method.}
    \label{fig:FST}
\end{figure}

Inspired by SAAA, we designed a novel method, Fusing by Stacking Together (FST), which uses the Stacked Attention \cite{yang2016stacked} to fuse information of images and questions, then form the answers by iteratively selecting tokens over a defined vocab. In general, the FST consists of four components: the Image Embedding module, the Question Embedding module, the MultiModal Fusion module, and the Answer Generator module (Figure \ref{fig:FST}).

The Image embedding of FST uses ResNeXt152++ \cite{jiang2020defense} to extract features (grid features in particular) from images. The Question Embedding consists of an LSTM \cite{hochreiter1997long} network to extract features from questions. Let $x_I \in \mathbb{R}^{s \times d_I}$ and $x_Q \in R^{d_Q}$, where $s$ is the total of spatial locations in images, be information extracted from the Image Embedding module and the Question Embedding module, respectively.

The MultiModal Fusion module consists of a Stacked Attention module similar to \cite{kazemi2017show}. Particularly, $x_Q$ is repeated to have the shape of $\mathbb{R}^{s \times d_Q}$. The attention weight is then obtained by the following formula:

\begin{align}
    a = softmax(W_x (W_I x_I^{T} + W_Q x_Q^{T}))^{T} \in \mathbb{R}^{s \times D}
\end{align}
where $W_x \in \mathbb{R}^{D \times D}$, $W_I \in \mathbb{R}^{D \times d_I}$ and $W_Q \in \mathbb{R}^{D \times d_Q}$, biases are reduced for clarification. The attended vector $a$ can be seen as stacked attention vectors that will be applied one by one on image features $x_I$. Then we obtained the fused features $x_f$ as follows:

\begin{align}
    x_f = \sum_{d \in \{1 \ldots D\}} (a_d \otimes x_I)
\end{align}
where $\otimes$ indicates the broad-cast multiplication.

In the Answer Generator module, answers are generated by conditioning on the fused features $x_f$ which are expressed as follows:

\begin{align}
    o_t = f(o_0, o_1, ..., o_{t-1} \mid x_f)
\end{align}
where $o_t$ is the output token at step $t$. To model this function $f$, we used the decoder of transformer architecture \cite{vaswani2017attention} with its masking technique as the Answer Generator module.

\subsection{Question-guided MultiModal Learning and Answer Generation}

\begin{figure}[ht]
    \centering
    \includegraphics[width=0.8\textwidth]{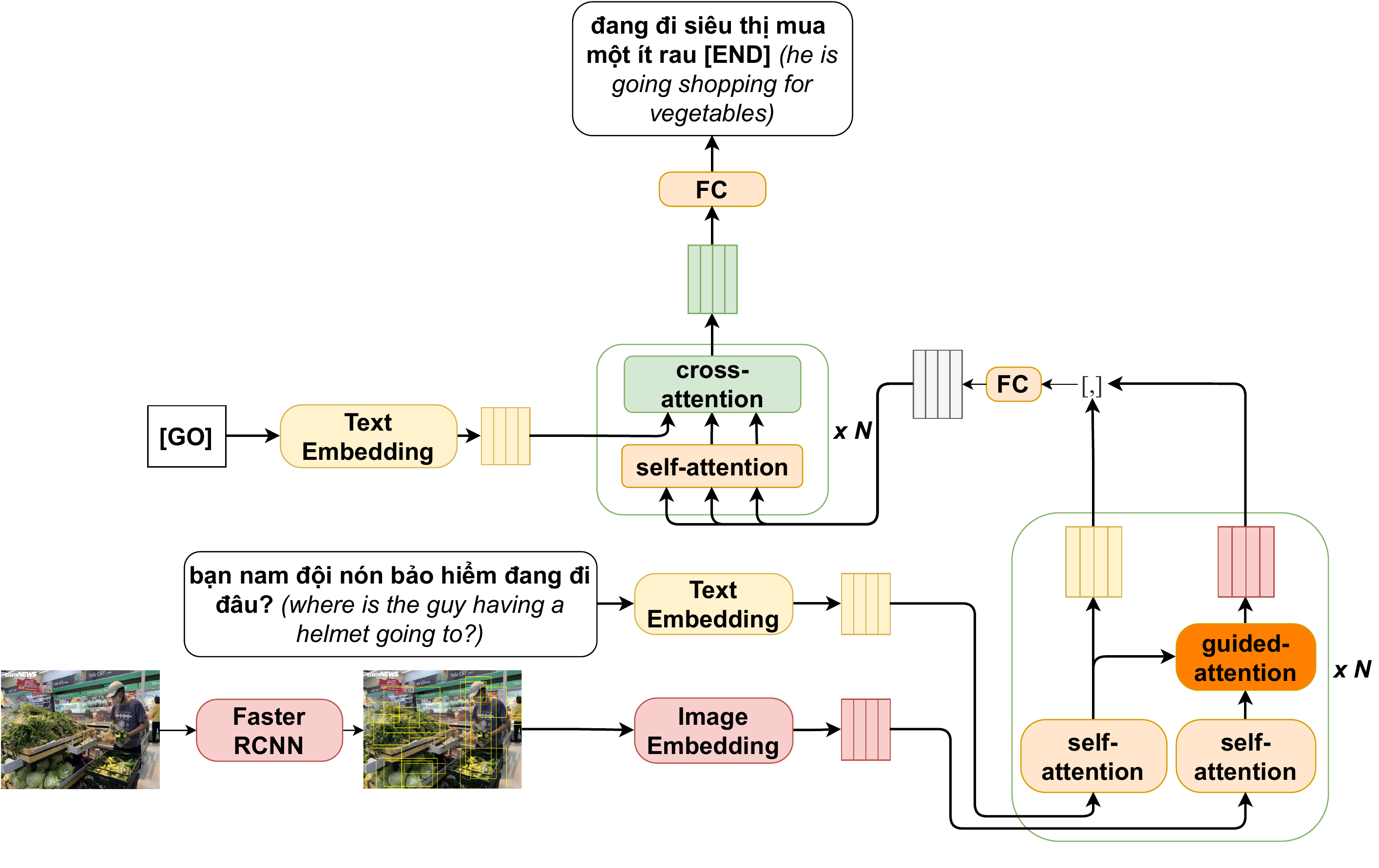}
    \caption{Question-guided MultiModal Learning and Answer Generation (QuMLAG) Method.}
    \label{fig:QuMLAG}
\end{figure}

We assume that the features of images should be transformed into features that keep enough visual information to answer the given questions. Then, by fusing these features with the features of questions, we obtain the fused information to conduct the answers. To this end, the Question-guided MultiModal Learning and Answer Generation (QuMLAG) was designed inspired by the Guided-attention (GA) mechanism proposed in \cite{yu2019deep}. QuMLAG consists of four components: the Image Embedding module, the Text embedding module, the MultiModal Fusion module, and the Answer Generator module (Figure \ref{fig:QuMLAG}).

The Image Embedding module receives the image features (region feature \cite{jiang2020defense}) extracted from FasterRCNN, then it passes these features through a fully connected layer to project their dimension to hidden dimension $dim$ hence producing the image features $x_I \in \mathbb{R}^{s \times dim}$ where $s$ is the total region of objects in images. The Question Embedding contains an LSTM network \cite{hochreiter1997long}. Questions firstly are embedded into high-dimensional embedded vectors by FastText \cite{bojanowski2017enriching}, then by applying the LSTM network, we obtained the linguistic features of questions $x_Q \in \mathbb{R}^{l \times dim}$ where $l$ is the total token of questions.

The MultiModal Fusion module of QuMLAG is designed based on the GA mechanism. In particular, this module consists of three multi-head attention modules. The first multi-head attention module is used to project image features extracted from FasterRCNN into the dim-dimension latent space of information, or in other words, this module is used to refine the $x_I$ features. The second multi-head attention is used with the same role as the first one but for the questions features $x_Q$. The final multi-head attention module is used to perform the GA attention mechanism, in which $x_I$ plays the role of query and $x_Q$ plays the role of key and value. The features vector $x_I$ after applying the Question-guided Attention using the third multi-head attention module eliminates the image features that do not assist in answering the given questions. Then output features $x_I$ and $x_Q$ are concatenated to yield fused features $x_f = [x_I, x_Q] \in \mathbb{R}^{(s+l) \times dim}$. 

The fused features vector $x_f$ is then fed to the Answer Generator module. Same as FST, the Answer Generator module is designed to implement the following formula:

\begin{align}
    o_t = f(o_0, o_1, ..., o_{t-1} \mid x_f)
\end{align}
we used the decoder module of transformer architecture \cite{vaswani2017attention} to implement the function $f$ as well as FST.

\subsection{MultiModal Learning and Pointer-augmented Answer Generator}

\begin{figure}[ht]
    \centering
    \includegraphics[width=0.6\textwidth]{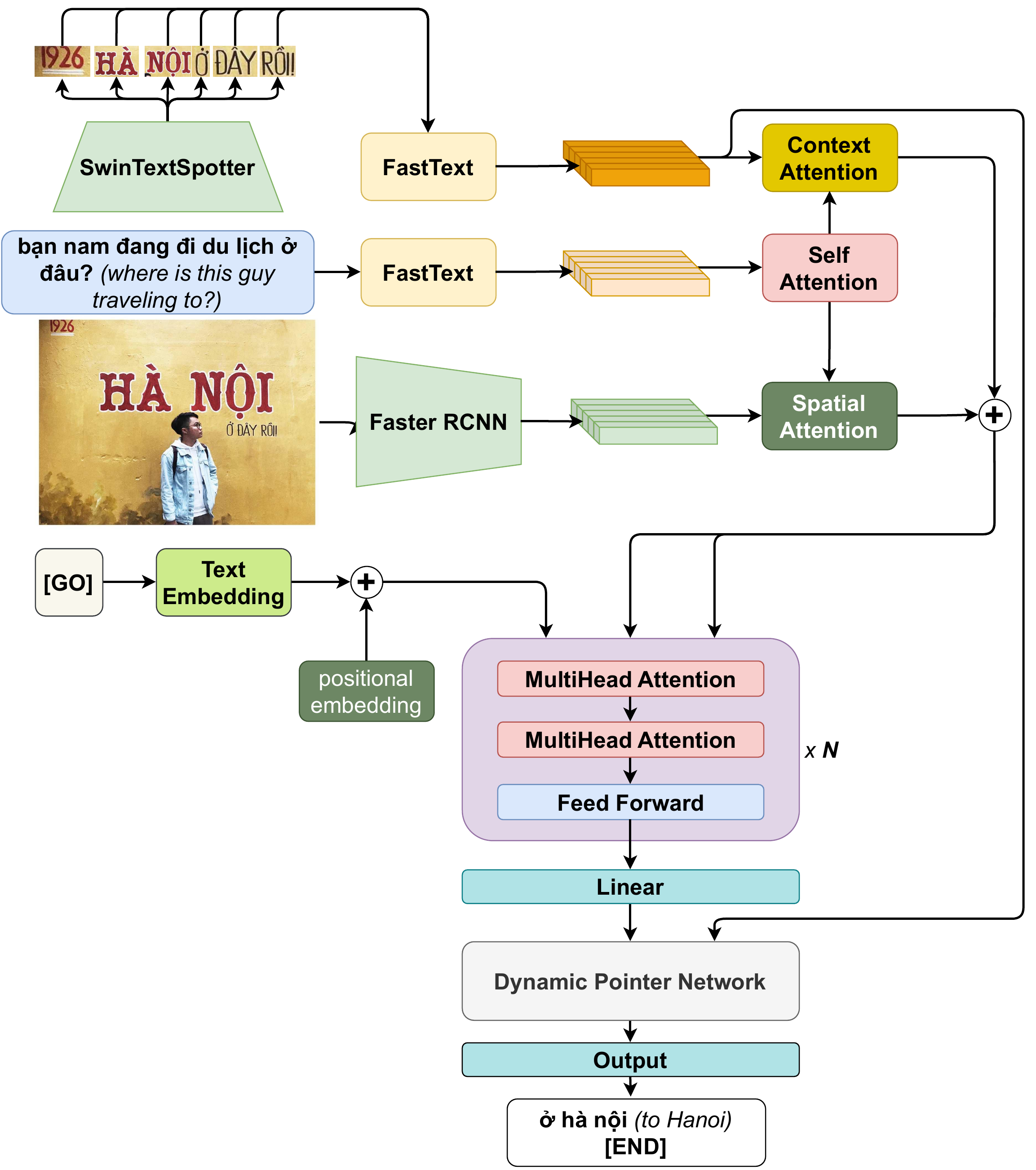}
    \caption{MultiModal Learning and Pointer-augmented Answer Generator (MLPAG) Method.}
    \label{fig:MLPAG}
\end{figure}

We aim to design a novel method that is inspired by the spirit of LoRRA but has the ability to dynamically select tokens from a defined vocab or scene texts from images while iteratively constructing answers. This method, the MultiModal Learning and Pointer-augmented Answer Generator (MLPAG) was designed using the same MultiModal Fusion mechanism as LoRRA \cite{singh2019towards}. In addition, as LoRRA can be able to read scene texts and use them in its answers, MLPAG was designed carefully to keep the spirit of LoRRA. In particular, MLPAG has five components: the Image Embedding module, the Scene Text Embedding module, the Question Embedding module, the MultiModal Fusion module, and the Answer Generator module (Figure \ref{fig:MLPAG}).

The Image Embedding module consists of a fully connected layer that projects image features (or region features \cite{jiang2020defense}) extracted from FasterRCNN \cite{ren2015faster} to hidden dimension $dim$ then yields the image features vector $x_I \in \mathbb{R}^{s \times dim}$ where $s$ is the total number of regions in images. The Question Embedding module contains a token embedding module that uses FastText \cite{bojanowski2017enriching}, and a fully connected layer to project embedded question features into embedded features $x_Q \in \mathbb{R}^{l \times dim}$ where $l$ is the total number of tokens in questions. The Scene Text embedding module contains an embedding module that uses FastText \cite{bojanowski2017enriching} to embed detected scene texts from images, and a fully connected layer to projects features vectors of FastText embedding to hidden dimension $dim$, then results in the features vector $x_S \in \mathbb{R}^{n \times dim}$ where $n$ is the total number of detected scene texts.

The MultiModal Fusion module of MLPAG contains three main components: Context Attention module, Spatial Attention module, and Self Attention module. The Self Attention module is a multi-head attention \cite{vaswani2017attention} module that is used to perform self-attention of $x_Q$ over itself, resulting in attended features $a_Q \in \mathbb{R}^{l \times dim}$. The Context Attention module is a multi-head attention module \cite{vaswani2017attention} which is used to perform the cross-attention of $x_S$ over $a_Q$. The Spatial Attention module is a multi-head attention module that is used to perform cross-attention of $x_I$ over $a_Q$. The two vectors $x_I$ and $x_S$ after being passed through the Spatial Attention and Context Attention, respectively, eliminate the visual information that is irrelevant to questions. Then the fusion information features are determined by

\begin{align}
    x_f = x_S \oplus x_I
\end{align} 
where $\oplus$ is the element-wise sum operator.

The Answer Generator module is then designed to generate answers iteratively. We use the transformer decoder \cite{vaswani2017attention} to model the following function:

\begin{align}
    o_t = f(o_0, o_1, ..., o_{t-1} \mid x_f)
\end{align}

Moreover, to provide MLPAG the ability to copy scene text from images to answers, we provide Answer Generator Module the Dynamic Pointer Network \cite{hu2020iterative}. In particular, let $h \in \mathbb{R}^{l \times dim}$ be the features vector prepared for producing output tokens $o \in \mathbb{R}^{l}$, the Dynamic Pointer Network takes into account $h$ and $x_S$ as follows:

\begin{align}
    ocr = (W_h h + b_h)(W_S x_S + b_S)^{T} \in \mathbb{R}^{l \times n}
\end{align}
where $W_h \in \mathbb{R}^{dim \times dim}$, $W_S \in \mathbb{R}^{dim \times dim}$, $b_h \in \mathbb{R}^{dim}$ and $b_S \in \mathbb{R}^{dim}$.

The hidden features $h$ on the other side are passed through a fully connected layer to project into vocab space $h' \in \mathbb{R}^{l \times v}$  where $v$ is the size of defined vocab. Then the final output $o \in \mathbb{R}^{l}$ is determined by 
\begin{align}
o = max([h', ocr])
\end{align}
the concatenation operator $[,]$ is performed along the last dimension.

\section{Experiments and Results}
\label{sect:experiment}

\subsection{Baseline Models}

We compare our proposed models with several powerful baselines described as follows.

\begin{itemize}
    \item \textbf{SAAA}: This is a strong baseline proposed by Google \cite{kazemi2017show} for the VQAv1 dataset \cite{VQA}, SAAA follows the stack attention mechanism \cite{yang2016stacked} to combine features from images and features from questions, then used this combination to select answer over a defined set, thus this is a sort of classification approach for the VQA task.
    
    \item \textbf{MCAN}: Proposed on the VQAv2 dataset \cite{yu2019deep}, MCAN was designed to use image features that guide the features of questions to yield the combined features, then \cite{yu2019deep} based on these combined features through the reduction module to reduce the features, MCAN selects the appropriate answer.
    
    \item \textbf{LoRRA}: Proposed together with the TextVQA dataset \cite{singh2019towards}, LoRRA was designed to have the ability to use scene texts available in images and select from these scene texts together with a defined set of answers an appropriate answer. Although it is able to read texts in images, LoRRA is a sort of classification approach for the VQA task.
    \item \textbf{M4C}: Proposed on the TextVQA dataset \cite{hu2020iterative} by Facebook, this method is the first method that tackles the VQA task as a generation task. M4C was designed with BERT \cite{Devlin2019BERTPO} as the question embedding and it uses the encoder module of BERT \cite{Devlin2019BERTPO} as the whole encoder, or the multimodal encoder, for embedding all forms of embedded features (object features or region features of objects in images \cite{Anderson2017BottomUpAT}, question features from BERT model, visual features of scene texts available in images and features of previous tokens in answers). Moreover, as treated the VQA task as a generation task rather than a classification task, M4C uses another way of copying scene texts in images to answers. In particular, instead of selecting scene texts in images as answers like LoRRA, Hu et al. \cite{hu2020iterative} designed the \textit{Dynamic Pointer Network} that fuses the features of embedded features of all tokens in vocab with the embedded features of scene texts then M4C has to select whether tokens from vocab or scene texts should appear at the step t in the answer generation process.
\end{itemize}

\subsection{Evaluation Metrics}
Inspired by the machine translation task and the image captioning task, we measure the distance between machine-generated answers (hypothesis - hypo) and human-given answers (reference - ref). Particularly, we used BLEU (BLEU@1, BLEU@2, BLEU@3, BLEU@4) \cite{papineni-etal-2002-bleu}, ROUGE (ROUGE-L) \cite{ganesan2018rouge}, METEOR \cite{banerjee-lavie-2005-meteor} and CIDEr \cite{vedantam2015cider} for evaluating the performances of visual question answering models on our dataset.

\subsubsection{BLEU}
This metric mainly depends on the color of the precision metric. Papineni et al. \cite{papineni-etal-2002-bleu} designed the BLEU metric following two observations (1) occurrence of n-gram tokens in hypo should not exceed its occurrence in ref, and (2) hypo having a length longer than one of ref should be assigned a low weight (penalty weight). Particularly, the score of a hypo given its ref is specified as:

\begin{align}
score_{token} = \frac{Count_{clip}(token)}{Count(token)}
\end{align}

then based on this formula, the score for all hypo in the dataset is designed as:

\begin{align}
p_n = \frac{\sum_{h \in hypothesis}\sum_{token \in h}Count_{clip}(token)}{\sum_{h \in hypothesis}\sum_{token \in h}Count(token)}
\end{align}
where $n$ is the value of the used n-gram.

The formula of $p_n$ already tackles the case that the length of hypo is greater than one of ref, the remaining case to be considered is the length of hypo is smaller than one of ref. To tackle this case, let $c$ be the length of all hypo in the dataset and $r$ the length of all ref in the dataset, the penalty weight for hypo having a length greater than the length of ref is designed as follows:

\begin{align}
    BP = e^{1-\frac{r}{c}}
\end{align}
obviously BP = 1 when $c > r$.

Finally, the BLEU score is retrieved as:

\begin{align}
    log BLEU = min(1-\frac{r}{c}, 0) + \sum_{n=1}^N w_n log p_n
\end{align}

In this paper we used $n \in \{1, 2, 3, 4\}$ which are BLEU@1, BLEU@, BLEU@3 and BLEU@4, respectively.

\subsubsection{ROUGE}
ROUGE shares the same characteristic as recall. Ganesan et al. \cite{ganesan2018rouge} designed the ROUGE metric in order to specify the ratio of common n-gram tokens between hypo and ref with n-gram tokens in ref. Apart from here, the remaining issue we encounter is the way we specify the common tokens. In this paper, we used the most common type of ROUGE metric which is the ROULE-L using the Longest Common Subsequence method (LCS) to specify the common n-gram tokens.

In particular, Ganesan et al. \cite{ganesan2018rouge} specified the recall $R$ and precision $P$ based on LCS between hypo and ref as follow:

\begin{align}
    R_{LCS} = \frac{LCS(hypo, ref)}{m}
\end{align}

\begin{align}
    P_{LCS} = \frac{LCS(hypo, ref)}{n}
\end{align}
then ROUGE-L is specified as:

\begin{align}
    ROUGE = \frac{(1 + \beta)^2R_{LCS}P_{LCS}}{R_{LCS} + \beta^2 P_{LCS}}
\end{align}
where $m$ is the length of the ref and $n$ is the length of hypo.

\subsubsection{METEOR}
BLEU and ROUGE define tokens based on n-gram tokens, while METEOR \cite{banerjee-lavie-2005-meteor} approached another way to measure the similarity between hypo and ref. Banerjee et al. \cite{banerjee-lavie-2005-meteor} assumed there are many cases when n-gram tokens swapped their position the meaning of the whole sentence is not changed, but BLEU and ROUGE metrics assign low scores for these cases. To tackle this situation, Banerjee et al. \cite{banerjee-lavie-2005-meteor} first defined the \textit{alingment} between hypo and ref. Alignments in their turn are defined as the set of mappings, where each mapping is specified as a connection between tokens in hypo and ref. Note that in this case a token is defined as a 1-gram token.

\begin{figure}[ht]
    \centering
    \begin{subfigure}{0.49\textwidth}
        \includegraphics[width=\textwidth]{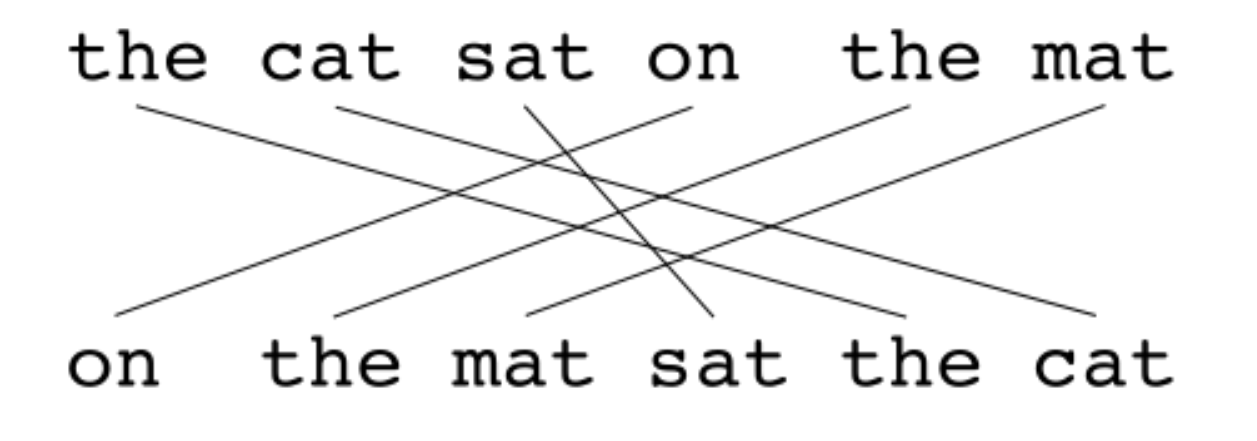}
        \caption{}
    \end{subfigure}
    \begin{subfigure}{0.49\textwidth}
        \includegraphics[width=\textwidth]{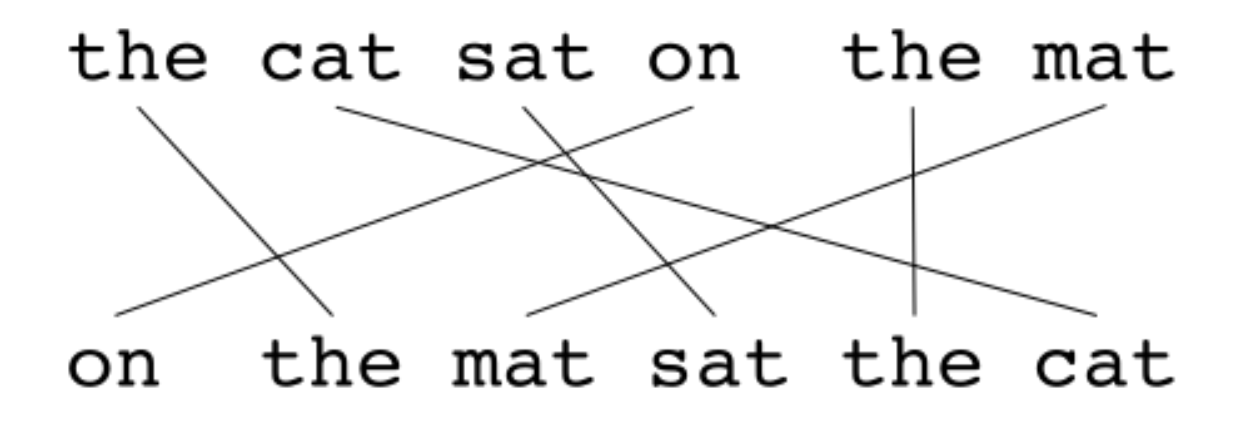}
        \caption{}
    \end{subfigure}
    \caption{There are many alignments between hypo and its ref.}
    \label{fig:meteor_alignment}
\end{figure}

Actually, there are a lot of alignments between a particular hypo and ref, but the selected alignment is one having the least intersections. For example, in Figure \ref{fig:meteor_alignment}, the first alignment is selected as the alignment between hypo and ref. Then the precision $P$ between hypo and ref based on their alignment is defined as:
\begin{align}
    P = \frac{m}{w_h}
\end{align}
and the recall $R$ between hypo and ref between them is defined as:
\begin{align}
    R = \frac{m}{w_r}
\end{align}
then the correlation between $P$ and $R$ is:
\begin{align}
    F_{mean} = \frac{10PR}{R + 9P}
\end{align}

Similar to BLEU, METEOR specifies the penalty weight for hypo that is longer than or shorter than its ref by taking into account the common tokens of hypo and ref. In particular, the penalty weight $p$ is defined as:
\begin{align}
    p = 0.5 \left(\frac{c}{u_m}\right)^3
\end{align}
where $c$ is the length of common unigrams of hypo and its ref and $u_m$ is the total unigrams appearing in both hypo and ref.

Finally, with the penalty weight and correlation between precision $P$ and recall $R$ the METEOR score between hypo and its ref is specified as:
\begin{align}
    M = F_{mean}(1-p)
\end{align}

\subsubsection{CIDEr}
Although having overcome the disadvantages of BLEU and ROUGE, METEOR still can not take into account the semantic similarity of hypo and ref. In particular, Vedantam et al. \cite{vedantam2015cider} indicated by defining the alignment between hypo and ref based on their mappings, METEOR implicitly assigns the same weight for all unigrams. Nonetheless, there are many cases that answer containing tokens that are not relevant to the inquiries information at all, and there are many cases as well when the hypo contains an exact token which makes the hypo closer to the ref but METEOR treats this token as the same way as other tokens. Vedantam et al. \cite{vedantam2015cider} state that such crucial tokens in hypo should be given higher weight when calculating the distance between hypo and ref, hence they proposed CIDEr having the ability to take into account the semantic similarity of hypo and ref.

For more details, CIDEr is constructed based on two observations (1) n-grams available in ref should not appear in hypo and (2) n-grams that concurrently appear among many images carry information not relevant to any particular images. To model these two observations, Vedantam et al. \cite{vedantam2015cider} took advantage of Term Frequency and Inverse Document Frequency (TF-IDF). Let $h_k(s)$ is the total occurrence of n-gram $w_k$ in sentence $s$, we have:

\begin{align}
    g_k(ij) = \frac{h_k(r_{ij})}{\sum_{w_l \in \Omega} h_l(r_{ij})} log\left( \frac{|\text{Q}|}{\sum_{Q_p \in Q}min(1, \sum_q h_k (s_{qp})} \right)
\end{align}

in which $g_k(ij)$ is the TF-IDF weight of n-gram $w_k$ in ref i of question j, $\Omega$ is the set of all n-grams and $Q$ is the set of all questions in dataset.

After determining the TF-IDF weight for all n-grams, the CIDEr score of hypo and its ref is specified based on the cosine similarity between them:

\begin{align}
    CIDEr(h_i, r_{ij}) = \frac{1}{m} \sum_j \frac{g^n(h_i) \cdot g^n(r_{ij})}{g^n(h_i) \cdot g^n(r_{ij})}
\end{align}

\subsection{Feature Extraction}

\subsubsection{Image Features Extraction} \label{sec:image_features}

In this paper, we used the FasterRCNN \cite{ren2015faster} with ResNeXt152++ \cite{jiang2020defense} as its backbone to extract region features \cite{Anderson2017BottomUpAT} from images for LoRRA, FST, MCAN, QuMLAG, and M4C. We also followed \cite{jiang2020defense} to use ResNeXt152++ to achieve grid features for SAAA and FST.

\subsubsection{Scene Text Features Extraction} \label{sec:scene_text_feature}
Hu et al. \cite{hu2020iterative} used the Rossetta system \cite{borisyuk2018rosetta} to achieve bounding boxes of scene texts available in images, then they used ROI Alignment to extract region features for each scene text from features map from the backbone of FasterRCNN \cite{ren2015faster} and they used these features as the features of describing the appearance of scene texts in images. Moreover, they used PHOC \cite{6857995} to extract the character appearance or detailed appearance of scene texts and used FastText \cite{bojanowski2017enriching} to get their linguistic meaning. Finally, contextual appearance features, detailed appearance features, and linguistic meaning of scene texts \cite{hu2020iterative} combine them to yield the context features for scene texts.

Nevertheless, Rossetta system is not publicly available and it does not support Vietnamese. Thus we have to use other scene text models that were trained on the Vietnamese scene text datasets such as VinText \cite{m_Nguyen-etal-CVPR21}. Particularly we used SwinTextSpotter \cite{huang2022swintextspotter} to extract bounding boxes as well as features for scene texts.

\subsection{Experimental Configuration}
Each method has its own set of hyperparameters as well as training configuration, thus in this section, we carefully detail the hyperparameters as well as the relevant configuration for each of the baselines and proposal generator methods. In general, we trained all baselines and proposed methods with batch data of size 64. For classifier-based methods, we fixed the learning rate at 0.01 and used the cross-entropy loss as the objective function. Note that in the original configuration of classier-based methods \cite{kazemi2017show,yu2019deep,singh2019towards}, authors used multi-label cross-entropy loss rather than cross-entropy loss. However, our experiments imply that using multi-label cross-entropy loss did not work and most classifier-based methods obtained results converging to 0 on all metrics. For generator-based methods, we adapt the learning rate scheduler from the training process of transformer \cite{vaswani2017attention}. Particularly, learning rate scheduler used in \cite{vaswani2017attention} is formed as follow:

$$
    learning\_rate = d_{model}^{-0.5} \times min(step\_num^{-0.5}, step\_num \times warmup\_steps^{-1.5})
$$

In our settings, we set $warmup\_steps = 10,000$. We used the early stopping technique to stop the training process when the CIDEr score does not increase after 5 epochs. We use Adam \cite{Kingma2014AdamAM} as the optimization method for all experiments. All experiments were implemented using an A100 GPU. Particular configuration for each method is detailed as follows.

\subsubsection{Experimental Settings for Baseline Models}

{\bf SAAA}: Most of the configurations for training SAAA were kept as in \cite{kazemi2017show} except for grid features of images extracted using ResNeXt152++ \cite{jiang2020defense} as described in Section \ref{sec:image_features}.

{\bf MCAN}: For the Information Fusion module, we used 4 layers for the GA module and 4 layers as well for the SA module. Each attention module in GA and SA has 8 heads with hidden dimensions of size 512. Moreover, as questions in the OpenViVQA dataset are Vietnamese, we used FastText \cite{bojanowski2017enriching} to achieve word embeddings for question tokens rather than GloVe \cite{pennington-etal-2014-glove} as in \cite{yu2019deep}.

{\bf LoRRA}: LoRRA in our experiments is kept as its original version proposed in \cite{singh2019towards}. Nevertheless, we used FastText \cite{bojanowski2017enriching} for scene texts extracted from images as well as question tokens instead of using GloVe \cite{pennington-etal-2014-glove} for question tokens. Scene texts from images are recognized using SwinTextSpotter \cite{huang2022swintextspotter}. The Context Attention module, the Self Attention module, and the Spatial Attention module of the Information Fusion module  all have 8 heads with hidden dimensions of size 512.

 
{\bf M4C}: In our experiments, we kept all original configurations of M4C including BERT-based-uncased \cite{Devlin2019BERTPO} as the Question Embedding module and the encoder module of BERT as the MultiModal module. The BERT model in the Question Embedding module was loaded and updated from its pre-trained weights while training on the OpenViVQA dataset.

\subsubsection{Experimental Settings for Proposed Models}

{\bf FST}: We used ResNeXt152++ \cite{jiang2020defense} in order to extract grid features from images. The LSTM \cite{hochreiter1997long} of the Question Embedding module has the hidden dimension of size 512 with 1 layer. The Stack Attention module of the Fusion module uses 2 glimpses to perform stacked attention. For the Answer Generator module, we used the decoder of transformer \cite{vaswani2017attention} with 3 layers. Each layer contains an attention module performing the self-attention mechanism and an attention module performing the cross-attention mechanism. Each attention module has 8 heads and hidden dimensions of size 512.

{\bf QuMLAG}: GMCAN in our experiments used  FasterRCNN \cite{ren2015faster} with ResNeXt152++ \cite{jiang2020defense} as its backbone to extract region features from images. The Text Embedding module of GMCAN also uses FastText \cite{bojanowski2017enriching} instead of GloVe \cite{pennington-etal-2014-glove} to be adaptable for Vietnamese. The Information Fusion module includes 4 layers of the GA module and 4 layers as well for the SA \cite{yu2019deep} module. Each attention module in GA and SA has 8 heads with hidden dimensions of size 512. The Answer Generator module includes 3 layers, each layer contains an attention module for self-attention and an attention module for cross-attention. Each attention module has 8 heads with hidden dimensions of size 512.

{\bf MLPAG}: MLPAG uses FasterRCNN \cite{ren2015faster} with ResNeXt152++ \cite{jiang2020defense} to obtain region features of the images. Scene texts in images are recognized using SwinTextSpotter \cite{huang2022swintextspotter}. Both scene texts and questions are embedded using FastText \cite{ren2015faster}. The decoder used for the Answer Generator module of MLPAG contains 3 layers, each layer has two attention layers, one layer performs self-attention while the other performs cross-attention. Each attention layer has 8 heads with 512 as its hidden dimension. 

\subsection{Experimental Results} \label{sect:main_result}
We evaluated baselines and our three proposal methods and reported their results in Table \ref{tab:main_results}.

\begin{table}[ht]
    \resizebox{\textwidth}{!}{
    \begin{tabular}{lccccccc}
    \hline
    \multicolumn{1}{c}{\textbf{Method}} & \textbf{BLEU@1} & \textbf{BLEU@2} & \textbf{BLEU@3} & \textbf{BLEU@4} & \textbf{METEOR} & \textbf{ROUGE}  & \textbf{CIDEr}  \\ \hline
    SAAA                   & 0.0091          & 0.0043          & 0.0001          & 0.0000          & 0.0533          & 0.0605          & 0.1230          \\
    MCAN                   & 0.2366          & 0.1808          & 0.1446          & 0.1179          & 0.1531          & 0.2718          & 1.0613          \\
    LoRRA                  & 0.1923          & 0.1430          & 0.1128          & 0.0917          & 0.1271          & 0.2137          & 0.8005          \\ 
    M4C                  &    0.3737      &     0.2755      &    0.2078       &    0.1597       &    0.2107       &     0.3789      &    1.5073       \\
    FST (ours)                   & 0.1710          & 0.1154          & 0.0813          & 0.0582          & 0.1157          & 0.2010          & 0.6141          \\
    QuMLAG (ours)                   & 0.3548          & 0.2863          & 0.2157          & 0.2037          & 0.2129          & 0.3786          & 1.7082          \\
    MLPAG (ours)                  & 0.3811          & 0.2927          & 0.2328          & 0.1889          & 0.2230           & 0.3772          & 1.6104          \\ \hline
    \end{tabular}}
    \caption{Exprimental results of baselines and proposed methods on the OpenViVQA dataset.}
    \label{tab:main_results}
\end{table}

\begin{figure}[ht]
    \centering
    \begin{subfigure}{0.47\textwidth}
        \includegraphics[width=\textwidth]{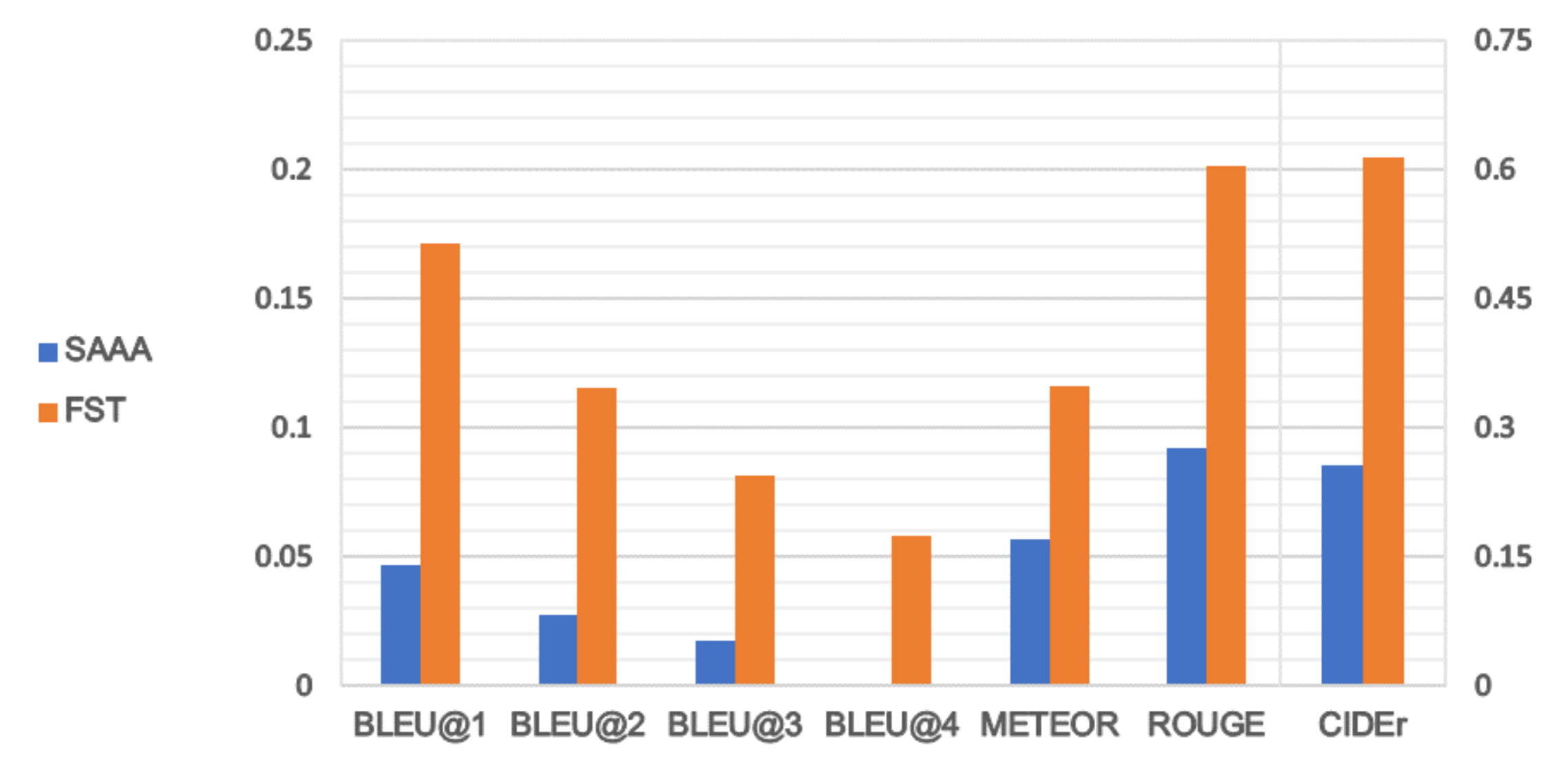}
    \end{subfigure}
    \begin{subfigure}{0.47\textwidth}
        \includegraphics[width=\textwidth]{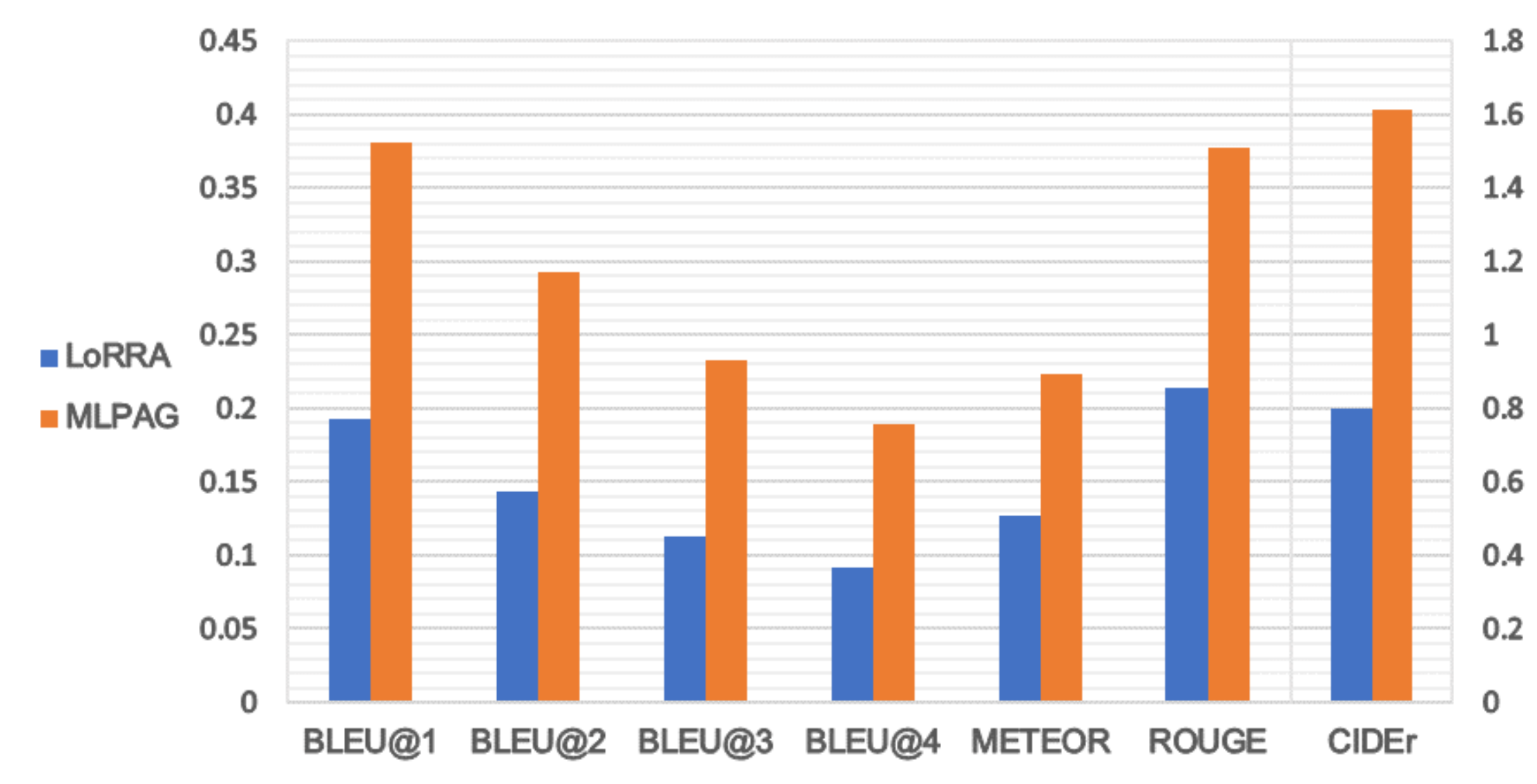}
    \end{subfigure}
    \begin{subfigure}{0.5\textwidth}
        \includegraphics[width=\textwidth]{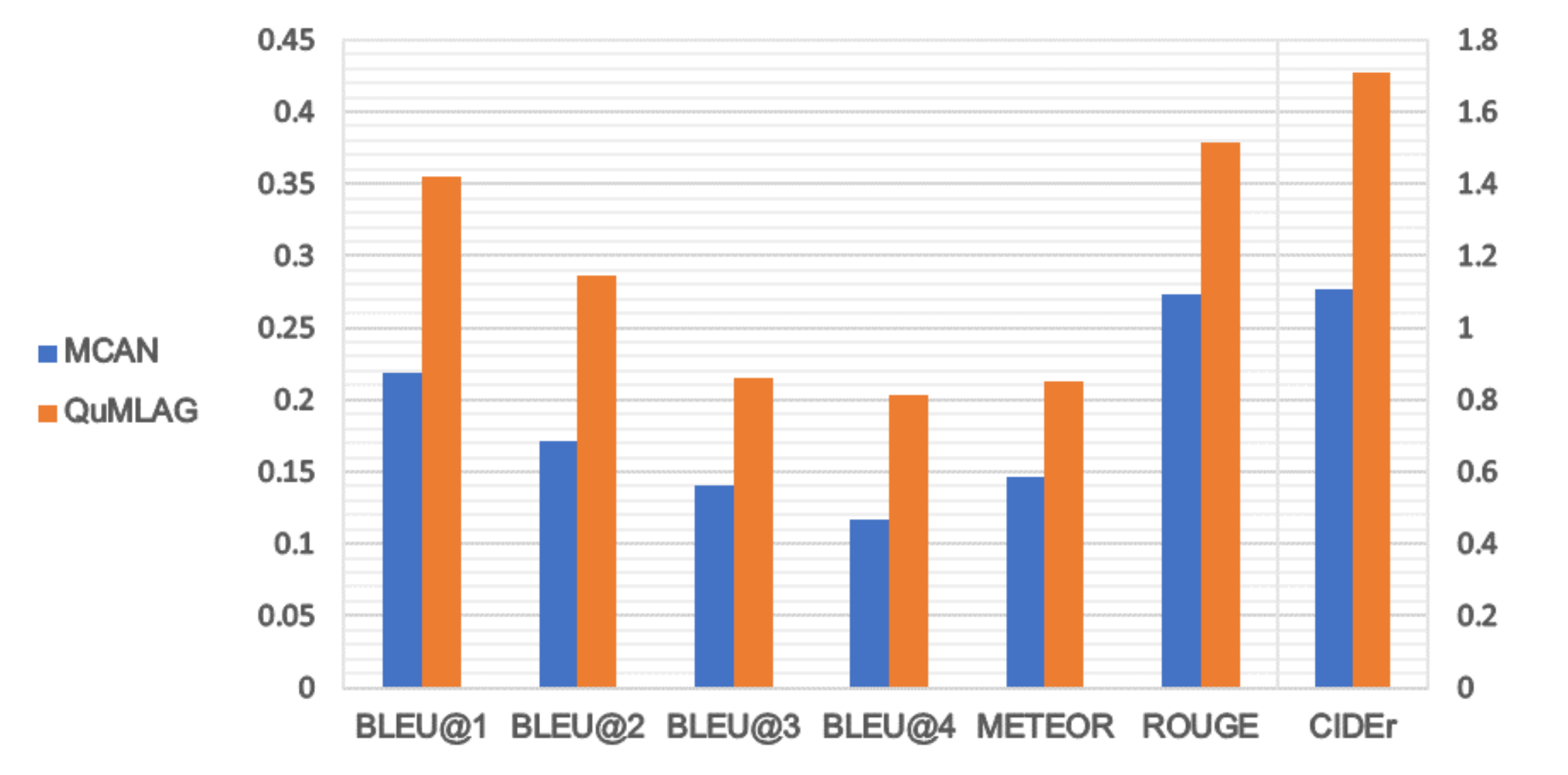}
    \end{subfigure}
    \caption{Our proposed methods achieved better results compared to their respective baselines.}
    \label{fig:classifier_generator}
\end{figure}

According to Table \ref{tab:main_results}, our proposed methods achieved better results compared to all baselines (Figure \ref{fig:classifier_generator}). These results proved that classifier-based methods can not perform well on the OpenViVQA dataset and we believe such methods can not work as well on open-ended VQA datasets in general. In particular, FST only outperforms SAAA which also uses the Stacked Attention mechanism. Question-guided Attention works better than Stack Attention, even it helps QuMLAG obtain higher scores than MLPAG without reading and using scene texts on BLEU@4, ROUGE, and CIDEr metrics. Especially M4C has the largest amount of parameters but they yielded lower results than our proposal. We will give detailed comments on M4C in Section \ref{sec:m4c_result}.

\section{Results Analysis}
\label{sect:result_analysis}



\subsection{Effect of Question and Answer Lengths on Experimental Results}

\begin{figure}
    \centering
    \begin{subfigure}{0.45\textwidth}
        \centering
        \begin{tabular}{ll}
            \hline
            \textbf{Group} & \textbf{Length (n)}     \\ \hline
            Short (S)      & $n \le 5$               \\
            Medium (M)     & $5 < n \le 10$  \\ 
            Long (L)       & $10 < n \le 15$ \\ 
            Very Long (XL) & $n > 15$              \\ \hline
        \end{tabular}
        \caption{}
        \label{tab:qa_groups}
    \end{subfigure}
    \begin{subfigure}{0.45\textwidth}
        \includegraphics[width=\textwidth]{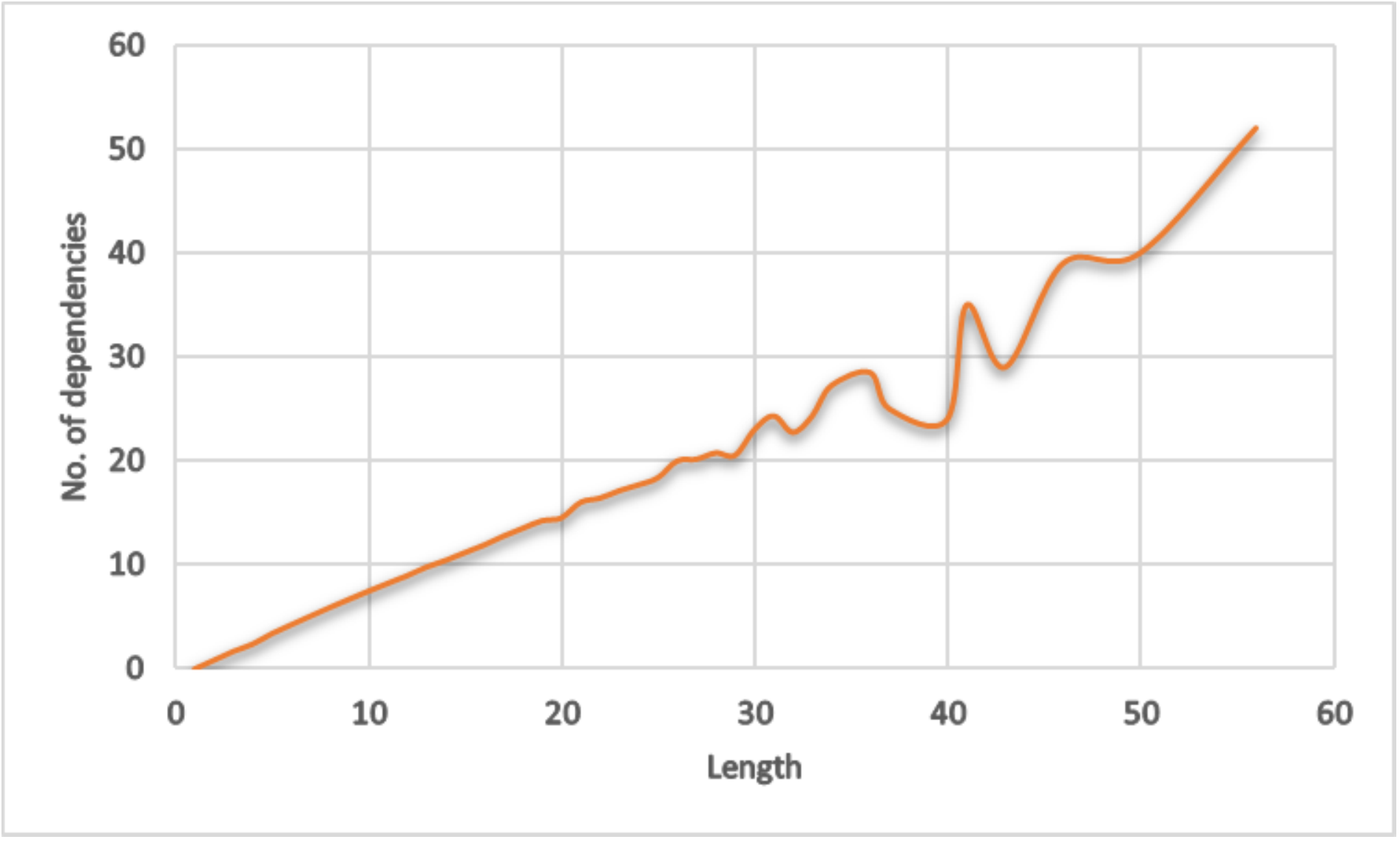}
        \caption{}
        \label{fig:dep_len}
    \end{subfigure}
    \caption{Groups of questions and answers based on their total number of tokens (a) and total number of dependencies based on the length of sentence (b).}
\end{figure}

Intuitively, we can use the LCS or LSS algorithm (Section \ref{sect:dataset}) to analyze the results of the experimental methods using the total number of semantic dependencies or semantic trees. However, such a way is not general as it concentrates on the linguistic complexity of languages. We argue that analyzing based on the length of sentences is more natural as people do not pay attention to how complicated the words they said but to how is the length of sentences they made. People who prefer using words or phrases to respond to given questions usually speak faster than those who respond politely in sentences. Moreover, as shown in Figure \ref{fig:dep_len}, the longer the sentences, the more linguistically complicated they are. This is intuitive as more tokens lead to more dependencies occurring between them. Thus, analyzing the results based on the length of sentences is more comprehensive because it covers both habits of people when using languages and the linguistic complexity of sentences.


\begin{figure}[ht]
    \centering
    \begin{subfigure}{0.3\textwidth}
        \includegraphics[width=\textwidth]{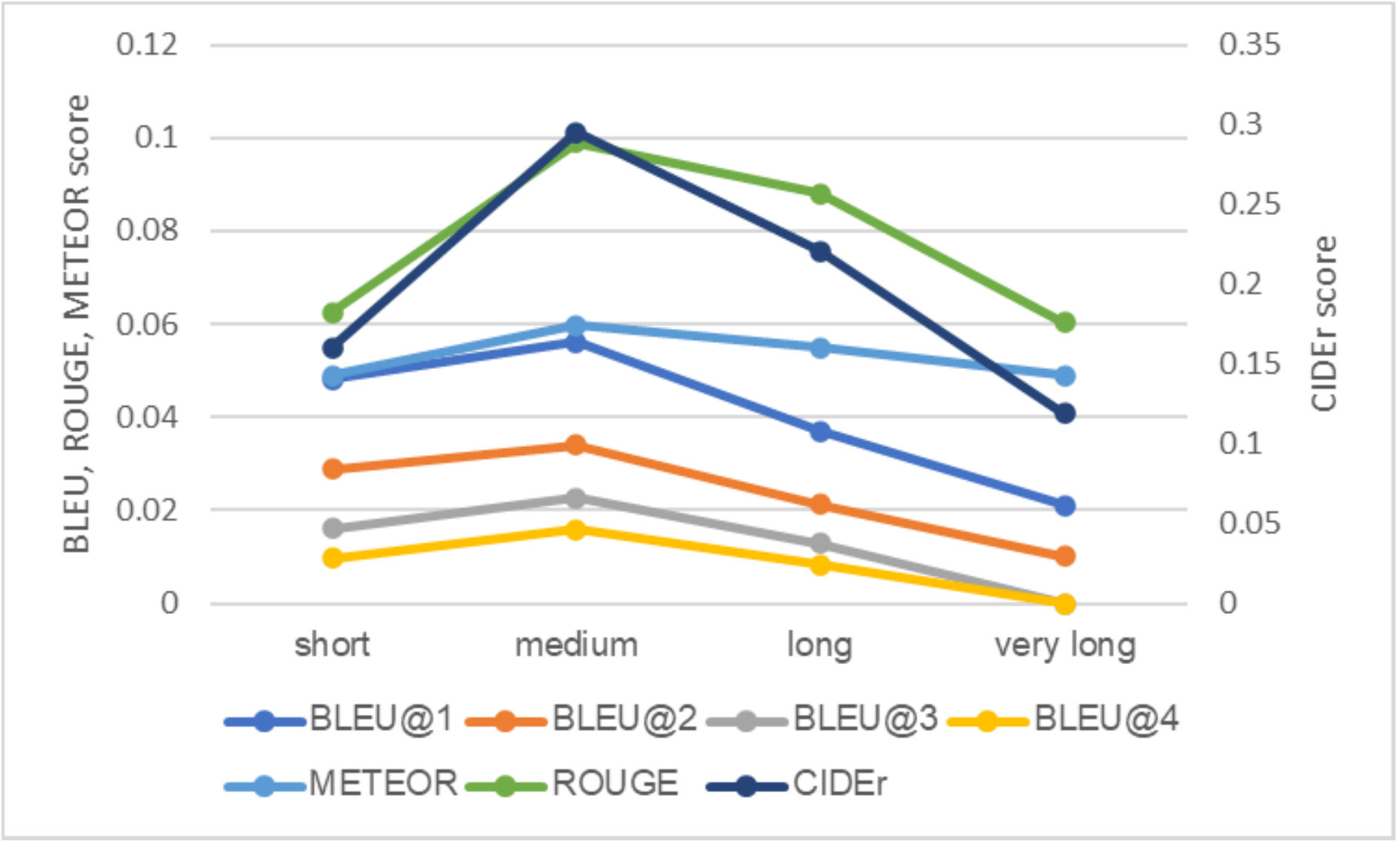}
        \caption{SAAA}
    \end{subfigure}
    \begin{subfigure}{0.3\textwidth}
        \includegraphics[width=\textwidth]{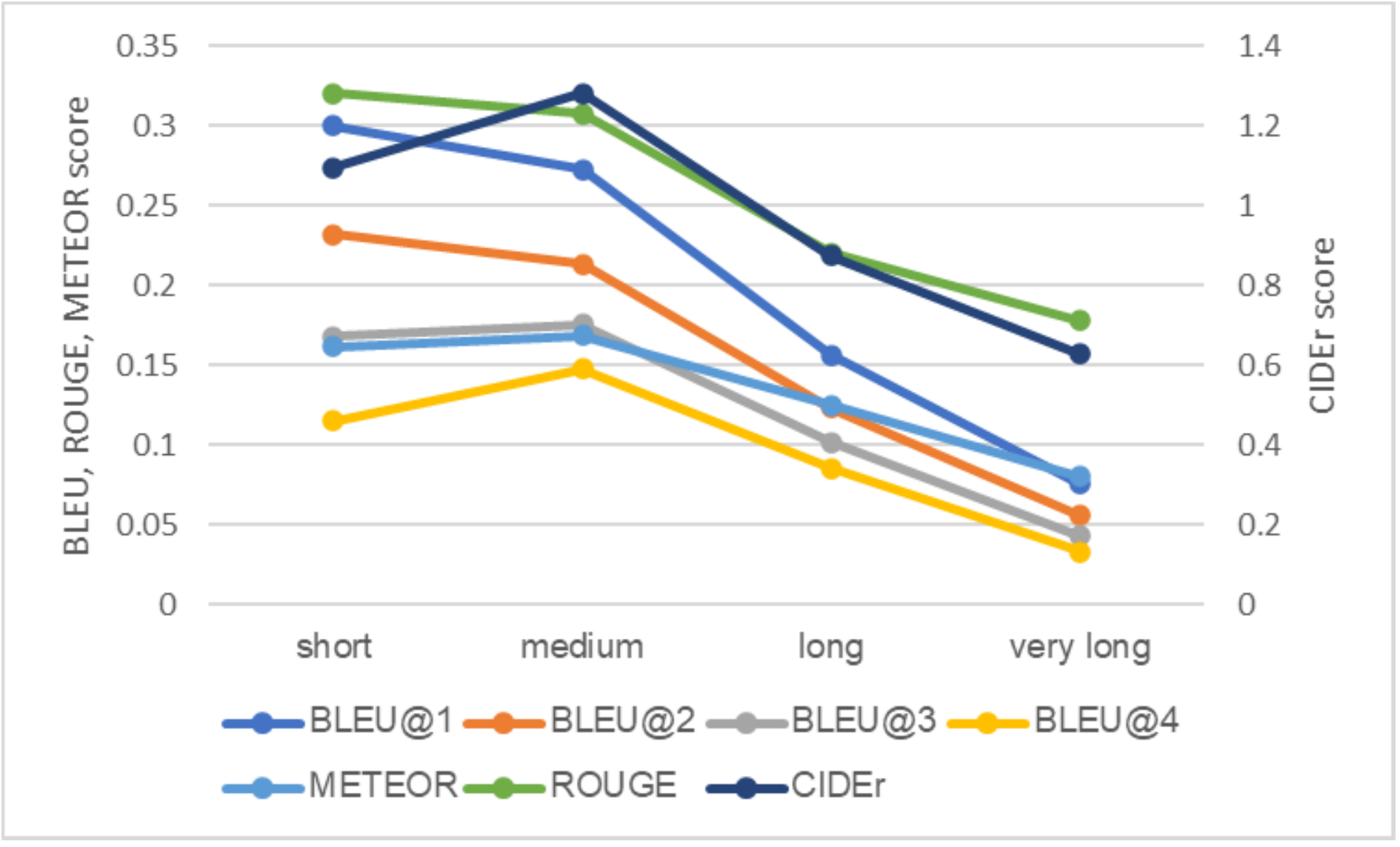}
        \caption{MCAN}
    \end{subfigure}
    \begin{subfigure}{0.3\textwidth}
        \includegraphics[width=\textwidth]{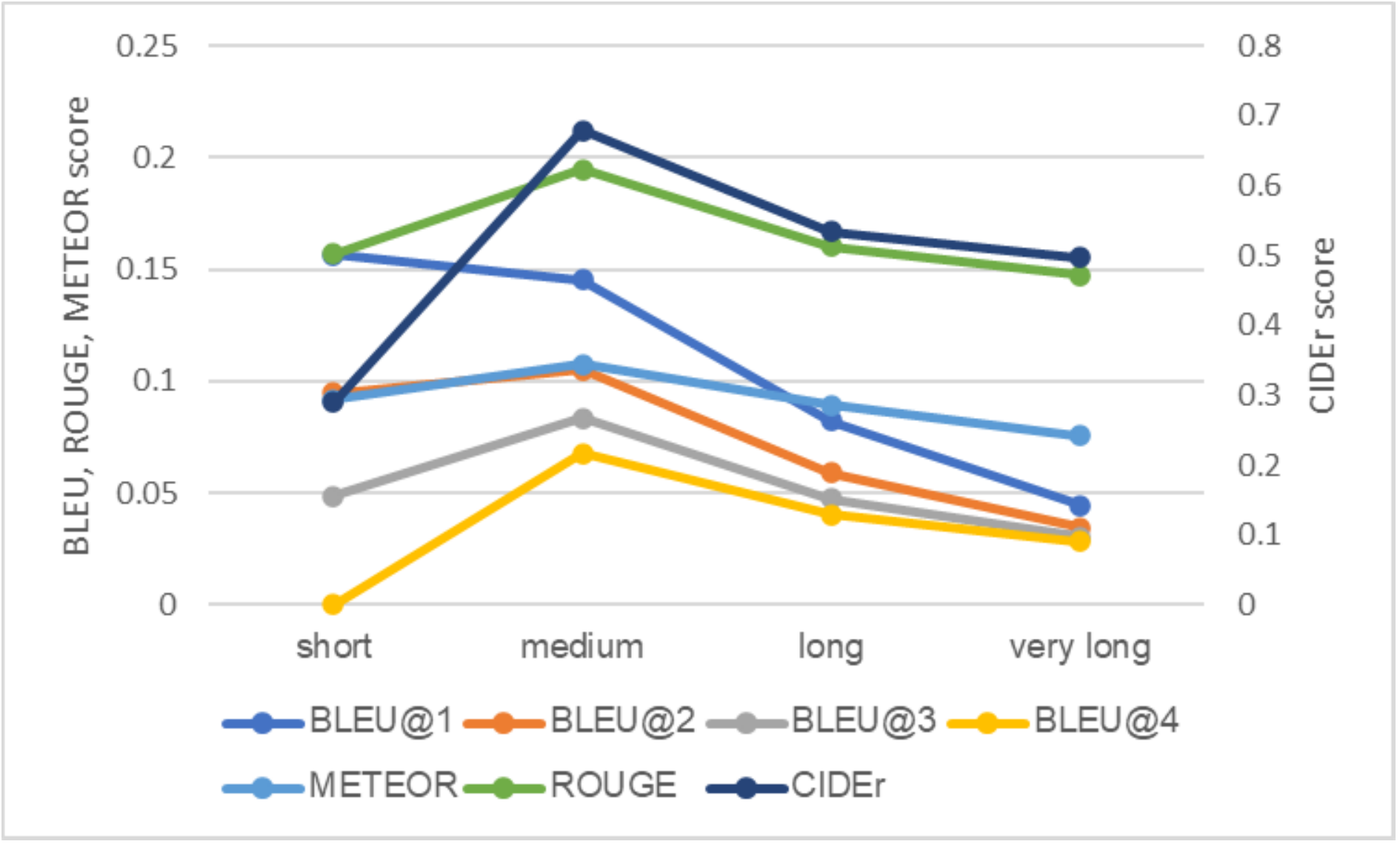}
        \caption{LoRRA}
    \end{subfigure}
    \begin{subfigure}{0.3\textwidth}
        \includegraphics[width=\textwidth]{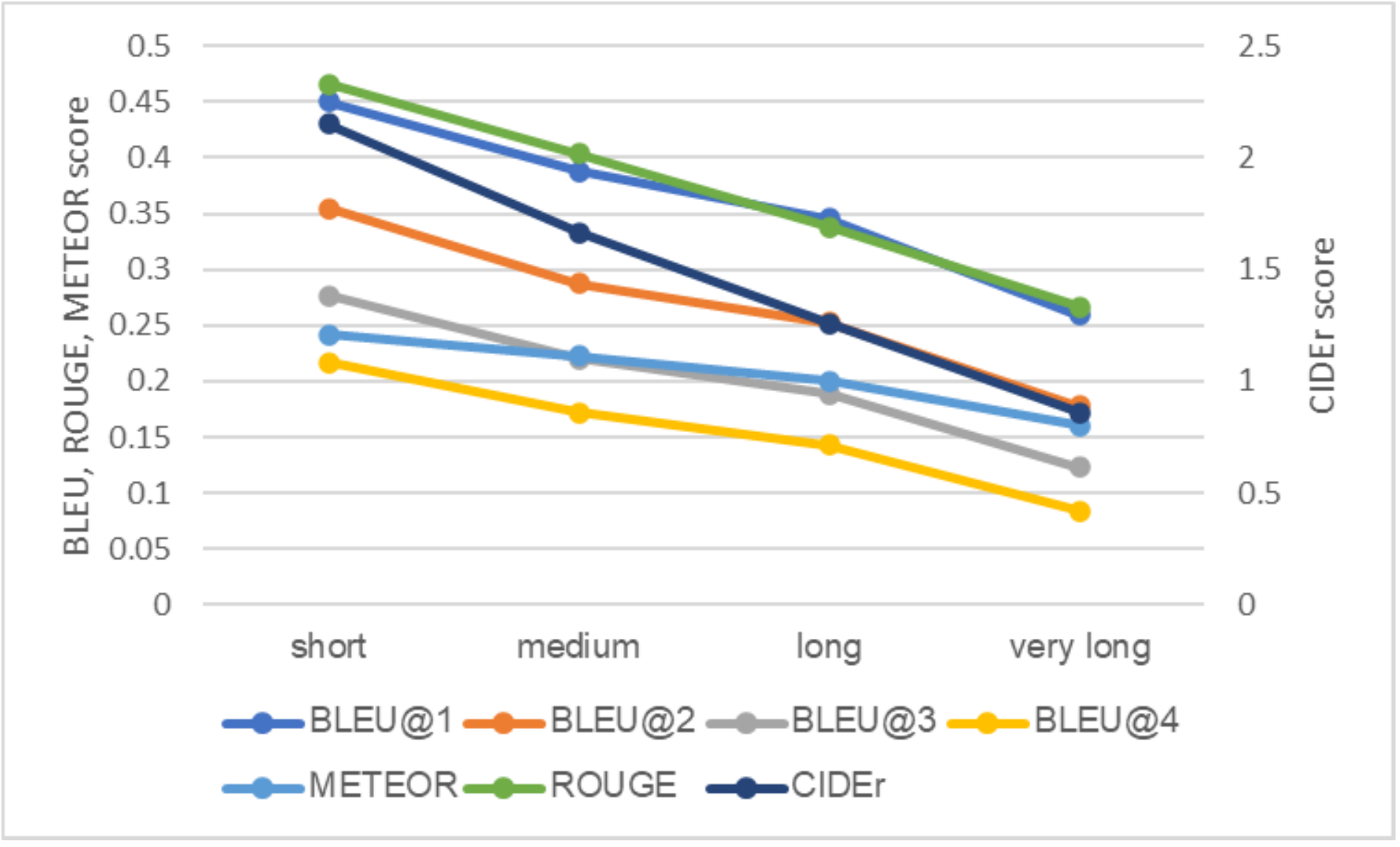}
        \caption{M4C}
    \end{subfigure}
    \begin{subfigure}{0.3\textwidth}
        \includegraphics[width=\textwidth]{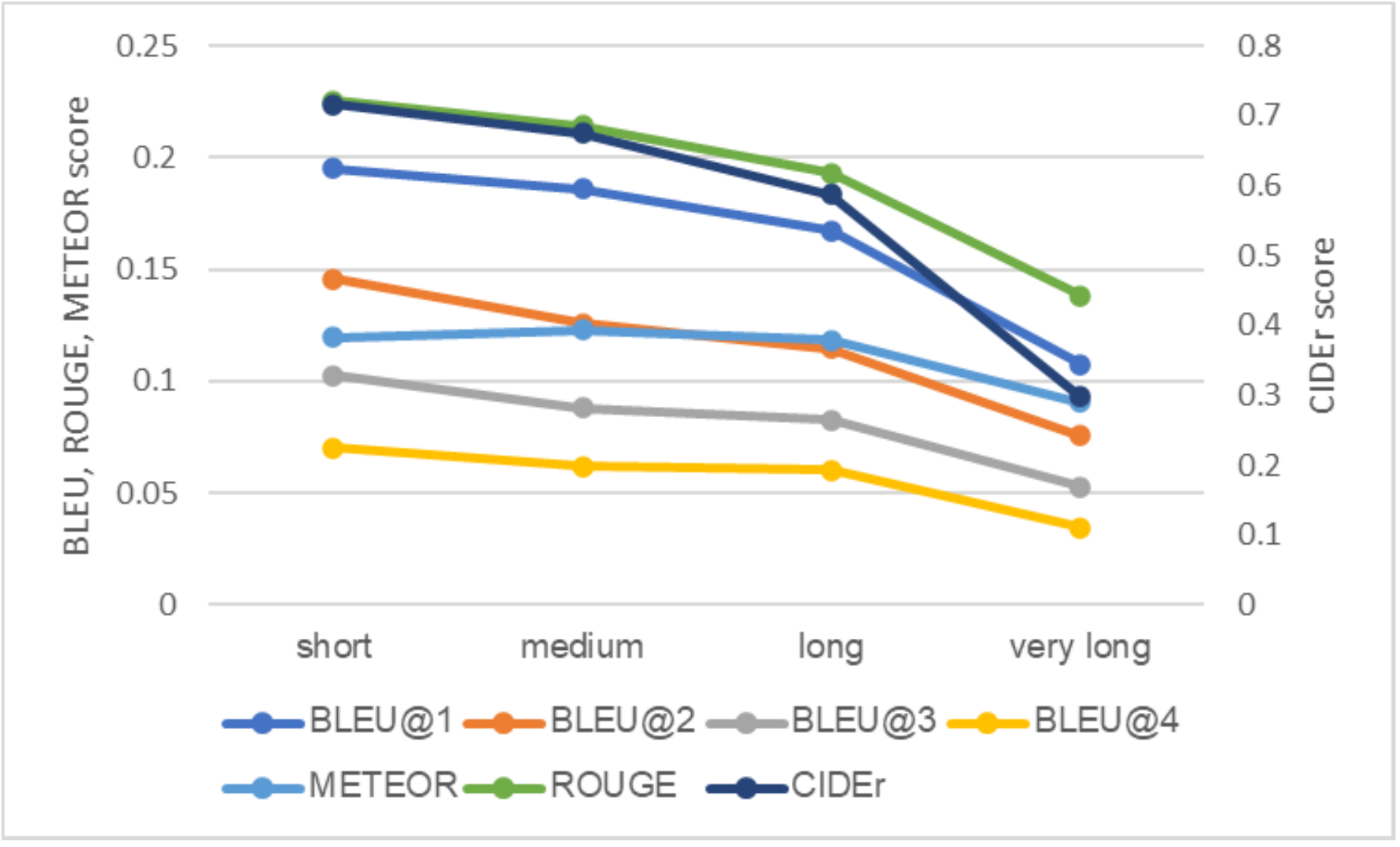}
        \caption{FST}
    \end{subfigure}
    \begin{subfigure}{0.3\textwidth}
        \includegraphics[width=\textwidth]{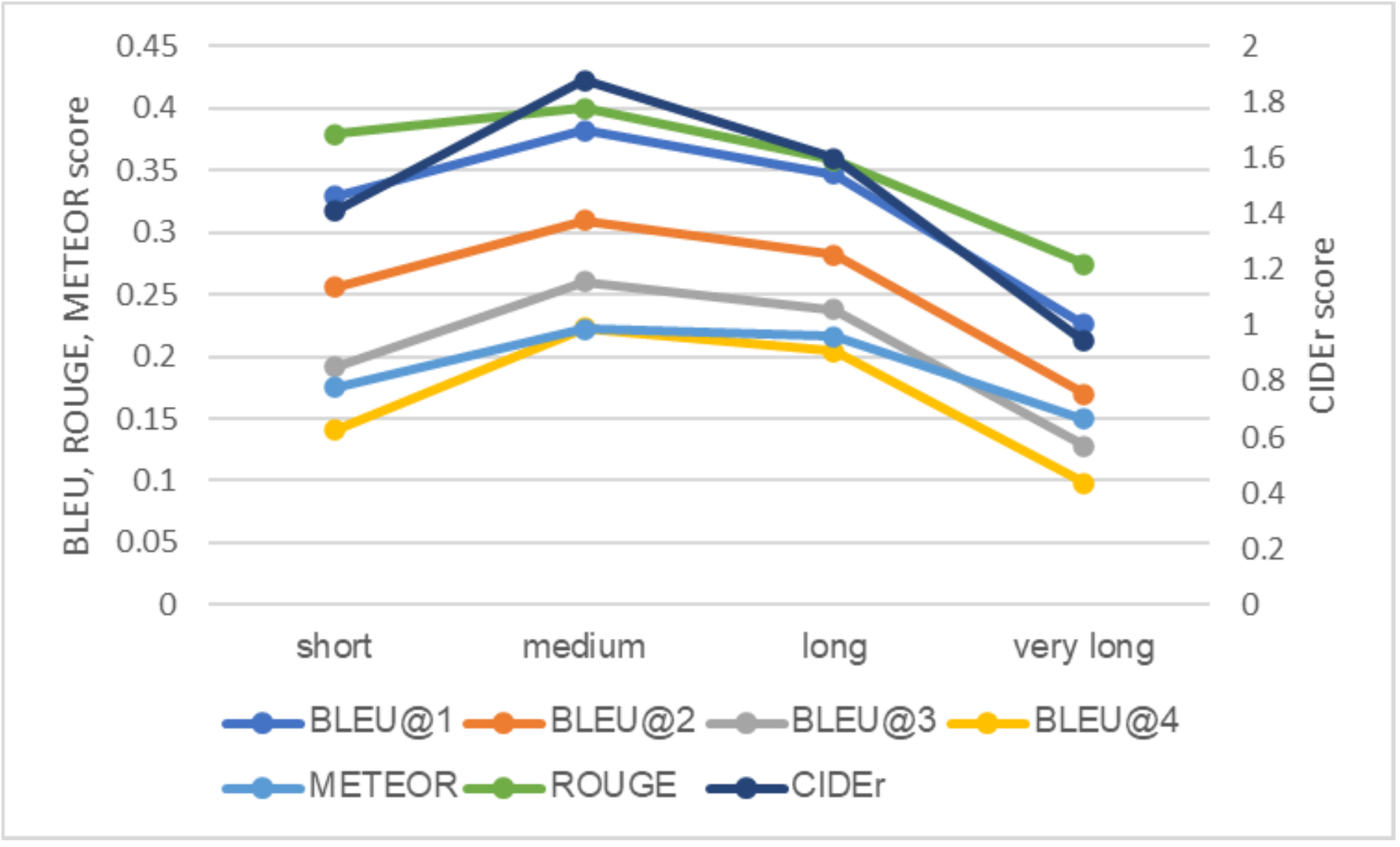}
        \caption{QuMLAG}
    \end{subfigure}
    \begin{subfigure}{0.3\textwidth}
        \includegraphics[width=\textwidth]{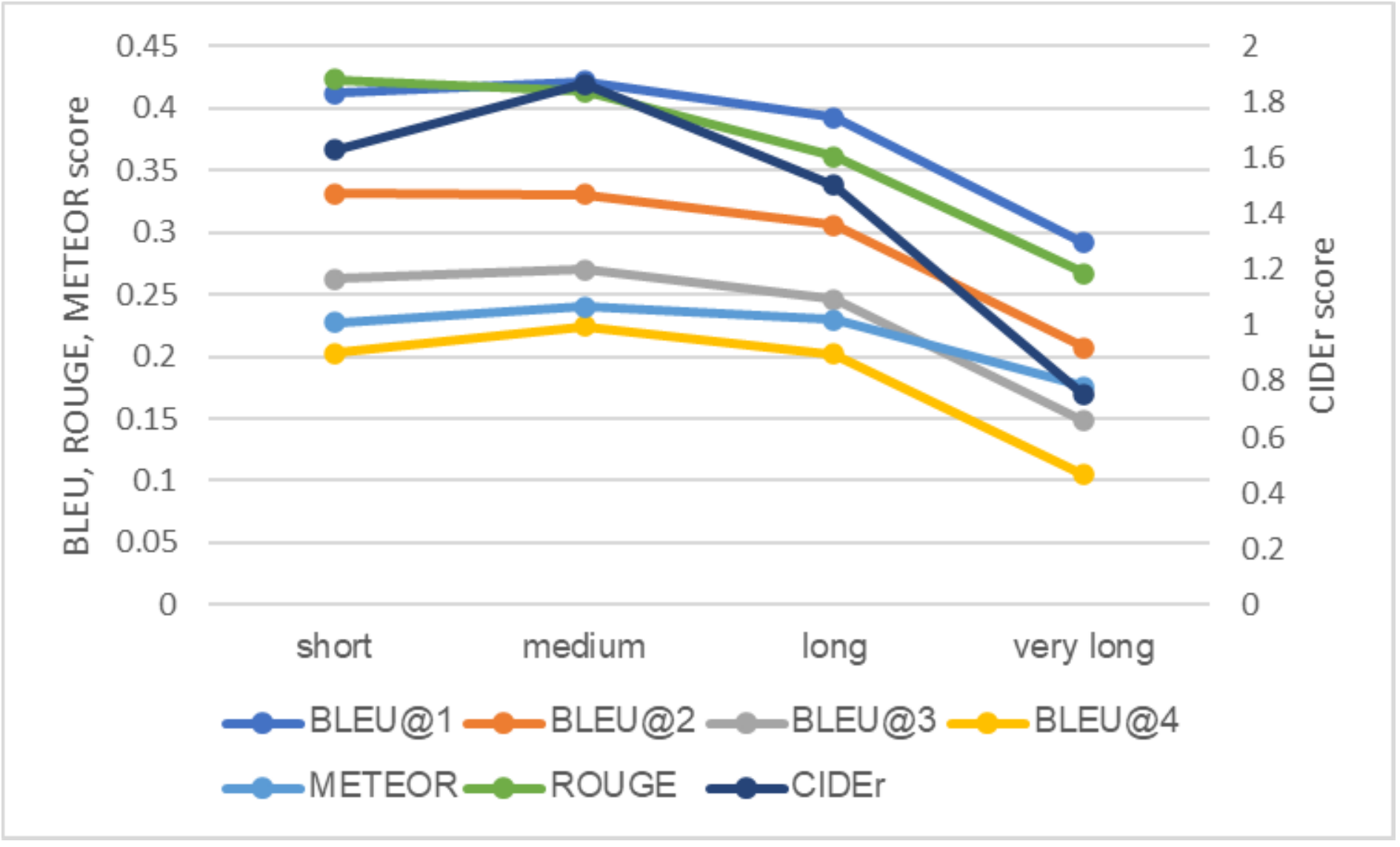}
        \caption{MLPAG}
    \end{subfigure}
    \caption{Results of experimental methods on length-based groups of questions.}
    \label{fig:question_length_results}
\end{figure}

For ease of analysis, we divide questions and answers into different groups based on their length, or the total number of tokens they have. Tokens in this Section are achieved in the same way as in Section \ref{sect:dataset} by using VnCoreNLP \cite{vu-etal-2018-vncorenlp} to perform word segmentation. More specifically, we define the groups of questions and answers as in Figure \ref{tab:qa_groups}. We evaluate the baselines and our proposed methods on each group of questions and answers with this definition. At first, we start with examining the results of the baselines and proposed models on groups of questions. Details are given in Figure \ref{fig:question_length_results}.


According to Figure \ref{fig:question_length_results}, most of the classifier-based methods achieve the highest results on group M questions on most metrics. Furthermore, the BLEU scores of the classifier-based methods are relatively low, even lower than 0.2. These scores show that the answers of classifier-based methods poorly match the ground truth answer, so their results are unreliable to interpret and conclude. On the other hand, the generator-based methods achieve the best results on S questions, then gradually decline when the length of questions increases, except for QuMLAG which has the best results on group M questions, and MLPAG which has approximately the same results on group S and M questions then downgrades its score when receiving longer questions. These results indicate that generator-based methods find it easier to cope with the group S and M questions as they are the most-appeared questions in the dataset. However, it also points out that these methods struggle to read and understand long questions in Vietnamese whose linguistic complexity is proportional to the length of respective sentences (Figure \ref{fig:dep_len}).

\begin{figure}[ht]
    \centering
    \begin{subfigure}{0.3\textwidth}
        \includegraphics[width=\textwidth]{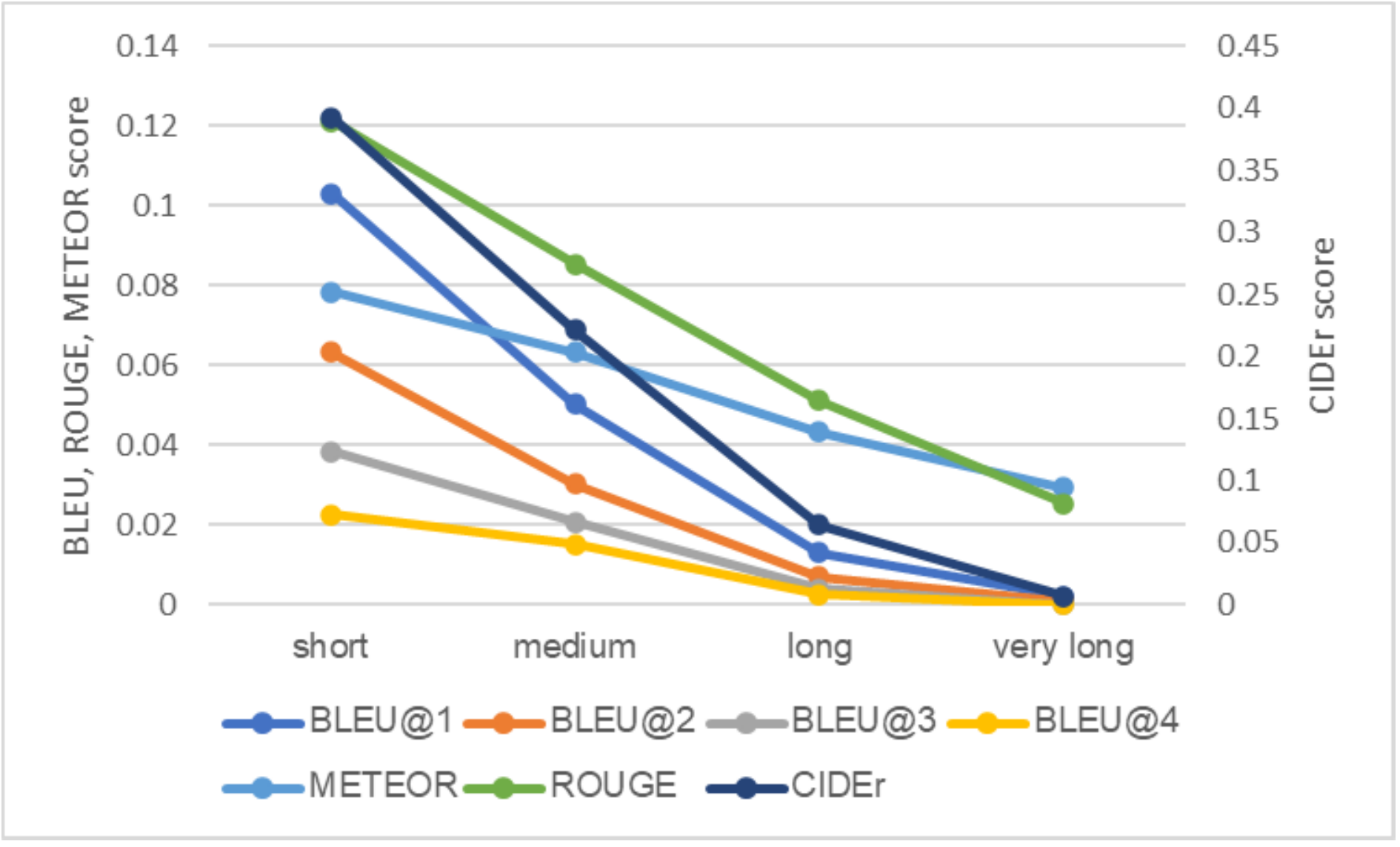}
        \caption{SAAA}
    \end{subfigure}
    \begin{subfigure}{0.3\textwidth}
        \includegraphics[width=\textwidth]{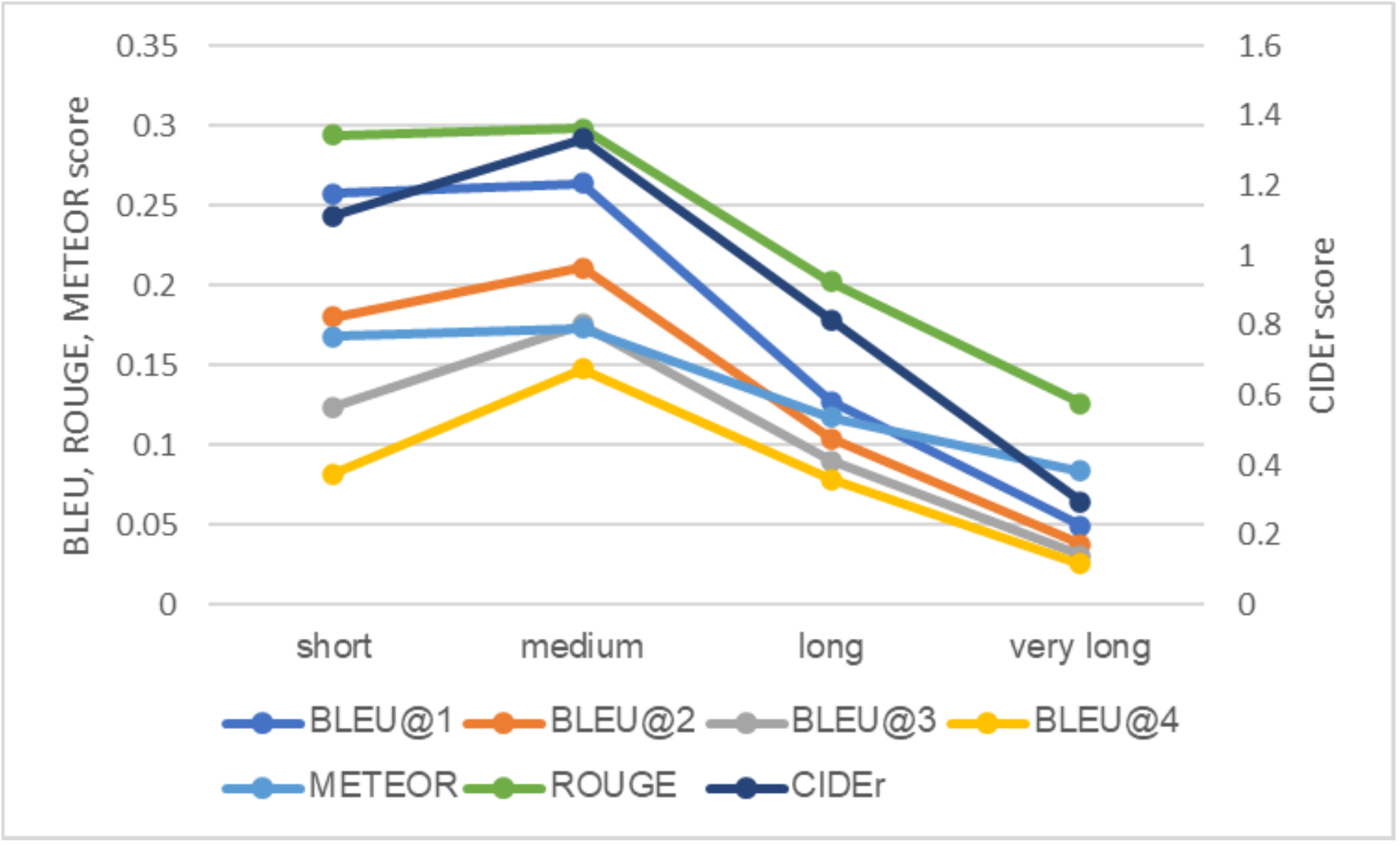}
        \caption{MCAN}
    \end{subfigure}
    \begin{subfigure}{0.3\textwidth}
        \includegraphics[width=\textwidth]{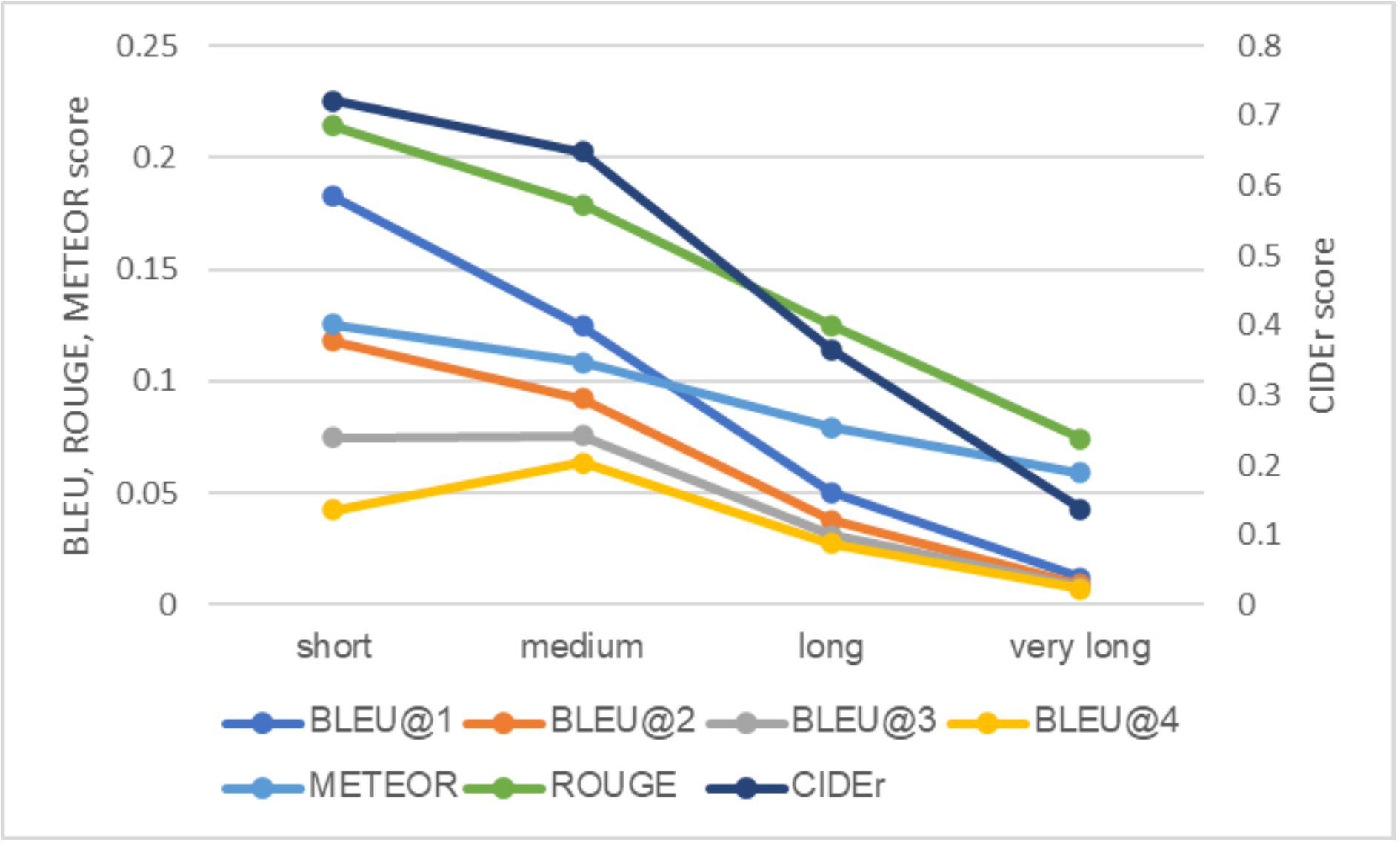}
        \caption{LoRRA}
    \end{subfigure}
    \begin{subfigure}{0.3\textwidth}
        \includegraphics[width=\textwidth]{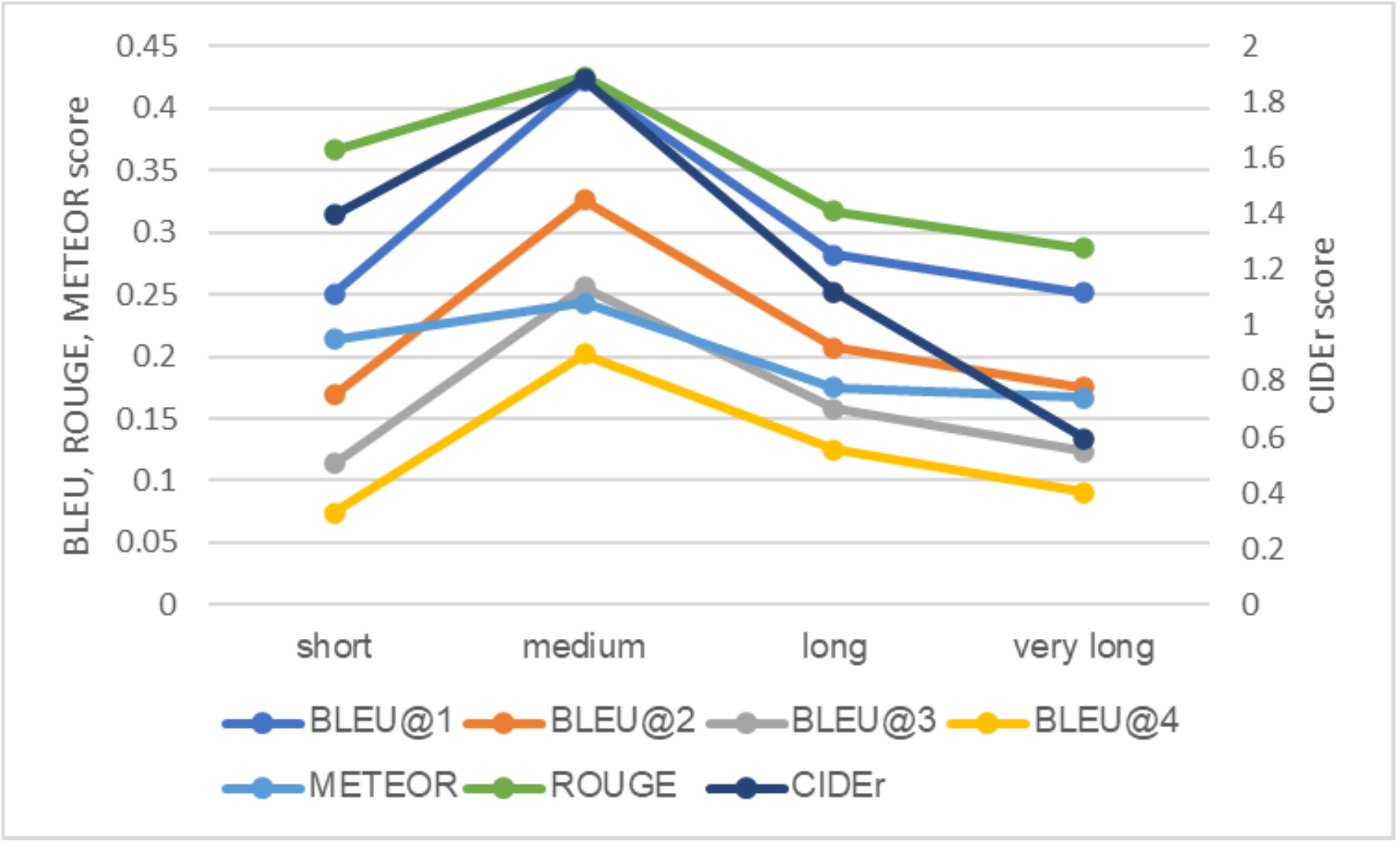}
        \caption{M4C}
    \end{subfigure}
    \begin{subfigure}{0.3\textwidth}
        \includegraphics[width=\textwidth]{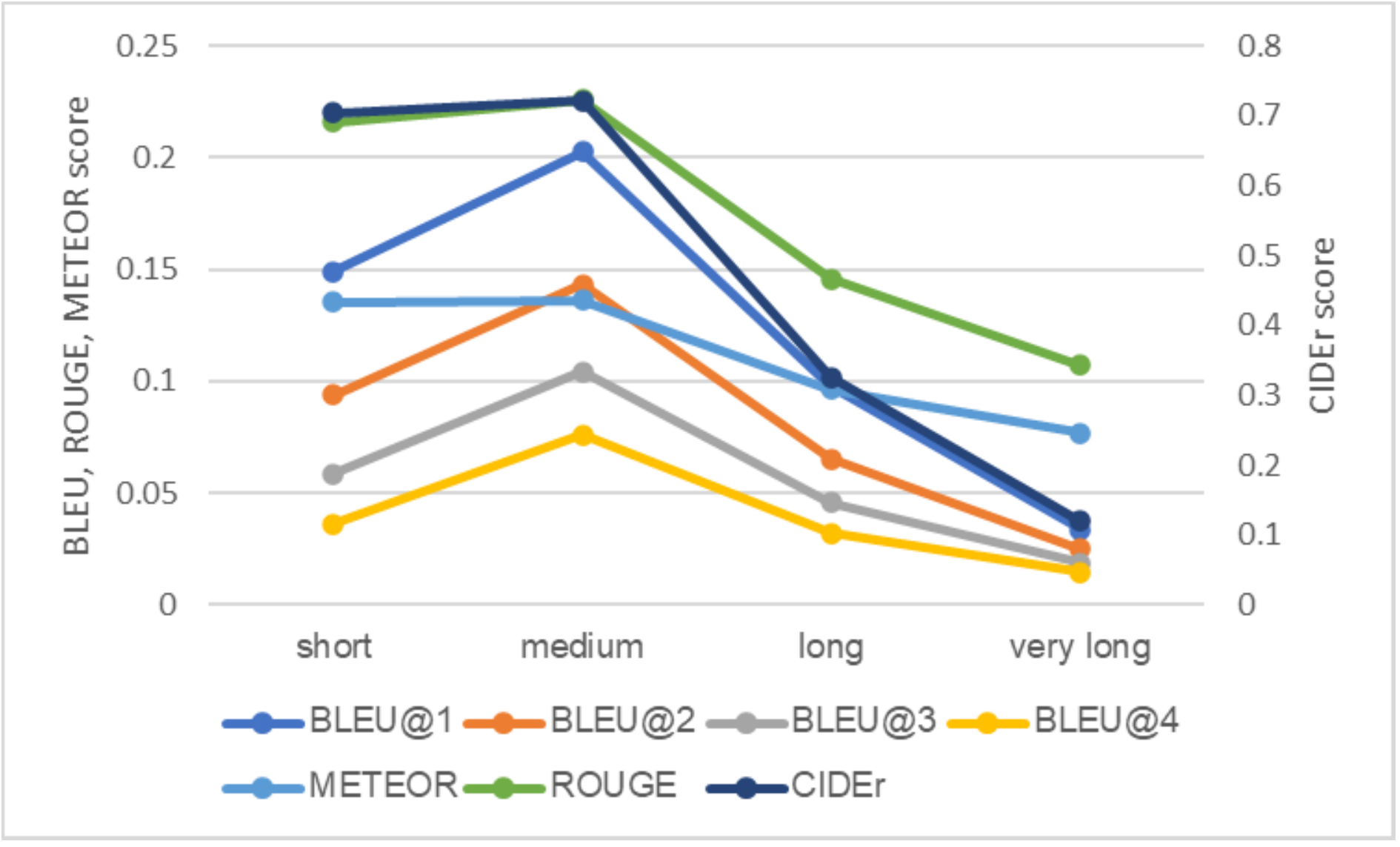}
        \caption{FST}
    \end{subfigure}
    \begin{subfigure}{0.3\textwidth}
        \includegraphics[width=\textwidth]{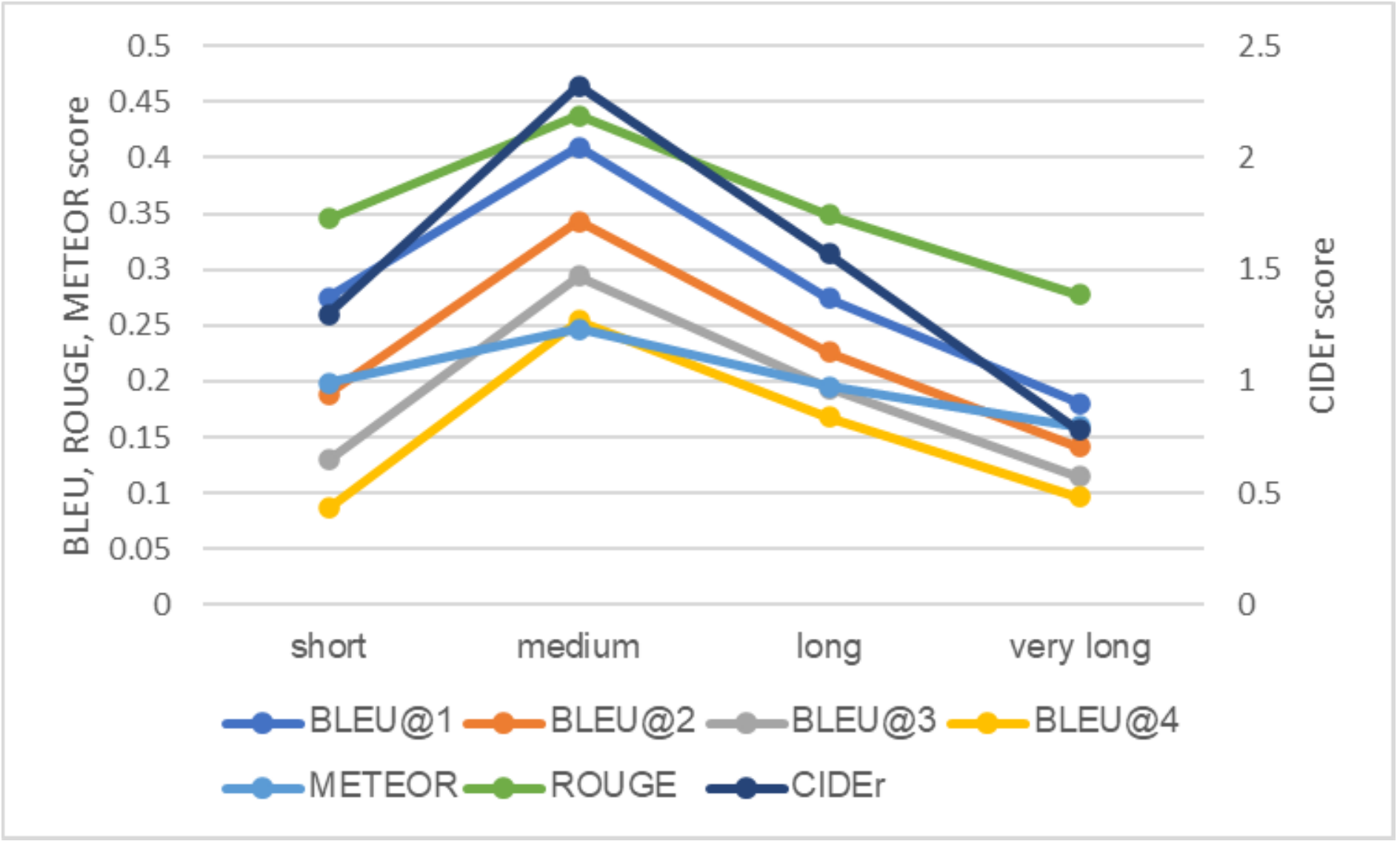}
        \caption{QuMLAG}
    \end{subfigure}
    \begin{subfigure}{0.3\textwidth}
        \includegraphics[width=\textwidth]{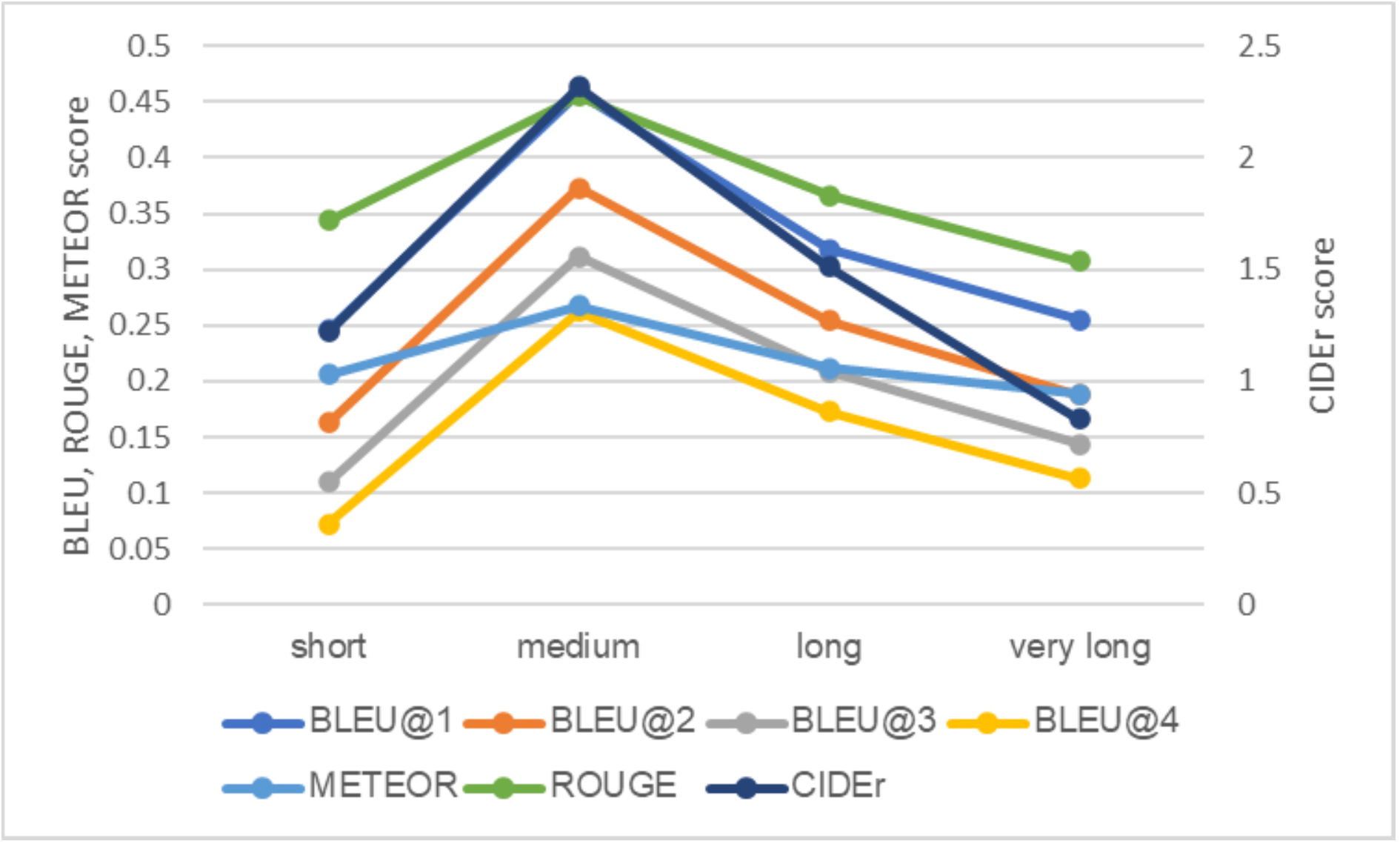}
        \caption{MLPAG}
    \end{subfigure}
    \caption{Results of experimental methods on length-based groups of answers.}
    \label{fig:answer_length_results}
\end{figure}

Turning to the groups of answers, similar to the analysis based on the length of questions, the results of classifier-based methods are low, especially in BLEU@2, BLEU@3, and BLEU@4 scores, which leads to analyzing and interpreting their results being not reliable. Hence, we concentrate on the results of the generator-based methods. According to Figure \ref{fig:answer_length_results}, all generator-based methods share the same pattern of their results on all groups of answers, and all of them achieve the best scores on group M answers. From Figure \ref{fig:datasets-answer-length-statistics}, most of the answers are in group M. This distribution directly influences the results of the methods as they are trained primarily to generate group M answers, leading to the results shown in Figure \ref{fig:answer_length_results}. On the other hand, generator-based methods have difficulty generating longer answers, such as those in groups L and XL. This indicates that these methods suffer hassle while yielding long answers in Vietnamese.

Surprisingly, although the answers of group S occupy a significant proportion in the answer distribution, the results of the generator-based methods on these answers are not as good as on the answers of group M, even though generally group S answers are more straightforward to construct than other groups of answers. To clarify this phenomenon, we consider investigating how the answers of generator-based methods match the ground truth answers. We find that on questions with short answers, the generated answers of generator-based methods have approximately 9.29\% of their tokens repeated from the questions, while this rate is 13\% on questions with medium answers. On questions having long and very long answers, this rate is 13.13\% and 10.87\%, respectively. This implies that on group M answers, generator-based models tend to repeat some first tokens of questions when starting to construct the answers as the way Vietnamese people normally do while answering (e.g. "\textbf{chiếc xe đạp đang được dựng} ở đâu?" (where is the bike leaning against?) $\Rightarrow$ "\textbf{chiếc xe đạp đang được dựng} ở bên cạnh cái giếng" (the bike is leaning against the well)). Accordingly, the generator-based methods easily obtain the highest results on group M answers as they have many matched tokens with the ground truth answers. However, on longer answers, their results drop due to the lack of information in the generated answers compared to human-given answers. In addition, short answers are primarily words or phrases. However, generator-based methods tend to produce group M answers. Such answers construction strategy of generator-based methods not only causes redundant tokens that affect the overall results but may also potentially give wrong information about objects, locations, quantities, colors, or scene texts in answers, hence obtaining low scores on group S answers.

\subsection{Effect of Question Types on Experimental Results}

\begin{figure}[ht]
    \centering
    \begin{subfigure}{0.32\textwidth}
        \includegraphics[width=\textwidth]{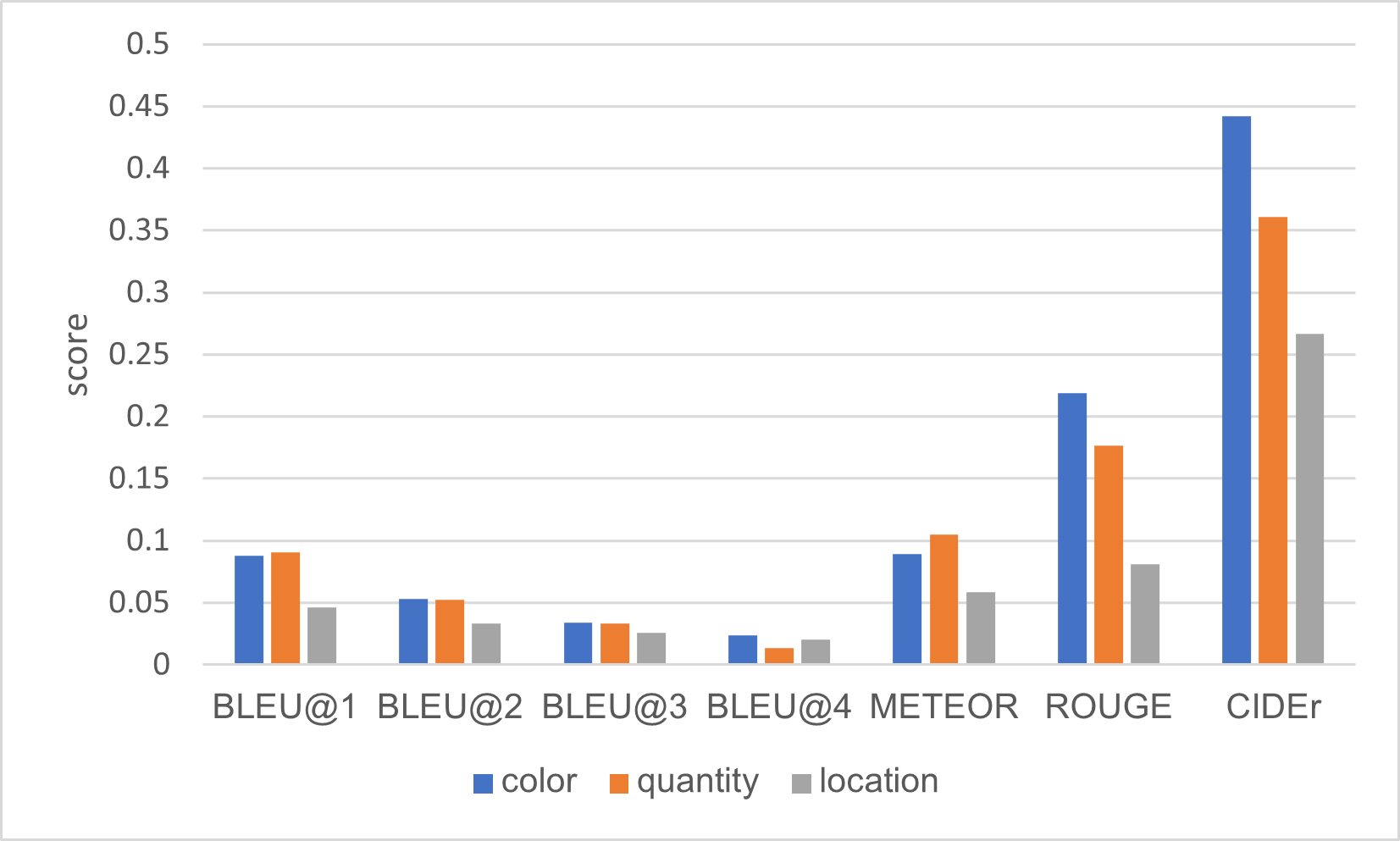}
        \caption{SAAA}
    \end{subfigure}
    \begin{subfigure}{0.32\textwidth}
        \includegraphics[width=\textwidth]{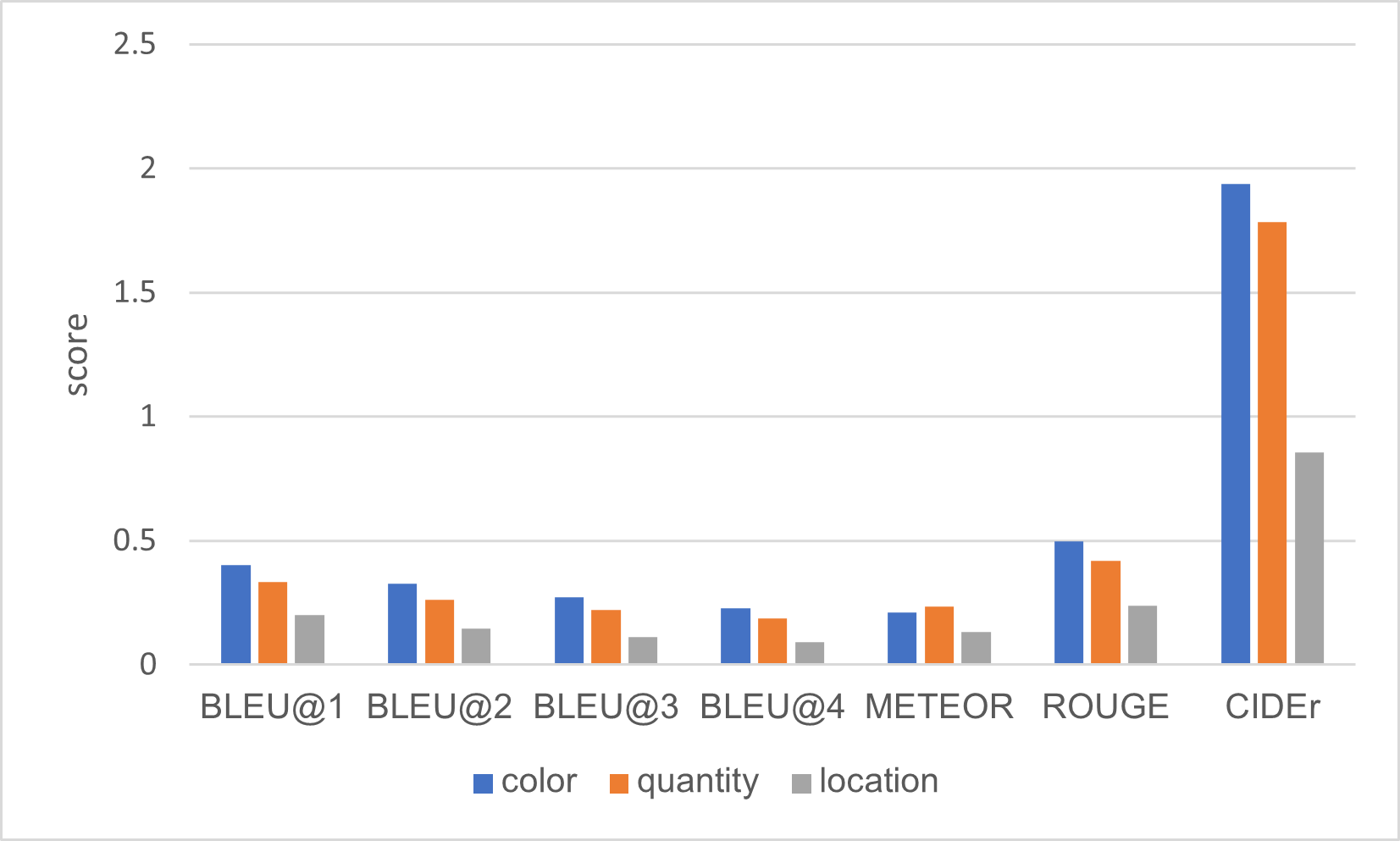}
        \caption{MCAN}
    \end{subfigure}
    \begin{subfigure}{0.32\textwidth}
        \includegraphics[width=\textwidth]{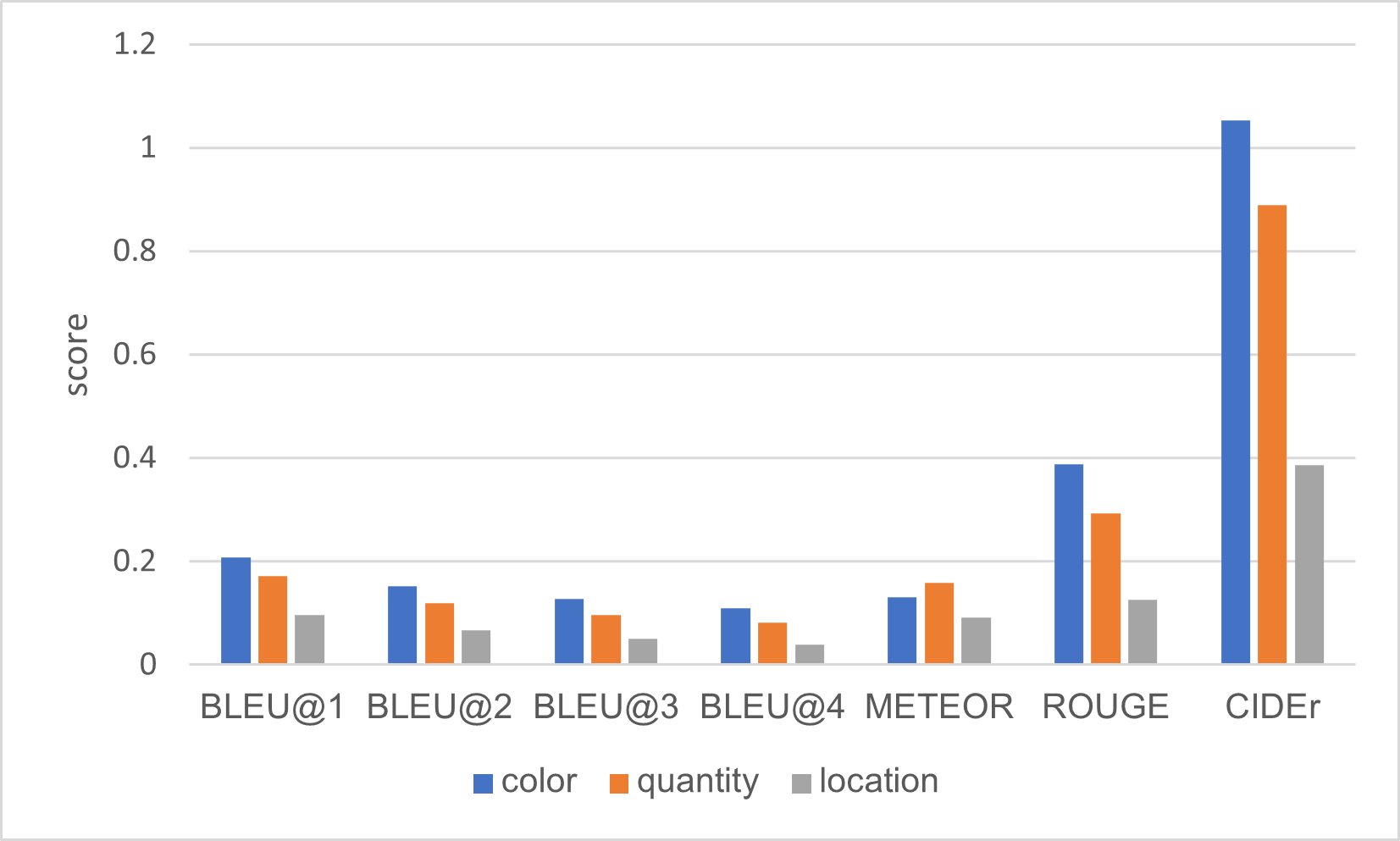}
        \caption{LoRRA}
    \end{subfigure}
    \begin{subfigure}{0.32\textwidth}
        \includegraphics[width=\textwidth]{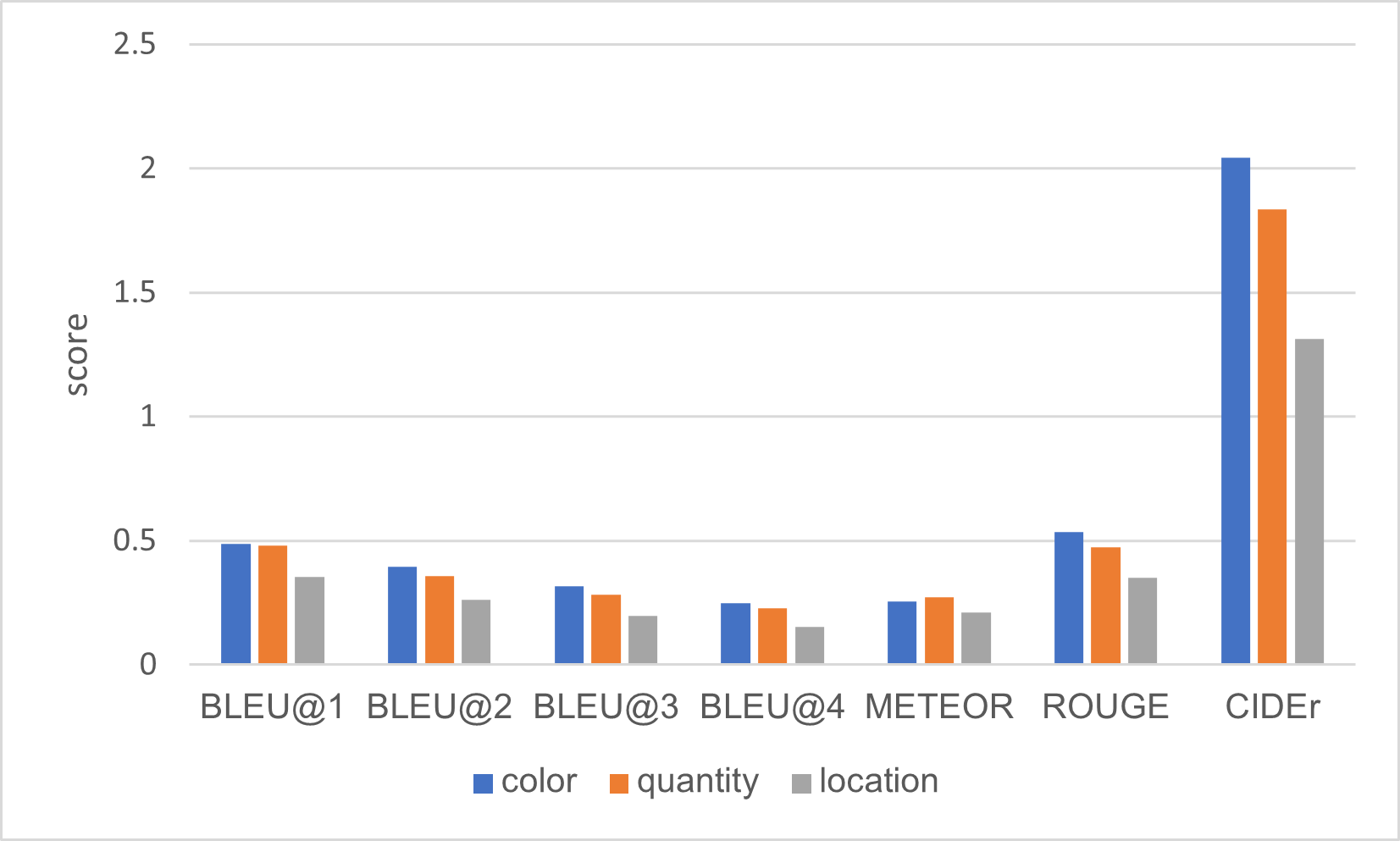}
        \caption{M4C}
    \end{subfigure}
    \begin{subfigure}{0.32\textwidth}
        \includegraphics[width=\textwidth]{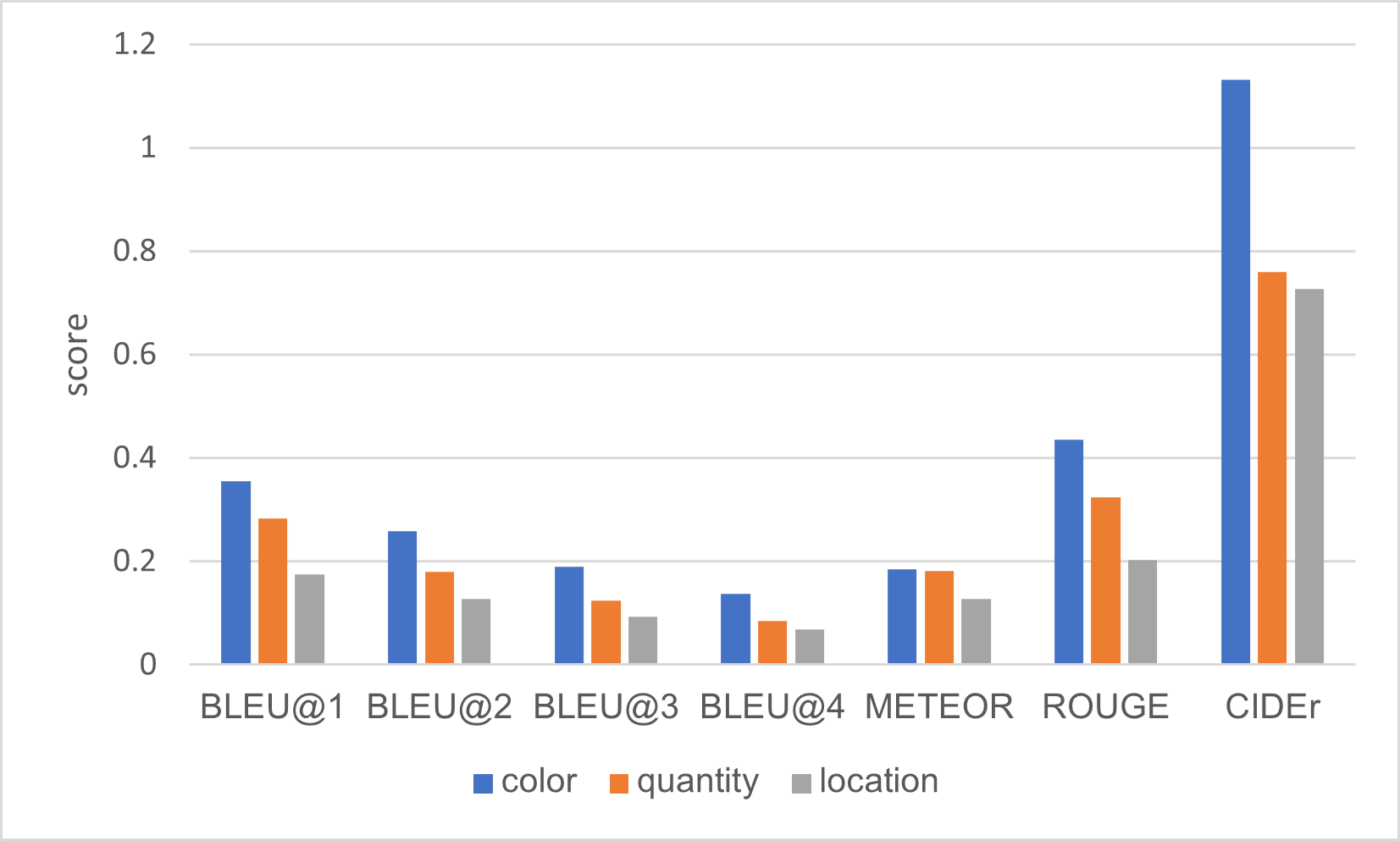}
        \caption{FST}
    \end{subfigure}
    \begin{subfigure}{0.32\textwidth}
        \includegraphics[width=\textwidth]{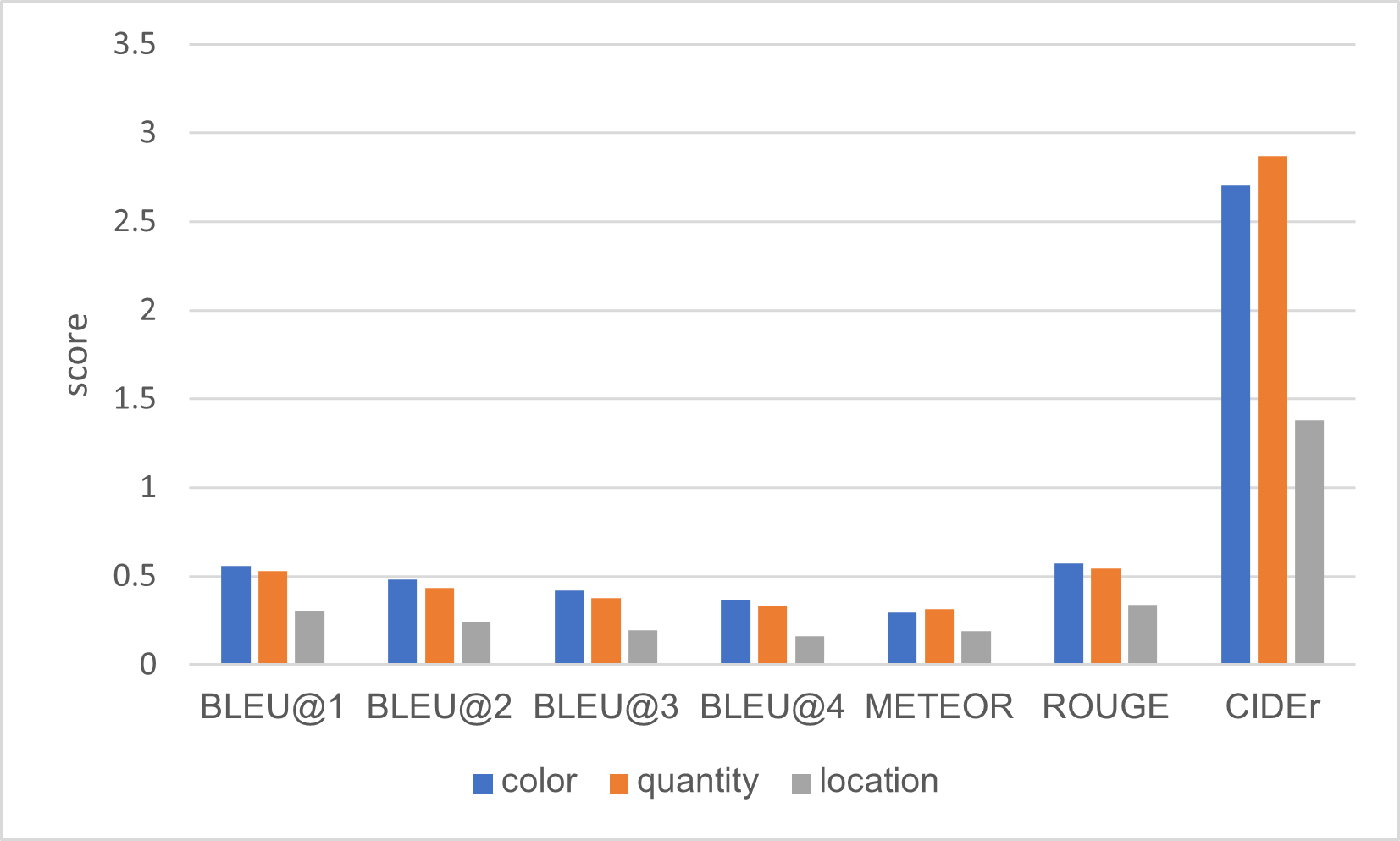}
        \caption{QuMLAG}
    \end{subfigure}
    \begin{subfigure}{0.32\textwidth}
        \includegraphics[width=\textwidth]{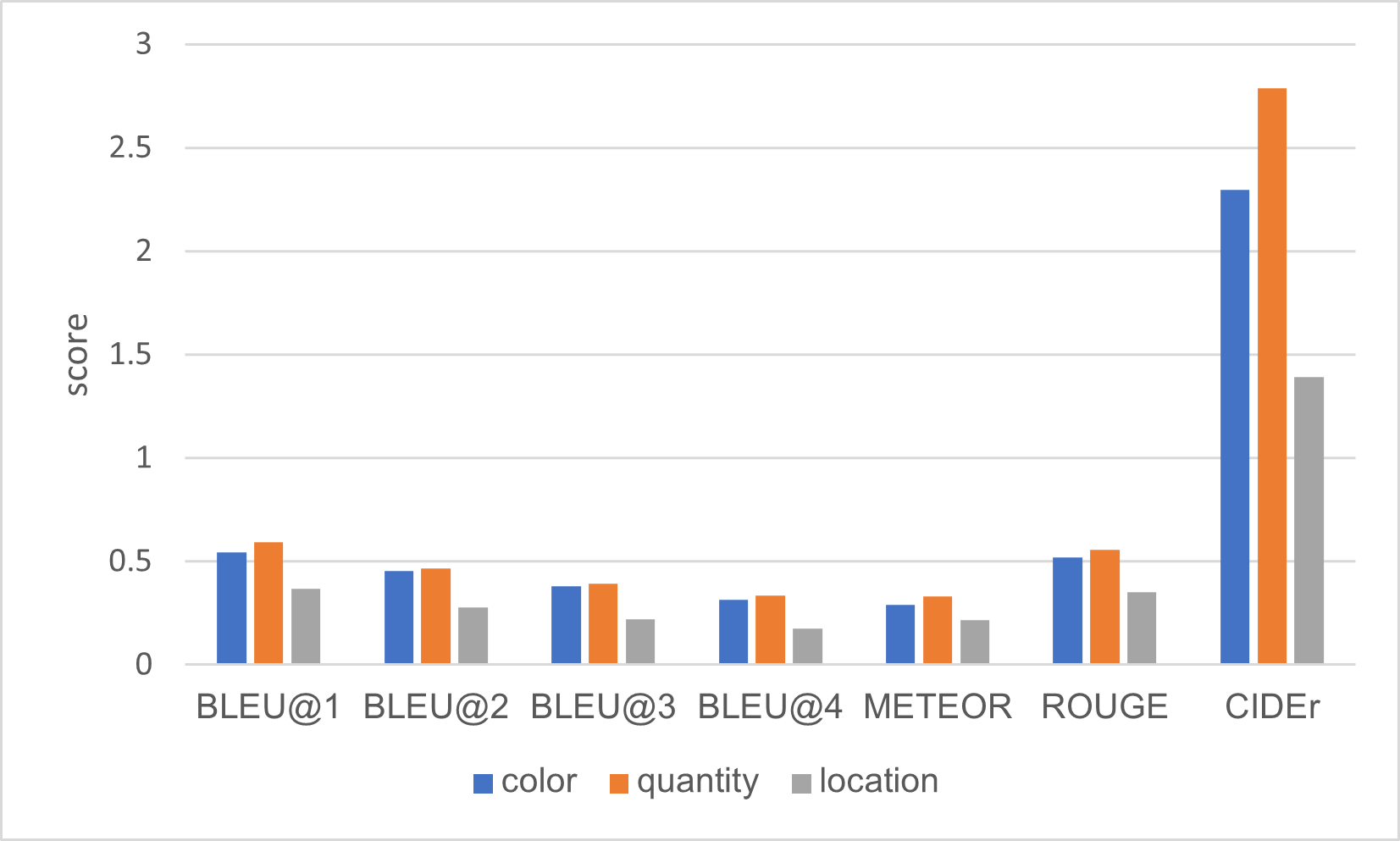}
        \caption{MLPAG}
    \end{subfigure}
    \caption{Results on question types for each experimental method.}
    \label{fig:types_per_models_results}
\end{figure}

In order to analyze how models perform on 
different targeted contents of questions, here we examine the predictions for questions on several fairly variated aspects such as colors, quantities, and locations. We define the type of question based on such aspects that questions may exploit. This analysis monitors how models may pay attention to visual details and corporate descriptive words into the desired answers. We first use regular expressions to prepare rule-based algorithms for question-type classification. For each type of question, we calculate the evaluation scores for every experimental method just like in the above sections. As illustrated in Figure \ref{fig:types_per_models_results}, all models elicit the same pattern that the questions on locations are more challenging than those on colors and quantities. This may be drawn from the fact that location information is more complicated, unlike colors and quantities which can be interpreted more simply.

\begin{figure}[ht]
    \centering
    \begin{subfigure}{0.47\textwidth}
        \includegraphics[width=\textwidth]{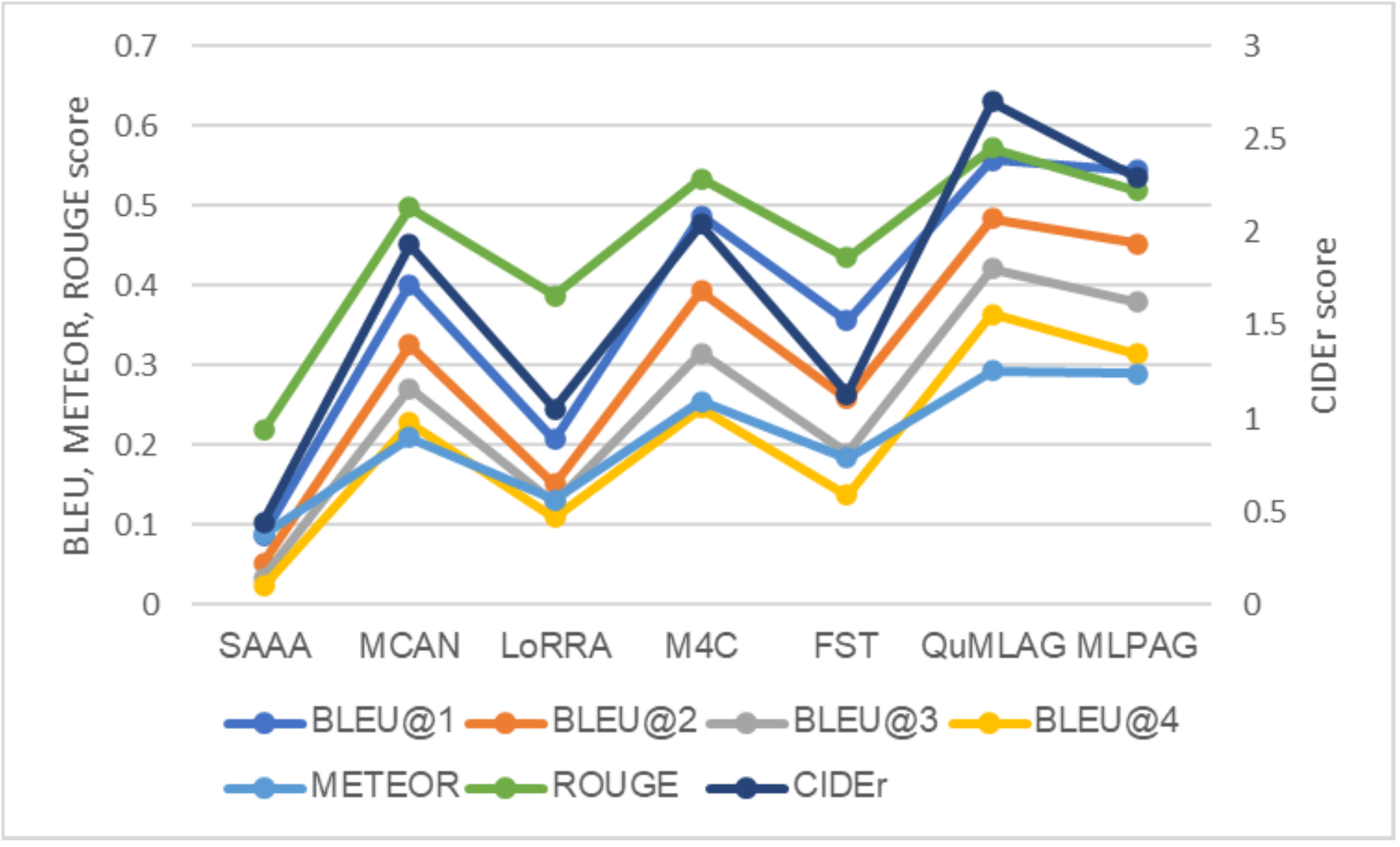}
        \caption{Colors}
    \end{subfigure}
    \begin{subfigure}{0.47\textwidth}
        \includegraphics[width=\textwidth]{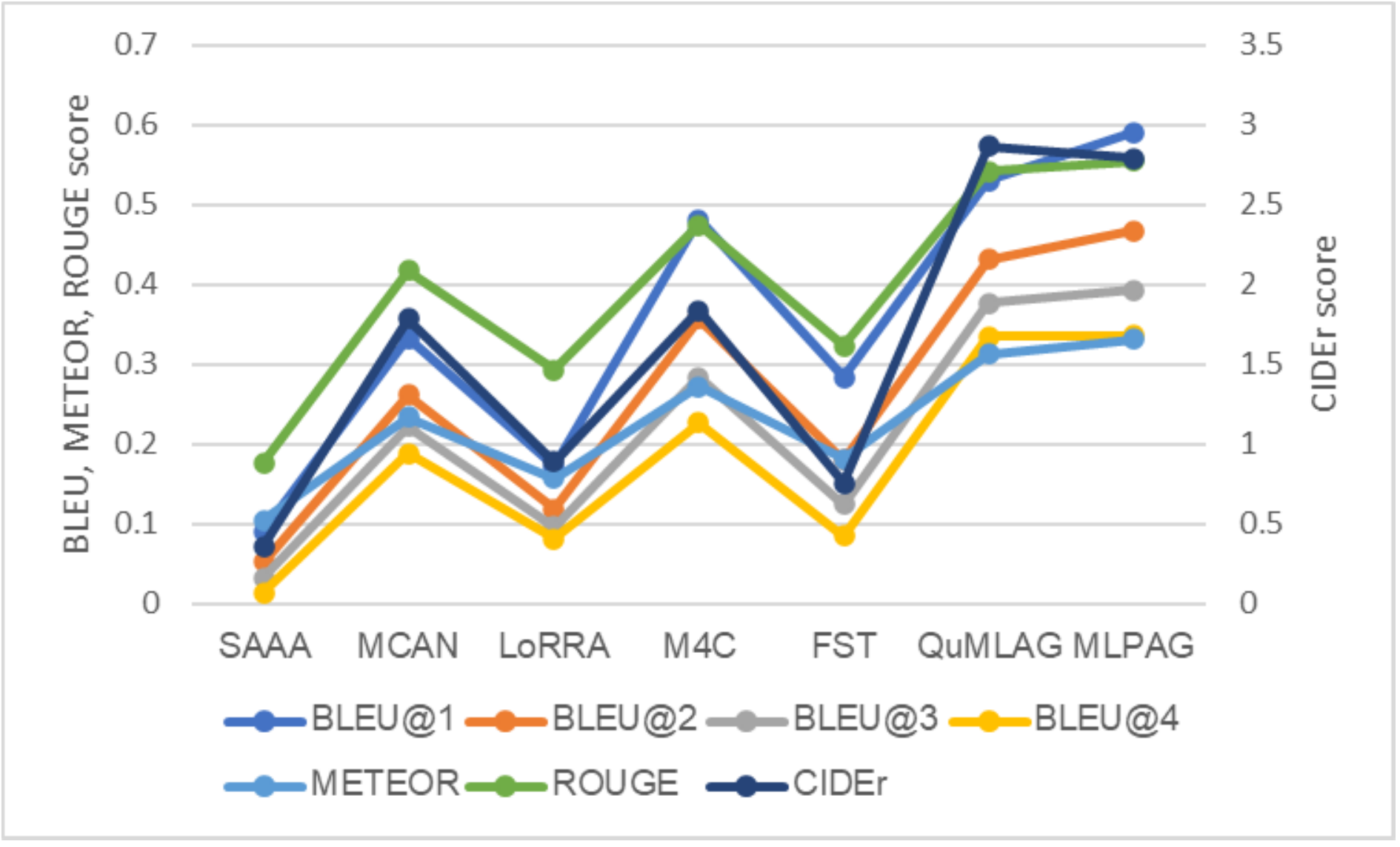}
        \caption{Quantities}
    \end{subfigure}
    \begin{subfigure}{0.47\textwidth}
        \includegraphics[width=\textwidth]{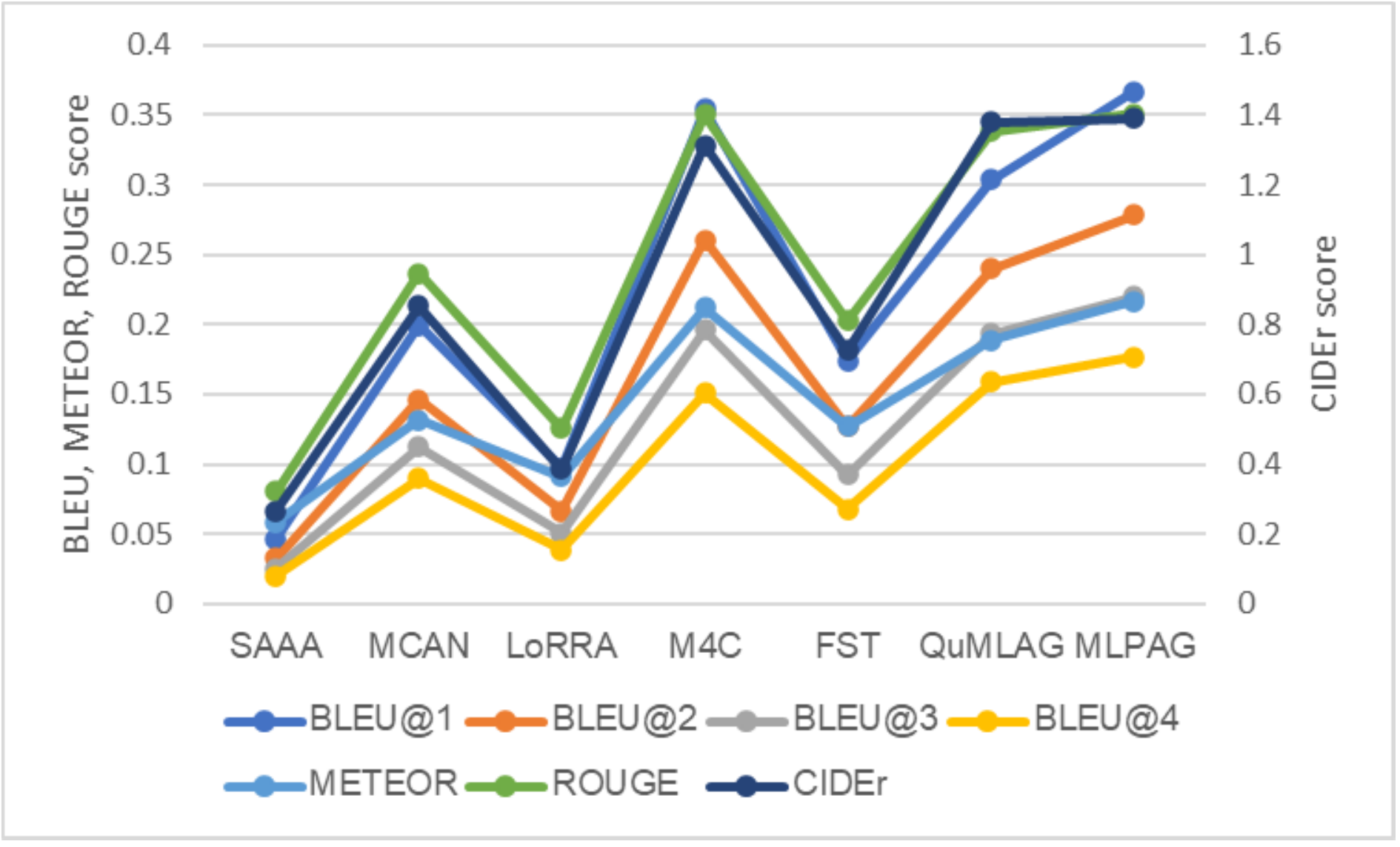}
        \caption{Locations}
    \end{subfigure}
    \caption{Results of experimental methods on each question type.}
    \label{fig:models_per_type_results}
\end{figure}

In Figure \ref{fig:models_per_type_results}, we compare the scores of the experimental methods for each type of question.
In particular, more sophisticated models based on transformer architecture perform better in describing the details required to answer these types of questions. On the overall pattern, the models show a similar order of performance for each type where the generative alternatives achieve higher scores than the original models. However, these findings are proportional to the main result in Section \ref{sect:main_result} and may blend in the overall performance of experimental methods on giving the whole answers, which do not describe how correctly the models produce the exact words to interpret colors, quantities, or locations in the answers for those types of question.

To analyze the correctness of experimental methods in giving answers for each type of question, we suggest using accuracy, as defined as follows:

\begin{align}
    {Accuracy}_{type} = \frac{{Number\_of\_correct\_terms}_{type}}{{Total\_questions}_{type}}
    \label{eq:accuracy}
\end{align}

\begin{figure}[!ht]
    \centering
    \includegraphics[width=0.7\textwidth]{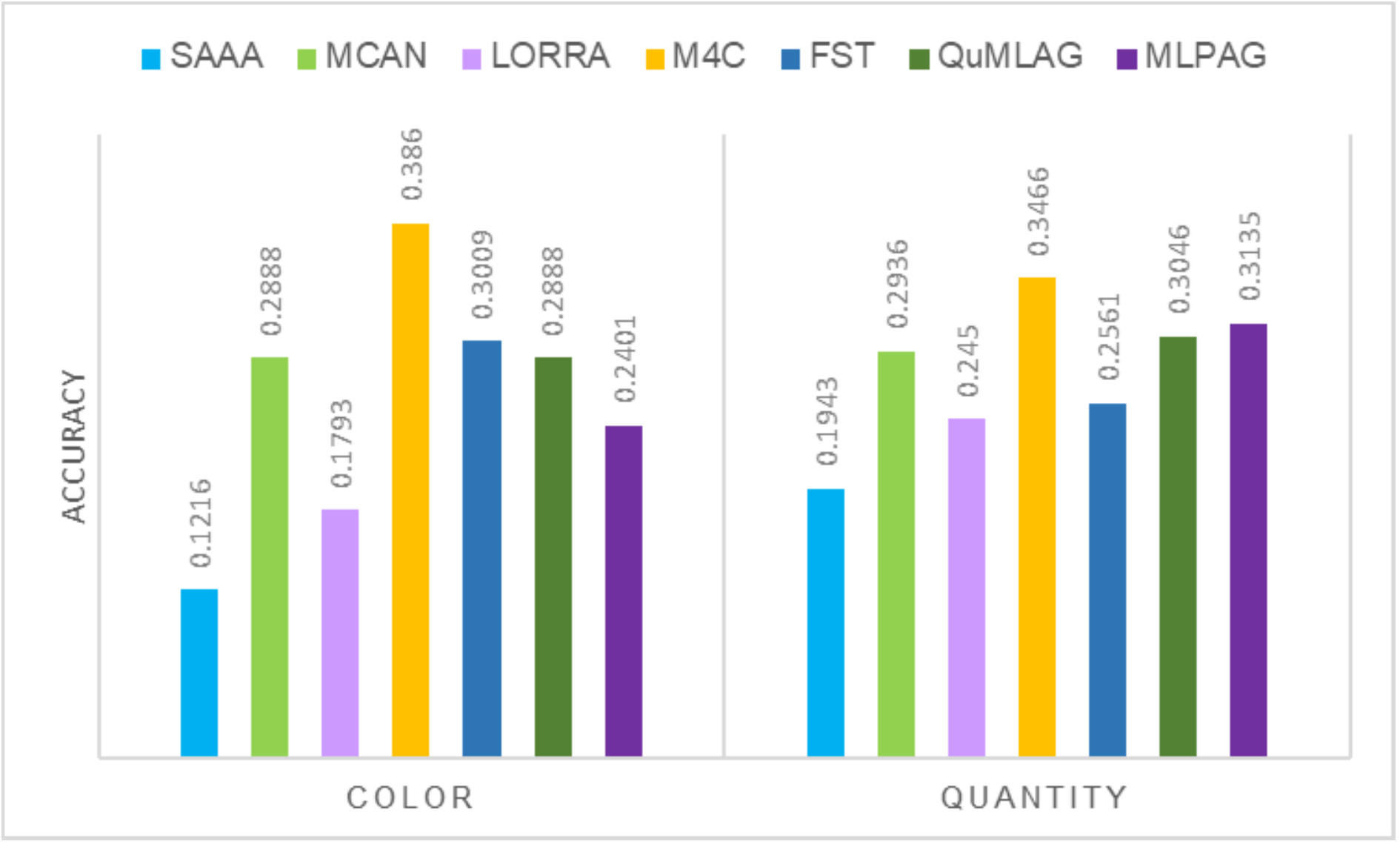}
    \caption{Accuracy of experimental methods on giving the correct term in prediction for each type of question.}
    \label{fig:accuracy_results}
\end{figure}

We used this metric to measure the percentage of giving the right terms as in the gold answers. For questions on colors, we track the color word in the prediction and compare it with the one in the corresponding gold answer. We also do the same for questions on quantities but not for locations since this factor is highly variable in the text presentation. Therefore, it is more feasible to measure the correctness of color and quantity predictions, which are enough to observe how well models may achieve at attending to the visual properties of objects. The plots in Figure \ref{fig:accuracy_results} show that M4C outperforms others in color and quantity predictions. The inter- and intra- modality relations modeling of M4C \cite{hu2020iterative} allows it to effectively associate visual information along with the collocation between the words in questions and answers, which facilitates itself with visual and textual information in the prediction of details such as colors and quantities. On the other hand, the generative renovation of baseline methods yields substantial accuracy increment on predictions of such terms. This suggests that when modified to formulate longer answers, the experimental methods tend to use these terms more correctly in the answers, which proves that the attentive generator module plays an important role in reinforcing the quality and details of the predictions in the final stage.

\subsection{Effect of Scene Text Images on Experimental Results}

Logically M4C and MLPAG were provided with Dynamic Pointer Network \cite{hu2020iterative} in their decoder module, or in other words, they similarly have the ability to use scene texts in their answers so they must have better results compared to non-read ability methods such as SAAA, MCAN, FST, and QuMLAG. However, from Table \ref{tab:main_results} the actual results of M4C and MLPAG were even lower than those of QuMLAG which is not able to read and use scene texts, especially MLPAG even yielded better results than M4C. To analyze these results, we first observed the difference in answer features among VQA datasets. According to Figure \ref{fig:datasets-answer-length-statistics}, the TextVQA dataset although was claimed to have open-ended questions and answers, its answers distribution shares the same characteristic as other Non-text QA VQA datasets such as VQAv2, which means most answers in TextVQA have lengths of 1, 2 and 3. Or, simply speaking, in the TextVQA dataset, scene texts are copied from images to be the answers. Therefore M4C method, when trained on the TextVQA dataset, tends to learn how to copy scene texts in images to use them as answers rather than using scene texts to provide detailed information in answers. When being trained on the OpenViVQA dataset, open-ended answers require the method must have the ability to decide which available scene texts it will "copy" and where is the most suitable position it will "paste" in the answers. Hence the complexity is increased, and the copying mechanism implemented by Dynamic Pointer Network does not work as expected.

\begin{figure}
    \centering
    \includegraphics[width=\textwidth]{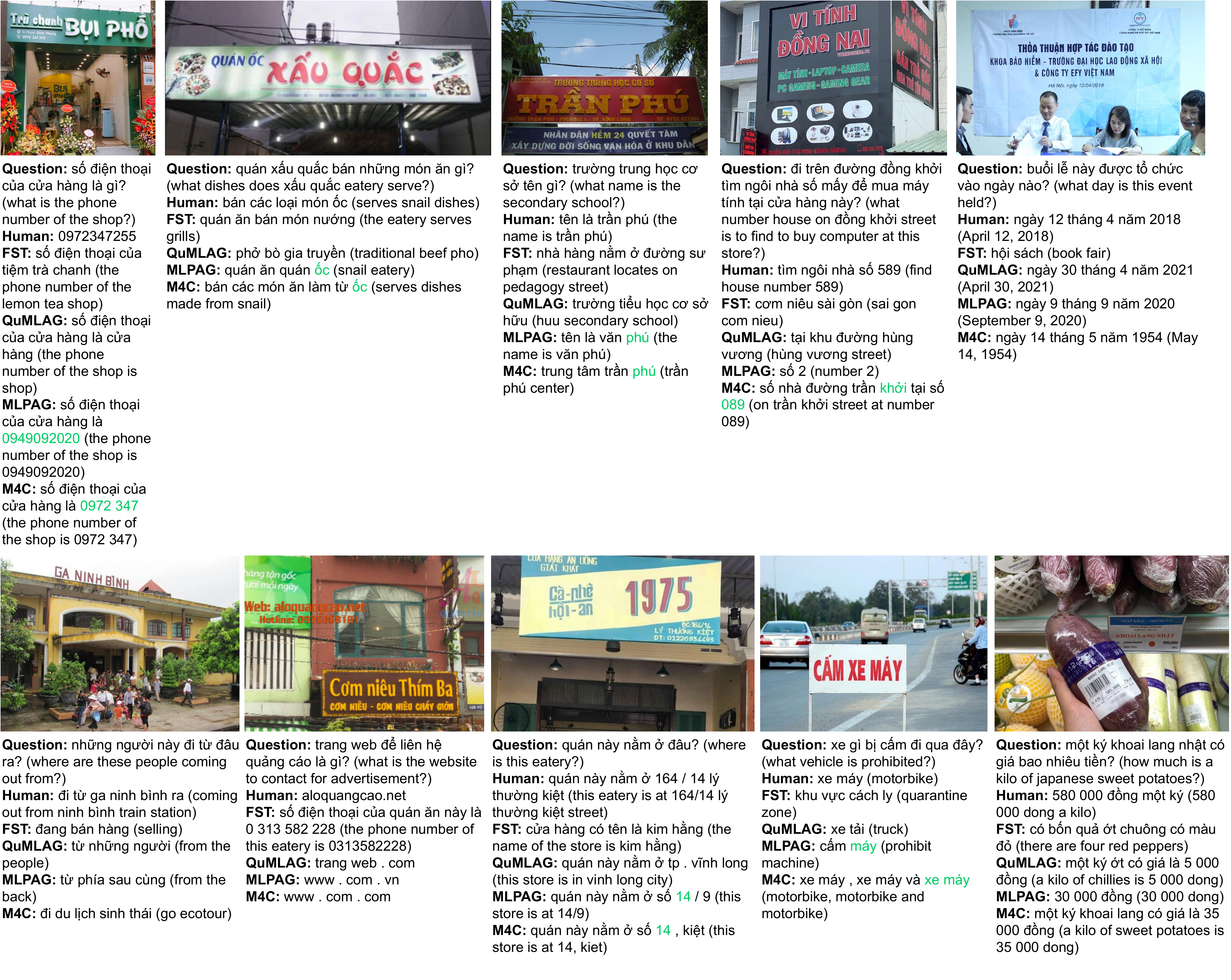}
    \caption{Several examples of our three proposed methods and M4C. Tokens in green are copied from scene texts in the images.}
    \label{fig:scene_text_examples}
\end{figure}

For non-reading architecture such as FST and QuMLAG, from Figure \ref{fig:scene_text_examples}, these two models obviously can not read scene text in images. Hence, they produced answers irrelevant to the given questions and images. In addition, MLPAG and M4C share the same behavior. According to Figure \ref{fig:scene_text_examples}, these two models did not work well in using scene text in answers. The frequency of using scene texts is low, and they usually use them at the end of answers. These results prove such Dynamic Pointer Network \cite{hu2020iterative} proposed for M4C is adaptable for simple answers as in the TextVQA dataset where the whole answers are scene texts copied from images in usual natural language, this module can not tackle well how to use scene text to provide additional information for given questions. We need a better solution for this challenge of the OpenViVQA dataset.

\subsection{Why are QuMLAG and MLPAG better than M4C?} \label{sec:m4c_result}

\begin{table}[ht]
\resizebox{\textwidth}{!}{
\begin{tabular}{lclllllll}
\hline
\multicolumn{2}{l}{}                  & \multicolumn{1}{c}{\textbf{BLEU@1}} & \multicolumn{1}{c}{\textbf{BLEU@2}} & \multicolumn{1}{c}{\textbf{BLEU@3}} & \multicolumn{1}{c}{\textbf{BLEU@4}} & \multicolumn{1}{c}{\textbf{METEOR}} & \multicolumn{1}{c}{\textbf{ROUGE}} & \multicolumn{1}{c}{\textbf{CIDEr}} \\ \hline
\multirow{4}{*}{Text-QA}     & FST  & 0.1020                              & 0.0674                              & 0.0445                              & 0.0302                              & 0.0809                              & 0.1331                             & 0.3566                             \\
                             & QuMLAG  & 0.3064                              & 0.2408                              & 0.1963                              & 0.1649                              & 0.1845                              & 0.3222                             & 1.2504                             \\
                             & MLPAG & \textbf{0.3589}                     & \textbf{0.2710}                     & \textbf{0.2115}                     & \textbf{0.1701}                     & \textbf{0.2075}                     & \textbf{0.3604}                    & \textbf{1.3758}                    \\
                             & M4C    & 0.3315                              & 0.2375                              & 0.1730                              & 0.1303                              & 0.1962                              & 0.3547                             & 1.3191                             \\ \hline
\multirow{4}{*}{Non-text QA} & FST  & 0.2514                              & 0.1743                              & 0.1259                              & 0.0909                              & 0.1559                              & 0.2747                             & 0.8638                             \\
                             & QuMLAG  & 0.4090                              & 0.3371                              & 0.2870                              & \textbf{0.2478}                     & 0.2408                              & \textbf{0.4354}                    & \textbf{2.1359}                    \\
                             & MLPAG & \textbf{0.4464}                     & \textbf{0.3565}                     & \textbf{0.2942}                     & 0.2449                              & \textbf{0.2534}                     & 0.4184                             & 1.9377                             \\
                             & M4C    & 0.3960                              & 0.3001                              & 0.2341                              & 0.1838                              & 0.2252                              & 0.4025                             & 1.6545                             \\ \hline
\end{tabular}}
\caption{Comparison of the three proposed methods and M4C between Text-QA and Non-text QA.}
\label{tab:text_non_text_qa}
\end{table}

From Table \ref{tab:main_results}, we saw the abnormal results that M4C, despite its better results on English datasets such as TextVQA and OCR-VQA, has lower results than QuMLAG and MLPAG. As reported in Table \ref{tab:text_non_text_qa}, on Text QAs, the MLPAG achieved the best results on all metrics while the M4C is lower. However, these two models share the same architecture in the Answer Generator module, and on several metrics (BLEU@2, BLEU@3, BLEU@4), M4C is lower than QuMLAG. On Non-text QA, MLPAG has better results than M4C on almost metrics. In addition, QuMLAG achieved the best results on BLEU@4, ROUGE, and CIDEr, and it leads MLPAG as well as M4C a far distance on CIDEr. Carefully observing, in both cases of Text-QA and Non-text QA in Table \ref{tab:text_non_text_qa}, as well as overall results in Table \ref{tab:main_results}, MLPAG and QuMLAG which share the same color that they have a smaller size (in terms of total parameters) achieved better results than M4C which is more extensive. This implies that on the OpenViVQA dataset, large-size models such as M4C are not effective; hence we need a method having better ideas to effectively tackle the OpenViVQA dataset rather than depending on the power of large models and computing resources.

\section{Conclusion and Future Work}
\label{sect:conclusion}

In this paper, we introduced the first high-quality and large-scale VQA benchmark in Vietnamese, then defined a novel and challenged open-ended VQA task for Vietnamese. Through our experiments, we showed the complexity of the linguistic aspects as well as the requirement of reading and using scene texts from images flexibly and effectively in answers to the OpenViVQA dataset so challenged that SOTA methods on VQAv2, OCR-VQA, and TextVQA datasets yielded downgraded results on this dataset. Our proposed methods FST, QuMLAG, and MLPAG in contrast preliminarily achieved better results when tackling the VQA task as an answers generator task, hence proving that former VQA approaches are not effective on our novel form of VQA. Moreover, although MLPAG is a sort of single-hop attention method, it achieved competitive results compared to QuMLAG (Table \ref{tab:main_results}) as a multi-hop attention thanks to its reading and using scene text ability (Table \ref{tab:text_non_text_qa}). We will address the drawbacks of these preliminary methods and then propose a better solution to effectively tackle the challenges from the OpenViVQA dataset.

Moreover, the OpenViVQA dataset in spite of being the largest VQA dataset in Vietnamese, its size is relatively small compared to English VQA datasets hence large vision-language models such as M4C did not reach the best performance as on English VQA dataset. We plan to widen the OpenViVQA dataset in terms of images and QAs in our next study as well as construct more answers for a question so that we can evaluate open-ended answers comprehensively. Moreover, inspired by Changpiny et al. \cite{changpinyo2022towards}, we are going to expand the OpenViVQA dataset to a multilingual VQA dataset, thus providing a high-quality resource for researching multilingual VQA including Vietnamese. The OpenViVQA dataset can be used as one of the high-quality resources for its manual annotation to evaluate pre-trained vision-language models, especially in Vietnamese.

\section{Several examples of well-known VQA datasets.}
\label{appendix-A}

\begin{figure}[ht]
    \centering
    \begin{subfigure}{0.47\textwidth}
        \includegraphics[width=\textwidth,height=1.25\textwidth]{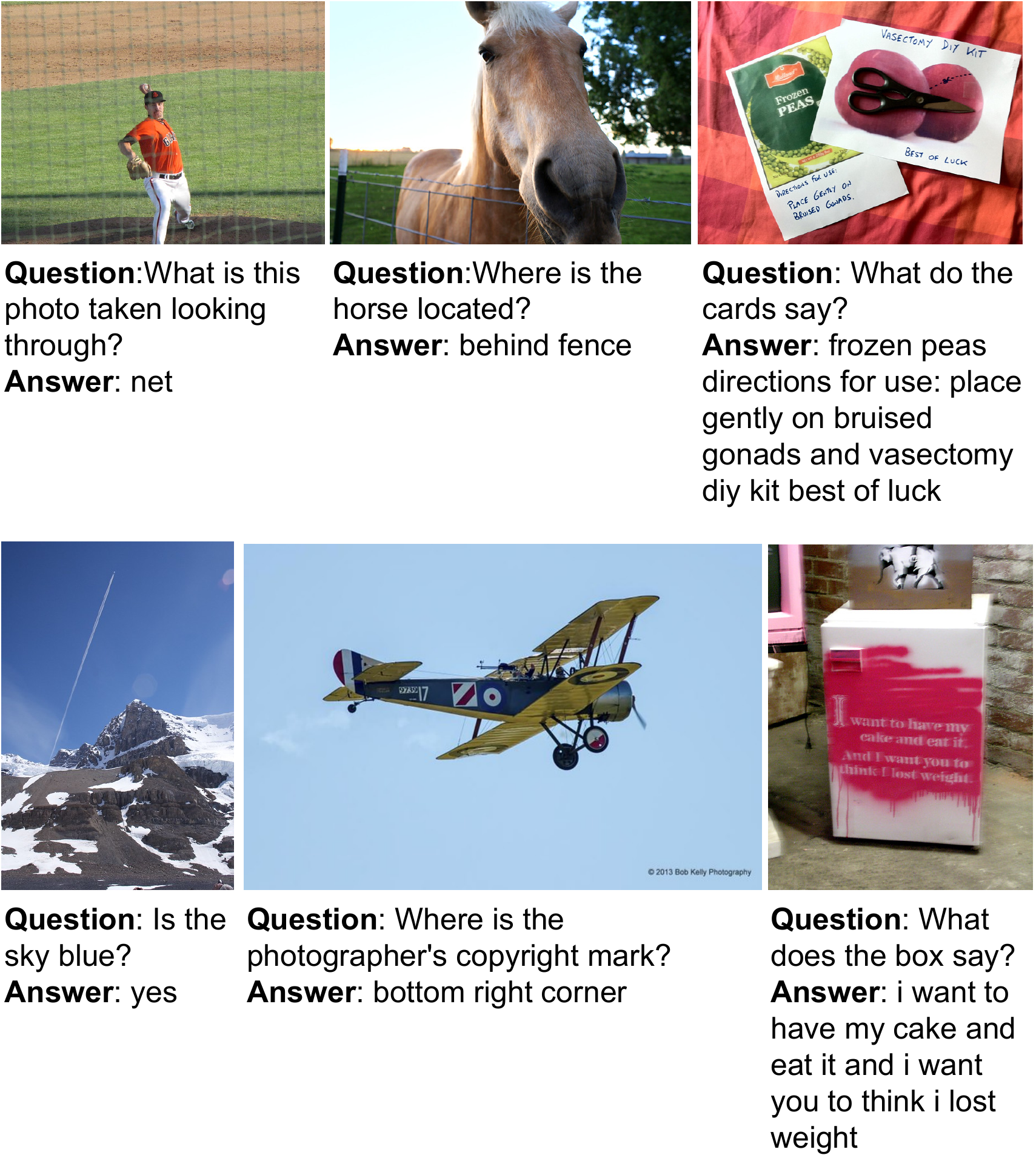}
        \caption{VQAv2}
    \end{subfigure}
    \begin{subfigure}{0.47\textwidth}
        \includegraphics[width=\textwidth,height=1.25\textwidth]{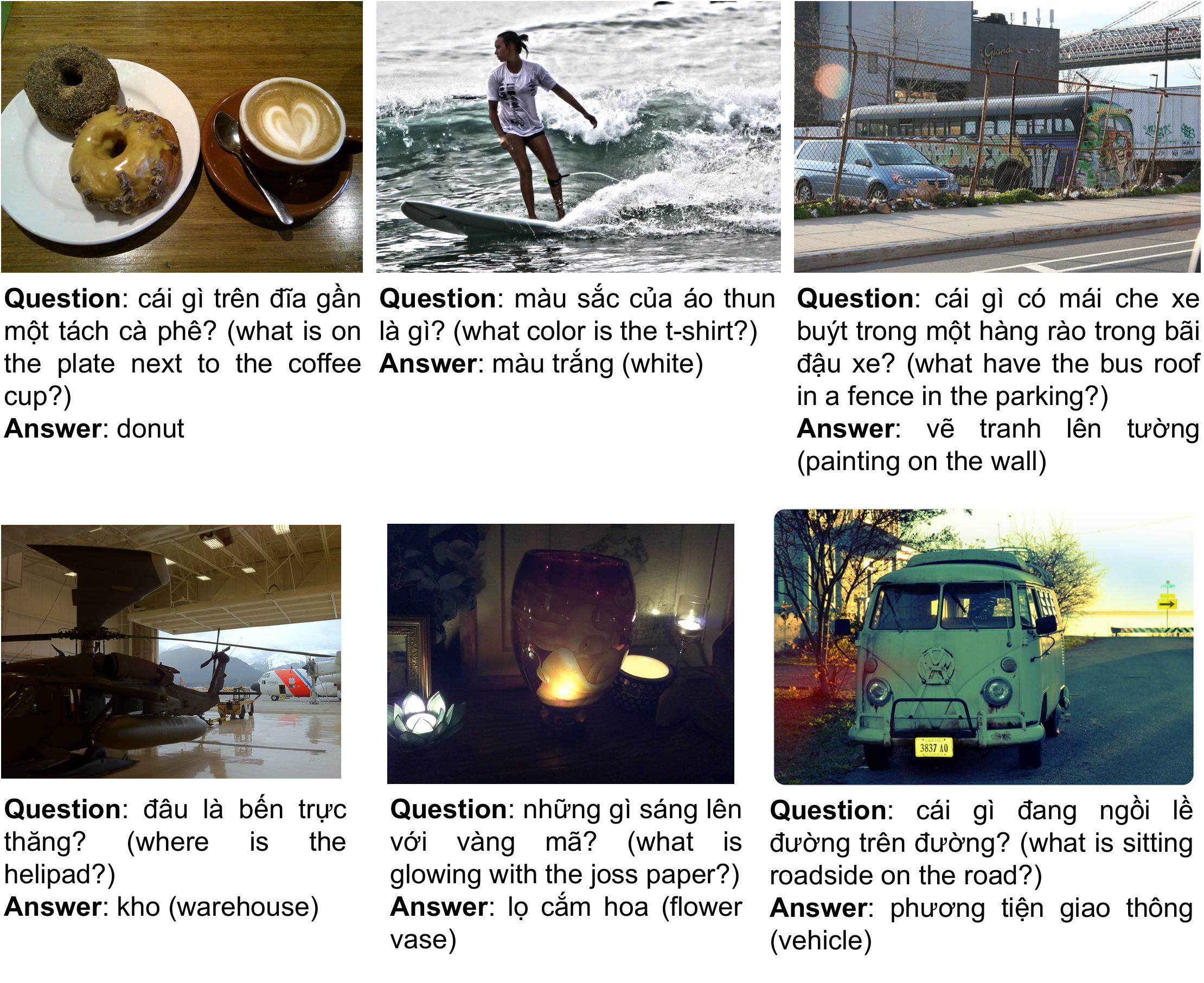}
        \caption{ViVQA}
    \end{subfigure}
    \begin{subfigure}{0.47\textwidth}
        \includegraphics[width=\textwidth,height=1.25\textwidth]{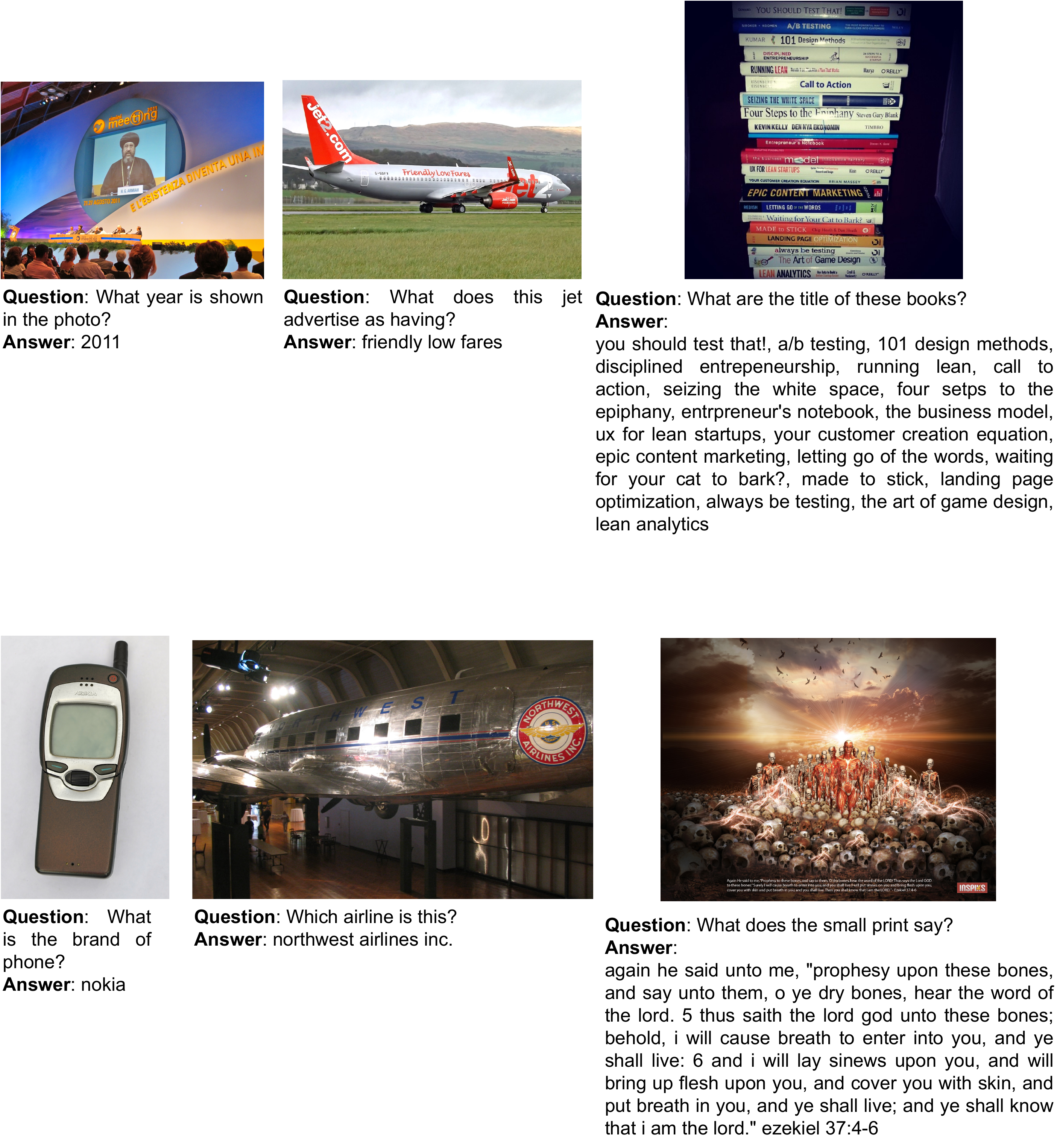}
        \caption{TextVQA}
    \end{subfigure}
    \begin{subfigure}{0.47\textwidth}
        \includegraphics[width=\textwidth,height=1.25\textwidth]{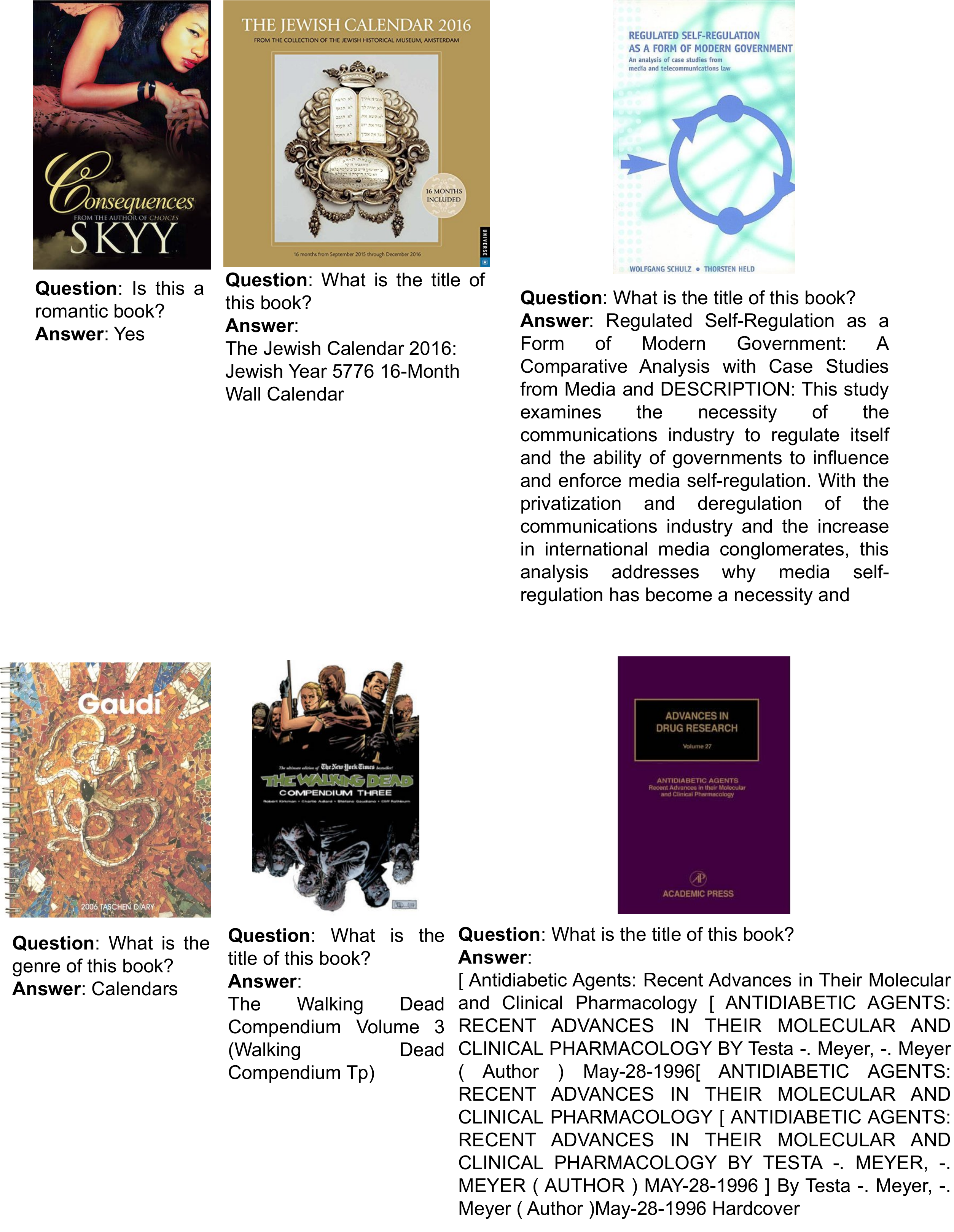}
        \caption{OCR-VQA}
    \end{subfigure}
    \caption{Examples of well-known VQA datasets. Three continuous columns represent questions with the shortest, medium, and longest (in terms of length), respectively.}
    \label{fig:dataset_examples}
\end{figure}

We present in Figure \ref{fig:dataset_examples} several questions with the shortest, medium, and longest answers to provide qualitative observation on VQA benchmarks. We can notice from Figure \ref{fig:question_length_results} and Figure \ref{fig:dataset_examples} that the questions in the OCR-VQA dataset are limited to a group of specific questions. Particularly this group is \{"What is the edition of this book?", "What is the genre of this book?", "What is the title of this book?", "What is the version of this book?", "What is the year printed on this calendar?", "What type of book is this?", "Which year's calendar is this"?, "Who is the author of this book?", "Who wrote this book?", "Is this book related to \textit{<title>}?", "Is this a/an \textit{<category>} book?"\} where \textit{<title>} and \textit{<category>} can be achieved from metadata of online books. This indicates authors of the OCR-VQA dataset collect metadata of books from the internet, then based on this metadata they provide appropriate questions selected from a defined set. For the TextVQA dataset, answers are simply scene texts in images, while in the OpenViVQA dataset answers containing scene text as a factor from images that provide more detailed information. For the VQAv2 dataset, answers are typically short (Figure \ref{fig:answer_length_results}), and its longest answers actually are scene texts in images. The same color as VQAv2 for ViVQA.

\section{Several examples of QA on colors, quantities and directions.} \label{appendix-B}

\begin{figure}[!ht]
    \centering
    \begin{subfigure}{0.8\textwidth}   \includegraphics[width=\textwidth]{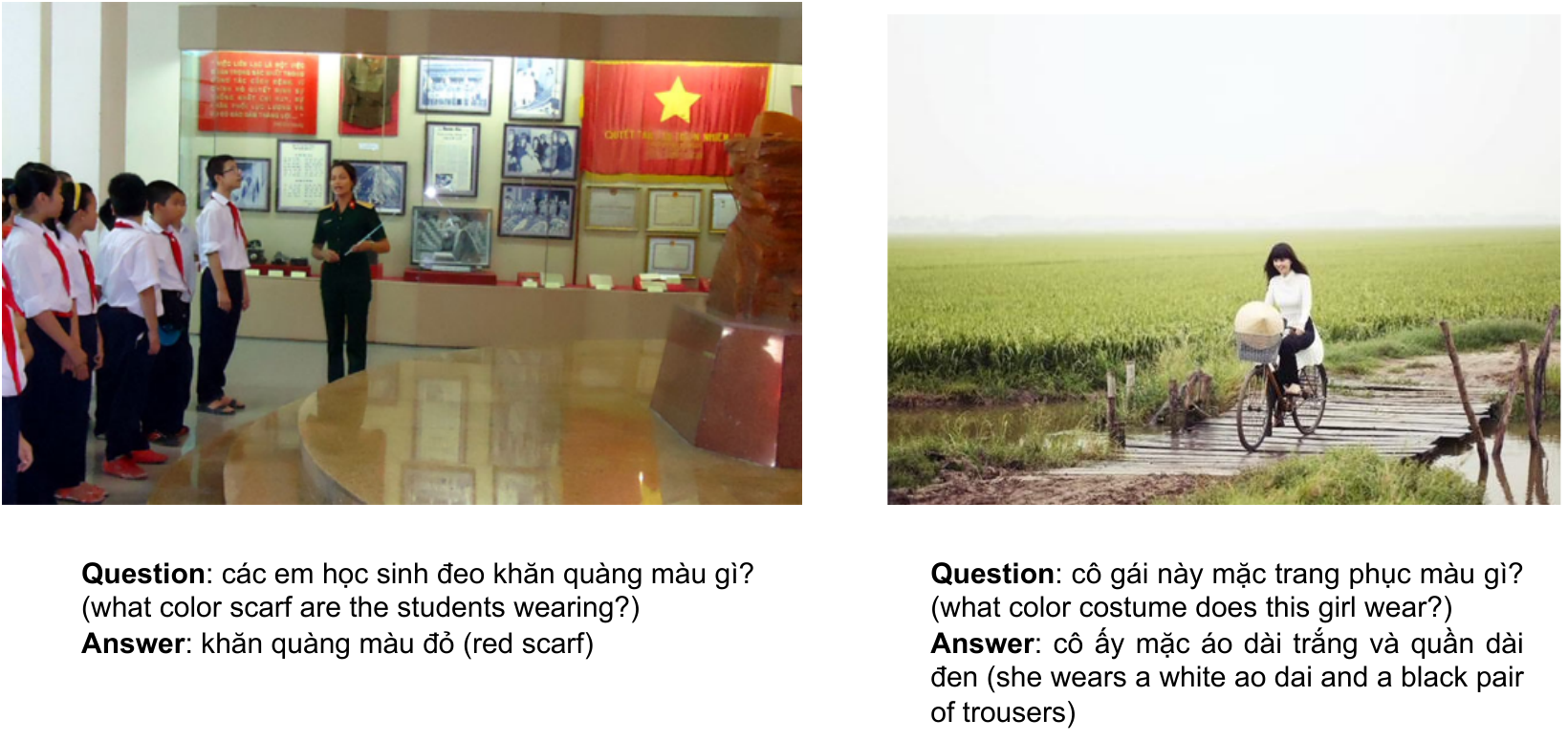}
        \caption{Colors}
    \end{subfigure}
    \begin{subfigure}{0.8\textwidth}
        \includegraphics[width=\textwidth]{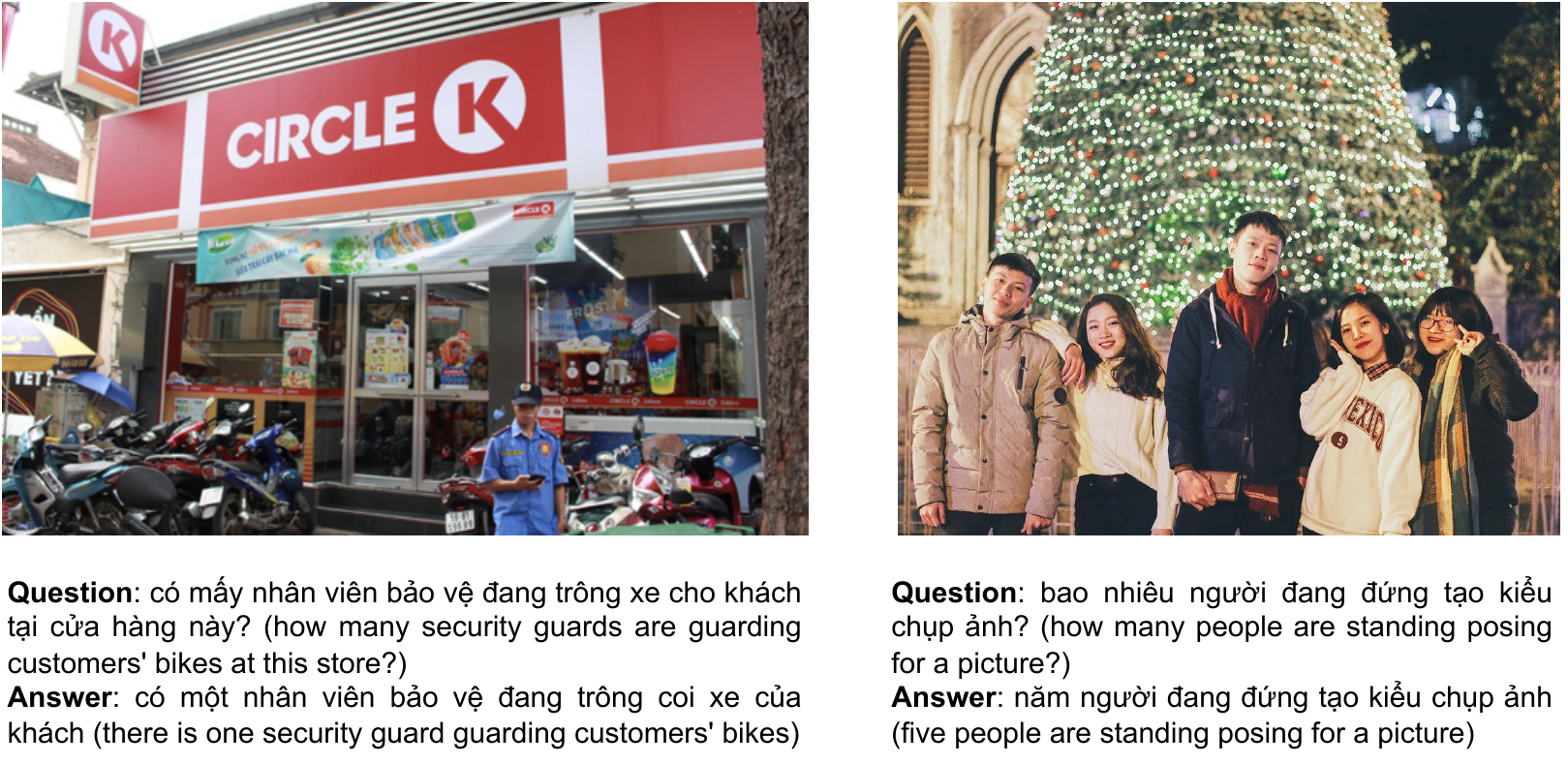}
        \caption{Quantities}
    \end{subfigure}
    \begin{subfigure}{0.8\textwidth}
        \includegraphics[width=\textwidth]{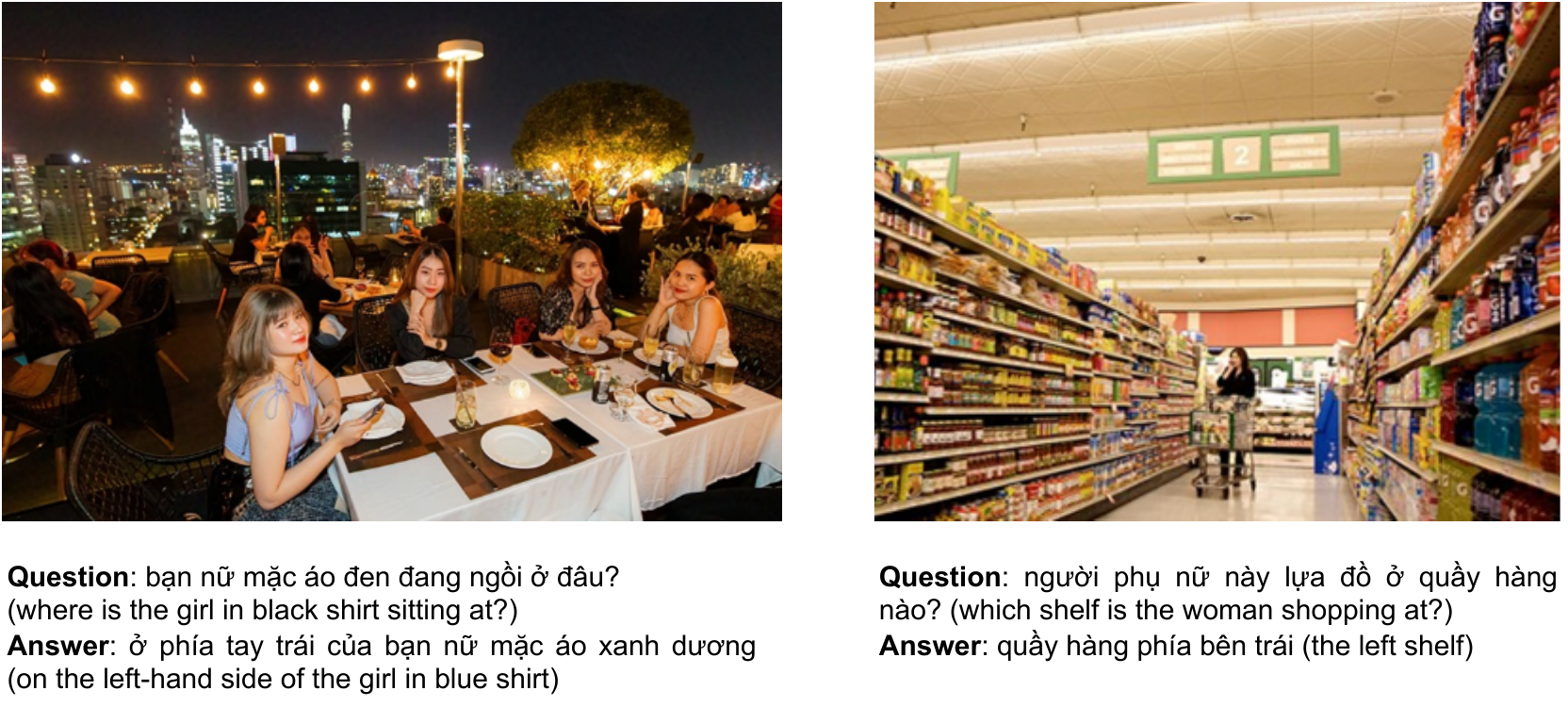}
        \caption{Directions}
    \end{subfigure}
    \caption{Examples of QAs for colors, quantities and directions in the OpenViVQA dataset}
    \label{fig:examples-on-type}
\end{figure}

Here are several typical examples (Figure \ref{fig:examples-on-type}) from the OpenViVQA dataset that we pick to demonstrate the ways our questions exploit the color, quantity and direction factors of visible objects, and also the corresponding answers containing the words that represent such factors.

\section{More examples of MLPAG and M4C on Text QAs} \label{appendix-C}

\begin{figure}[ht]
    \centering
    \includegraphics[width=\textwidth]{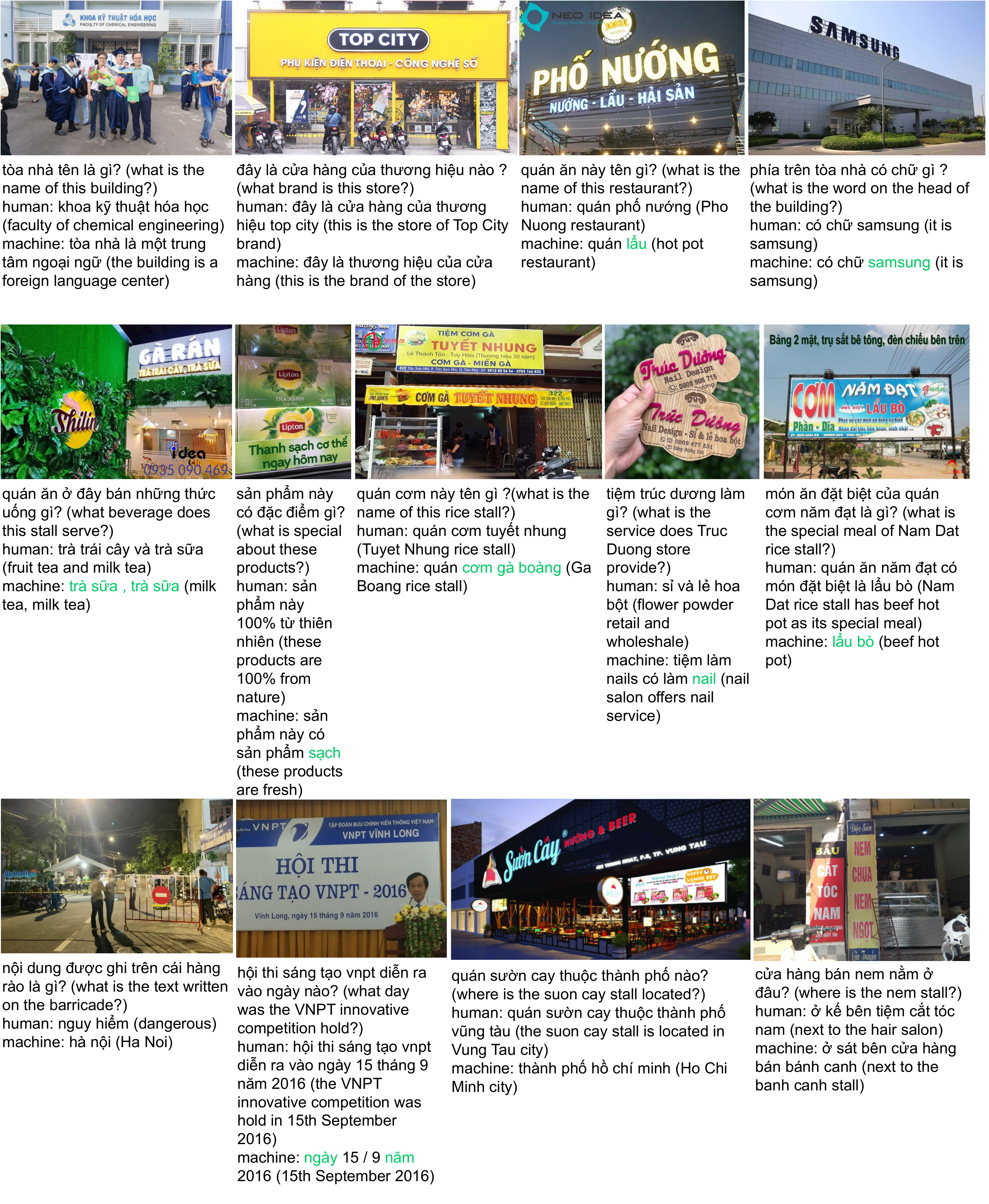}
    \caption{Examples of results of MLPAG. Green texts indicate tokens copied from scene texts in images.}
    \label{fig:examples_glorra}
\end{figure}

\begin{figure}[ht]
    \centering
    \includegraphics[width=\textwidth]{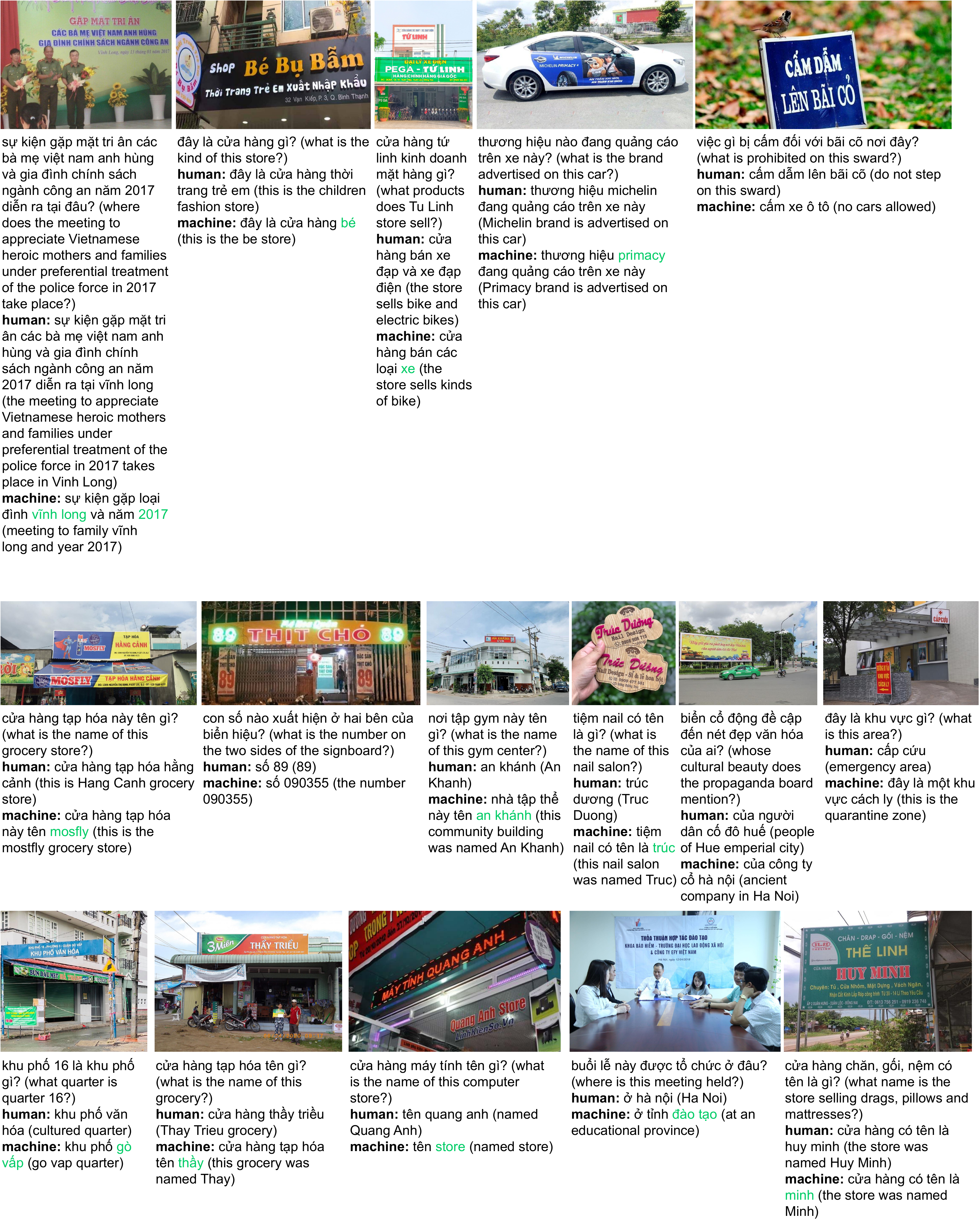}
    \caption{Examples of results of M4C. Green texts indicate tokens copied from scene texts in images.}
    \label{fig:examples_m4c}
\end{figure}

We provided more examples of MLPAG (Figure \ref{fig:examples_glorra}) and M4C (Figure \ref{fig:examples_m4c}) to have a comprehensive observation of their behavior on copying scene texts from images to answers.

\bibliography{main}

\begin{thebibliography}{10}

\bibitem{6857995}
J.~Almazán, A.~Gordo, A.~Fornés, and E.~Valveny.
\newblock Word spotting and recognition with embedded attributes.
\newblock {\em IEEE Transactions on Pattern Analysis and Machine Intelligence},
  36(12):2552--2566, 2014.

\bibitem{Anderson2017BottomUpAT}
P.~Anderson, X.~He, C.~Buehler, D.~Teney, M.~Johnson, S.~Gould, and L.~Zhang.
\newblock Bottom-up and top-down attention for image captioning and visual
  question answering.
\newblock {\em 2018 IEEE/CVF Conference on Computer Vision and Pattern
  Recognition}, pages 6077--6086, 2017.

\bibitem{VQA}
S.~Antol, A.~Agrawal, J.~Lu, M.~Mitchell, D.~Batra, C.~L. Zitnick, and
  D.~Parikh.
\newblock {VQA}: {V}isual {Q}uestion {A}nswering.
\newblock In {\em International Conference on Computer Vision (ICCV)}, 2015.

\bibitem{bahdanau2014neural}
D.~Bahdanau, K.~Cho, and Y.~Bengio.
\newblock Neural machine translation by jointly learning to align and
  translate.
\newblock {\em arXiv preprint arXiv:1409.0473}, 2014.

\bibitem{banerjee-lavie-2005-meteor}
S.~Banerjee and A.~Lavie.
\newblock {METEOR}: An automatic metric for {MT} evaluation with improved
  correlation with human judgments.
\newblock In {\em Proceedings of the {ACL} Workshop on Intrinsic and Extrinsic
  Evaluation Measures for Machine Translation and/or Summarization}, pages
  65--72, Ann Arbor, Michigan, June 2005. Association for Computational
  Linguistics.

\bibitem{bojanowski2017enriching}
P.~Bojanowski, E.~Grave, A.~Joulin, and T.~Mikolov.
\newblock Enriching word vectors with subword information.
\newblock {\em Transactions of the Association for Computational Linguistics},
  5:135--146, 2017.

\bibitem{borisyuk2018rosetta}
F.~Borisyuk, A.~Gordo, and V.~Sivakumar.
\newblock Rosetta: Large scale system for text detection and recognition in
  images.
\newblock In {\em Proceedings of the 24th ACM SIGKDD international conference
  on knowledge discovery \& data mining}, pages 71--79, 2018.

\bibitem{Changpinyo2022TowardsMV}
S.~Changpinyo, L.~Xue, I.~Szpektor, A.~V. Thapliyal, J.~Amelot, X.~Chen, and
  R.~Soricut.
\newblock Towards multi-lingual visual question answering.
\newblock {\em ArXiv}, abs/2209.05401, 2022.

\bibitem{changpinyo2022towards}
S.~Changpinyo, L.~Xue, I.~Szpektor, A.~V. Thapliyal, J.~Amelot, X.~Chen, and
  R.~Soricut.
\newblock Towards multi-lingual visual question answering.
\newblock {\em arXiv preprint arXiv:2209.05401}, 2022.

\bibitem{chen2020uniter}
Y.-C. Chen, L.~Li, L.~Yu, A.~El~Kholy, F.~Ahmed, Z.~Gan, Y.~Cheng, and J.~Liu.
\newblock Uniter: Universal image-text representation learning.
\newblock In {\em Computer Vision--ECCV 2020: 16th European Conference,
  Glasgow, UK, August 23--28, 2020, Proceedings, Part XXX}, pages 104--120.
  Springer, 2020.

\bibitem{cho2020x}
J.~Cho, J.~Lu, D.~Schwenk, H.~Hajishirzi, and A.~Kembhavi.
\newblock X-lxmert: Paint, caption and answer questions with multi-modal
  transformers.
\newblock {\em arXiv preprint arXiv:2009.11278}, 2020.

\bibitem{Devlin2019BERTPO}
J.~Devlin, M.-W. Chang, K.~Lee, and K.~Toutanova.
\newblock Bert: Pre-training of deep bidirectional transformers for language
  understanding.
\newblock {\em ArXiv}, abs/1810.04805, 2019.

\bibitem{fleiss1971measuring}
J.~L. Fleiss.
\newblock Measuring nominal scale agreement among many raters.
\newblock {\em Psychological bulletin}, 76(5):378, 1971.

\bibitem{ganesan2018rouge}
K.~Ganesan.
\newblock Rouge 2.0: Updated and improved measures for evaluation of
  summarization tasks.
\newblock {\em arXiv preprint arXiv:1803.01937}, 2018.

\bibitem{goyal2017making}
Y.~Goyal, T.~Khot, D.~Summers-Stay, D.~Batra, and D.~Parikh.
\newblock Making the v in vqa matter: Elevating the role of image understanding
  in visual question answering.
\newblock In {\em Proceedings of the IEEE conference on computer vision and
  pattern recognition}, pages 6904--6913, 2017.

\bibitem{He2015DeepRL}
K.~He, X.~Zhang, S.~Ren, and J.~Sun.
\newblock Deep residual learning for image recognition.
\newblock {\em 2016 IEEE Conference on Computer Vision and Pattern Recognition
  (CVPR)}, pages 770--778, 2015.

\bibitem{hochreiter1997long}
S.~Hochreiter and J.~Schmidhuber.
\newblock Long short-term memory.
\newblock {\em Neural computation}, 9(8):1735--1780, 1997.

\bibitem{spacy2}
M.~Honnibal and I.~Montani.
\newblock {spaCy 2}: Natural language understanding with {B}loom embeddings,
  convolutional neural networks and incremental parsing.
\newblock To appear, 2017.

\bibitem{hu2020iterative}
R.~Hu, A.~Singh, T.~Darrell, and M.~Rohrbach.
\newblock Iterative answer prediction with pointer-augmented multimodal
  transformers for textvqa.
\newblock In {\em Proceedings of the IEEE/CVF Conference on Computer Vision and
  Pattern Recognition}, pages 9992--10002, 2020.

\bibitem{huang2022swintextspotter}
M.~Huang, Y.~Liu, Z.~Peng, C.~Liu, D.~Lin, S.~Zhu, N.~Yuan, K.~Ding, and
  L.~Jin.
\newblock Swintextspotter: Scene text spotting via better synergy between text
  detection and text recognition.
\newblock In {\em Proceedings of the IEEE/CVF Conference on Computer Vision and
  Pattern Recognition}, pages 4593--4603, 2022.

\bibitem{huang2020pixel}
Z.~Huang, Z.~Zeng, B.~Liu, D.~Fu, and J.~Fu.
\newblock Pixel-bert: Aligning image pixels with text by deep multi-modal
  transformers.
\newblock {\em arXiv preprint arXiv:2004.00849}, 2020.

\bibitem{iwana2016judging}
B.~K. Iwana, S.~T. Raza~Rizvi, S.~Ahmed, A.~Dengel, and S.~Uchida.
\newblock Judging a book by its cover.
\newblock {\em arXiv preprint arXiv:1610.09204}, 2016.

\bibitem{jiang2020defense}
H.~Jiang, I.~Misra, M.~Rohrbach, E.~Learned-Miller, and X.~Chen.
\newblock In defense of grid features for visual question answering.
\newblock In {\em IEEE Conference on Computer Vision and Pattern Recognition
  (CVPR)}, 2020.

\bibitem{Kantharaj2022OpenCQAOQ}
S.~Kantharaj, X.~Do, R.~T.~K. Leong, J.~Q. Tan, E.~Hoque, and S.~R. Joty.
\newblock Opencqa: Open-ended question answering with charts.
\newblock {\em ArXiv}, abs/2210.06628, 2022.

\bibitem{kazemi2017show}
V.~Kazemi and A.~Elqursh.
\newblock Show, ask, attend, and answer: A strong baseline for visual question
  answering.
\newblock {\em arXiv preprint arXiv:1704.03162}, 2017.

\bibitem{Kingma2014AdamAM}
D.~P. Kingma and J.~Ba.
\newblock Adam: A method for stochastic optimization.
\newblock {\em CoRR}, abs/1412.6980, 2014.

\bibitem{OpenImages}
A.~Kuznetsova, H.~Rom, N.~Alldrin, J.~Uijlings, I.~Krasin, J.~Pont-Tuset,
  S.~Kamali, S.~Popov, M.~Malloci, A.~Kolesnikov, T.~Duerig, and V.~Ferrari.
\newblock The open images dataset v4: Unified image classification, object
  detection, and visual relationship detection at scale.
\newblock {\em IJCV}, 2020.

\bibitem{li2020unicoder}
G.~Li, N.~Duan, Y.~Fang, M.~Gong, and D.~Jiang.
\newblock Unicoder-vl: A universal encoder for vision and language by
  cross-modal pre-training.
\newblock In {\em Proceedings of the AAAI Conference on Artificial
  Intelligence}, volume~34, pages 11336--11344, 2020.

\bibitem{li2019visualbert}
L.~H. Li, M.~Yatskar, D.~Yin, C.-J. Hsieh, and K.-W. Chang.
\newblock Visualbert: A simple and performant baseline for vision and language.
\newblock {\em arXiv preprint arXiv:1908.03557}, 2019.

\bibitem{li2020oscar}
X.~Li, X.~Yin, C.~Li, P.~Zhang, X.~Hu, L.~Zhang, L.~Wang, H.~Hu, L.~Dong,
  F.~Wei, et~al.
\newblock Oscar: Object-semantics aligned pre-training for vision-language
  tasks.
\newblock In {\em Computer Vision--ECCV 2020: 16th European Conference,
  Glasgow, UK, August 23--28, 2020, Proceedings, Part XXX 16}, pages 121--137.
  Springer, 2020.

\bibitem{lin2014microsoft}
T.-Y. Lin, M.~Maire, S.~Belongie, J.~Hays, P.~Perona, D.~Ramanan,
  P.~Doll{\'a}r, and C.~L. Zitnick.
\newblock Microsoft coco: Common objects in context.
\newblock In {\em Computer Vision--ECCV 2014: 13th European Conference, Zurich,
  Switzerland, September 6-12, 2014, Proceedings, Part V 13}, pages 740--755.
  Springer, 2014.

\bibitem{lu2019vilbert}
J.~Lu, D.~Batra, D.~Parikh, and S.~Lee.
\newblock Vilbert: Pretraining task-agnostic visiolinguistic representations
  for vision-and-language tasks.
\newblock {\em Advances in neural information processing systems}, 32, 2019.

\bibitem{lu2022unified}
J.~Lu, C.~Clark, R.~Zellers, R.~Mottaghi, and A.~Kembhavi.
\newblock Unified-io: A unified model for vision, language, and multi-modal
  tasks.
\newblock {\em arXiv preprint arXiv:2206.08916}, 2022.

\bibitem{lu2016hierarchical}
J.~Lu, J.~Yang, D.~Batra, and D.~Parikh.
\newblock Hierarchical question-image co-attention for visual question
  answering.
\newblock {\em Advances in neural information processing systems}, 29, 2016.

\bibitem{luong2015effective}
M.-T. Luong, H.~Pham, and C.~D. Manning.
\newblock Effective approaches to attention-based neural machine translation.
\newblock {\em arXiv preprint arXiv:1508.04025}, 2015.

\bibitem{Mathew_2021_WACV}
M.~Mathew, D.~Karatzas, and C.~Jawahar.
\newblock Docvqa: A dataset for vqa on document images.
\newblock In {\em Proceedings of the IEEE/CVF Winter Conference on Applications
  of Computer Vision (WACV)}, pages 2200--2209, January 2021.

\bibitem{mishraICDAR19}
A.~Mishra, S.~Shekhar, A.~K. Singh, and A.~Chakraborty.
\newblock Ocr-vqa: Visual question answering by reading text in images.
\newblock In {\em ICDAR}, 2019.

\bibitem{phonlp}
L.~T. Nguyen and D.~Q. Nguyen.
\newblock {PhoNLP: A joint multi-task learning model for Vietnamese
  part-of-speech tagging, named entity recognition and dependency parsing}.
\newblock In {\em Proceedings of the 2021 Conference of the North American
  Chapter of the Association for Computational Linguistics: Demonstrations},
  pages 1--7, 2021.

\bibitem{m_Nguyen-etal-CVPR21}
N.~Nguyen, T.~Nguyen, V.~Tran, T.~Tran, T.~Ngo, T.~Nguyen, and M.~Hoai.
\newblock Dictionary-guided scene text recognition.
\newblock In {\em Proceedings of the {IEEE} Conference on Computer Vision and
  Pattern Recognition (CVPR)}, 2021.

\bibitem{papineni-etal-2002-bleu}
K.~Papineni, S.~Roukos, T.~Ward, and W.-J. Zhu.
\newblock {B}leu: a method for automatic evaluation of machine translation.
\newblock In {\em Proceedings of the 40th Annual Meeting of the Association for
  Computational Linguistics}, pages 311--318, Philadelphia, Pennsylvania, USA,
  July 2002. Association for Computational Linguistics.

\bibitem{pennington-etal-2014-glove}
J.~Pennington, R.~Socher, and C.~Manning.
\newblock {G}lo{V}e: Global vectors for word representation.
\newblock In {\em Proceedings of the 2014 Conference on Empirical Methods in
  Natural Language Processing ({EMNLP})}, pages 1532--1543, Doha, Qatar, Oct.
  2014. Association for Computational Linguistics.

\bibitem{ren2015faster}
S.~Ren, K.~He, R.~Girshick, and J.~Sun.
\newblock Faster r-cnn: Towards real-time object detection with region proposal
  networks.
\newblock {\em Advances in neural information processing systems}, 28, 2015.

\bibitem{singh2019towards}
A.~Singh, V.~Natarajan, M.~Shah, Y.~Jiang, X.~Chen, D.~Batra, D.~Parikh, and
  M.~Rohrbach.
\newblock Towards vqa models that can read.
\newblock In {\em Proceedings of the IEEE/CVF conference on computer vision and
  pattern recognition}, pages 8317--8326, 2019.

\bibitem{su2019vl}
W.~Su, X.~Zhu, Y.~Cao, B.~Li, L.~Lu, F.~Wei, and J.~Dai.
\newblock Vl-bert: Pre-training of generic visual-linguistic representations.
\newblock {\em arXiv preprint arXiv:1908.08530}, 2019.

\bibitem{tan2019lxmert}
H.~Tan and M.~Bansal.
\newblock Lxmert: Learning cross-modality encoder representations from
  transformers.
\newblock {\em arXiv preprint arXiv:1908.07490}, 2019.

\bibitem{visualmrc}
R.~Tanaka, K.~Nishida, and S.~Yoshida.
\newblock Visualmrc: Machine reading comprehension on document images.
\newblock {\em Proceedings of the AAAI Conference on Artificial Intelligence},
  35(15):13878--13888, May 2021.

\bibitem{teney2018tips}
D.~Teney, P.~Anderson, X.~He, and A.~Van Den~Hengel.
\newblock Tips and tricks for visual question answering: Learnings from the
  2017 challenge.
\newblock In {\em Proceedings of the IEEE conference on computer vision and
  pattern recognition}, pages 4223--4232, 2018.

\bibitem{tran-etal-2021-vivqa-vietnamese}
K.~Q. Tran, A.~T. Nguyen, A.~T.-H. Le, and K.~V. Nguyen.
\newblock Vivqa: Vietnamese visual question answering.
\newblock In {\em Proceedings of the 35th Pacific Asia Conference on Language,
  Information and Computation}, pages 546--554, Shanghai, China, 11 2021.
  Association for Computational Lingustics.

\bibitem{vaswani2017attention}
A.~Vaswani, N.~Shazeer, N.~Parmar, J.~Uszkoreit, L.~Jones, A.~N. Gomez,
  {\L}.~Kaiser, and I.~Polosukhin.
\newblock Attention is all you need.
\newblock {\em Advances in neural information processing systems}, 30, 2017.

\bibitem{vedantam2015cider}
R.~Vedantam, C.~Lawrence~Zitnick, and D.~Parikh.
\newblock Cider: Consensus-based image description evaluation.
\newblock In {\em Proceedings of the IEEE conference on computer vision and
  pattern recognition}, pages 4566--4575, 2015.

\bibitem{vu-etal-2018-vncorenlp}
T.~Vu, D.~Q. Nguyen, D.~Q. Nguyen, M.~Dras, and M.~Johnson.
\newblock {V}n{C}ore{NLP}: A {V}ietnamese natural language processing toolkit.
\newblock In {\em Proceedings of the 2018 Conference of the North {A}merican
  Chapter of the Association for Computational Linguistics: Demonstrations},
  pages 56--60, New Orleans, Louisiana, June 2018. Association for
  Computational Linguistics.

\bibitem{wang2021simvlm}
Z.~Wang, J.~Yu, A.~W. Yu, Z.~Dai, Y.~Tsvetkov, and Y.~Cao.
\newblock Simvlm: Simple visual language model pretraining with weak
  supervision.
\newblock {\em arXiv preprint arXiv:2108.10904}, 2021.

\bibitem{worley2015open}
P.~Worley.
\newblock Open thinking, closed questioning: Two kinds of open and closed
  question.
\newblock {\em Journal of Philosophy in Schools}, 2015.

\bibitem{yang2016stacked}
Z.~Yang, X.~He, J.~Gao, L.~Deng, and A.~Smola.
\newblock Stacked attention networks for image question answering.
\newblock In {\em Proceedings of the IEEE conference on computer vision and
  pattern recognition}, pages 21--29, 2016.

\bibitem{yu2019deep}
Z.~Yu, J.~Yu, Y.~Cui, D.~Tao, and Q.~Tian.
\newblock Deep modular co-attention networks for visual question answering.
\newblock In {\em Proceedings of the IEEE/CVF conference on computer vision and
  pattern recognition}, pages 6281--6290, 2019.

\bibitem{ZHANG2019268}
D.~Zhang, R.~Cao, and S.~Wu.
\newblock Information fusion in visual question answering: A survey.
\newblock {\em Information Fusion}, 52:268--280, 2019.

\bibitem{zhang2021vinvl}
P.~Zhang, X.~Li, X.~Hu, J.~Yang, L.~Zhang, L.~Wang, Y.~Choi, and J.~Gao.
\newblock Vinvl: Revisiting visual representations in vision-language models.
\newblock In {\em Proceedings of the IEEE/CVF Conference on Computer Vision and
  Pattern Recognition}, pages 5579--5588, 2021.

\bibitem{ZHANG20211}
S.~Zhang, M.~Chen, J.~Chen, F.~Zou, Y.-F. Li, and P.~Lu.
\newblock Multimodal feature-wise co-attention method for visual question
  answering.
\newblock {\em Information Fusion}, 73:1--10, 2021.

\bibitem{ZHANG2020116}
W.~Zhang, J.~Yu, H.~Hu, H.~Hu, and Z.~Qin.
\newblock Multimodal feature fusion by relational reasoning and attention for
  visual question answering.
\newblock {\em Information Fusion}, 55:116--126, 2020.

\bibitem{ZHENG202114}
W.~Zheng, L.~Yan, C.~Gou, and F.-Y. Wang.
\newblock Km4: Visual reasoning via knowledge embedding memory model with
  mutual modulation.
\newblock {\em Information Fusion}, 67:14--28, 2021.

\end{thebibliography}

\bibliographystyle{abbrv}







\end{document}